%% file: main.tex
\documentclass{article}

\usepackage{arxiv}

\usepackage[utf8]{inputenc} 
\usepackage[T1]{fontenc}    
\usepackage{url}            
\usepackage{booktabs}       
\usepackage{amsfonts}       
\usepackage{nicefrac}       
\usepackage{microtype}      
\usepackage{lipsum}
\usepackage{amsmath}
\usepackage{amsthm}
\usepackage{amssymb}   
\usepackage{bigints}
\usepackage{bm}
\usepackage{diagbox}
\usepackage{makecell}
\usepackage{multirow}
\usepackage{caption}
\usepackage{subcaption}
\usepackage{graphicx}
\usepackage[ruled,vlined]{algorithm2e}
\include{pythonlisting}
\usepackage{xcolor}
\usepackage{mathtools}
\usepackage{enumitem}

\usepackage{hyperref}
\hypersetup{
    colorlinks=true,
    citecolor=black,
    linkcolor=black,
    filecolor=black,
    urlcolor=black,
}

\newcommand{\R}{\mathbb{R}}

\theoremstyle{definition}

\numberwithin{equation}{section}

\theoremstyle{plain}

\newtheorem{theorem}{Theorem}[section]

\newtheorem{remark}[theorem]{Remark}

\title{Learning the solution operator of parametric partial differential equations with physics-informed DeepOnets}

\author{
  Sifan Wang \\
  Graduate Group in Applied Mathematics \\
  and Computational Science \\
  University of Pennsylvania\\
  Philadelphia, PA 19104 \\
  \texttt{sifanw@sas.upenn.edu} \\
  \And
  Hanwen Wang \\
  Graduate Group in Applied Mathematics \\
  and Computational Science \\
  University of Pennsylvania\\
  Philadelphia, PA 19104 \\
  \texttt{wangh19@sas.upenn.edu} \\
  \And
  Paris Perdikaris \\
  Department of Mechanichal Engineering \\
  and Applied Mechanics\\
  University of Pennsylvania\\
  Philadelphia, PA 19104 \\
  \texttt{pgp@seas.upenn.edu} \\
}

\begin{document}
\maketitle
\begin{abstract}
Deep operator networks (DeepONets) are receiving increased  attention thanks to their demonstrated capability to approximate  nonlinear operators between infinite-dimensional Banach spaces. However, despite their remarkable early promise, they typically require large training data-sets consisting of paired input-output observations which may be expensive to obtain, while their predictions may not be consistent with the underlying physical principles that generated the observed data. In this work,  we propose a novel model class coined as physics-informed DeepONets,  which introduces an effective regularization mechanism for biasing the outputs of DeepOnet models towards ensuring physical consistency. This is accomplished by leveraging automatic differentiation to impose the underlying physical laws via soft penalty constraints during model training. We demonstrate that this simple, yet remarkably effective extension can not only yield a significant improvement in the predictive accuracy of DeepOnets, but also greatly reduce the need for large training data-sets. To this end, a remarkable observation is that physics-informed DeepONets are capable of solving parametric partial differential equations (PDEs) without any paired input-output observations, except for a set of given initial or boundary conditions. We illustrate the effectiveness of the proposed framework through a series of comprehensive numerical studies across various types of PDEs.  Strikingly, a trained physics informed DeepOnet model can predict the solution of $\mathcal{O}(10^3)$ time-dependent PDEs in a fraction of a second -- up to three orders of magnitude faster compared a conventional PDE solver. The data and code accompanying this manuscript are publicly available at \url{https://github.com/PredictiveIntelligenceLab/Physics-informed-DeepONets}.
\end{abstract}


\section{Introduction}

As machine learning (ML) methodologies take center stage across diverse disciplines in science and engineering, there is an increased interest in adopting data-driven methods to analyze, emulate, and optimize complex physical systems. The dynamic behavior of such systems is often described by conservation and constitutive laws expressed as systems of partial differential equations (PDEs) \cite{courant2008methods}. A classical task then involves the use of analytical or computational tools to solve such equations across a range of scenarios, e.g. different domain geometries, input parameters, initial and boundary conditions (IBCs). Mathematically speaking, solving these so-called {\em parametric PDE} problems involves learning the {\em solution operator} that maps variable input entities to the corresponding latent solutions of the underlying PDE system. Tackling this task using traditional tools (e.g. finite element methods \cite{hughes2012finite}) bears a formidable cost, as independent simulations need to be performed for every different domain geometry, input parameter, or IBCs. This challenge has motivated a growing literature on reduced-order methods \cite{lucia2004reduced, kutz2016dynamic, benner2017model, schilders2008model, quarteroni2014reduced, mezic2005spectral, peherstorfer2016data} that leverage existing data-sets to build fast emulators, often at the price of reduced accuracy, stability, and generalization performance \cite{majda2018strategies, lassila2014model}. More recently, ML tools are actively developed to infer solutions of PDEs  \cite{psichogios1992hybrid, lagaris1998artificial, raissi2019physics, sun2020surrogate, zhu2019physics, karumuri2020simulator, sirignano2018dgm}, however most existing tools can only accommodate a fixed given set of input parameters or IBCs. Nevertheless, these approaches have found wide applicability across diverse applications including fluid mechanics \cite{raissi2020hidden, tartakovsky2020physics}, heat transfer \cite{hennigh2020nvidia}, bio-engineering \cite{kissas2020machine,sahli2020physics}, materials \cite{lu2020extraction, chen2020physics, goswami2020transfer}, and finance \cite{elbrachter2018dnn, han2018solving}, showcasing the remarkable effectiveness of ML techniques in learning black-box functions, even in high-dimensional contexts \cite{poggio2017and}. A natural question then arises: can ML methods be effective in building fast emulators for solving {\it parametric} PDEs?

Solving {\it parametric} PDEs requires learning {\it operators} (i.e. maps between infinite-dimensional function spaces) instead of functions (i.e. maps between finite-dimensional vector spaces), thus defining a new and relatively under-explored realm for ML-based approaches. 
Neural operator methods \cite{li2020neural, li2020multipole, li2020fourier} represent the solution map of parametric PDEs as an integral Hilbert-Schmidt operator, whose kernel is parametrized and learned from paired observations, either using local message passing on a graph-based discretization of the physical domain \cite{li2020neural, li2020multipole}, or using global Fourier approximations in the frequency domain \cite{li2020fourier}. By construction, neural operators methods are resolution independent (i.e. the model can be queried at any arbitrary input location), but they require large training data-sets, while their involved implementation often leads to slow and computationally expensive training loops. More recently, Lu {\em et al.} \cite{lu2019deeponet} has presented a novel operator learning architecture coined as DeepOnet that is motivated by the universal approximation theorem for operators \cite{chen1995universal, back2002universal}. DeepOnets still require large annotated data-sets consisting of paired input-output observations, but they provide a simple and intuitive model architecture that is fast to train, while allowing for a continuous representation of the target output functions that is independent of resolution. Beyond deep learning approaches, operator-valued kernel methods \cite{kadri2016operator, griebel2017reproducing} have also been demonstrated as a powerful tool for learning nonlinear operators, and they can naturally be generalized to neural networks acting on function spaces \cite{owhadi2020ideas}, but their applicability is generally limited due to their computational cost. Here we should again stress that the aforementioned techniques enable inference in abstract infinite-dimensional Banach spaces \cite{nelsen2020random}; a paradigm shift from current machine learning practice that mainly focuses on learning functions instead of operators. In fact, recent theoretical findings also suggest that the sample complexity of DeepOnets can circumvent the curse of dimensionality in certain scenarios \cite{lanthaler2021error}.

While the aforementioned methodologies have demonstrated early promise across a range of applications \cite{cai2020deepm, lin2020operator, liu2021learning}, their application to solving parametric PDEs faces two fundamental challenges. First, they require a large corpus of paired input-output observations. In many realistic scenarios, the acquisition of such data involves the repeated evaluation of expensive experiments or costly high-fidelity simulators, thus generating sufficient large training data-sets may be prohibitively expensive. In fact, ideally, one would wish to be able to train such models without any observed data at all (i.e. given only knowledge of the PDE form and its corresponding ICBs). The second challenge relates to the fact that, by construction, the methods outlined above can only return a crude approximation to the target solution operator in the sense that the predicted output functions are not guaranteed to satisfy the underlying PDE. Recent efforts \cite{khoo2017solving, zhu2019physics, geneva2020modeling, chen2020meta, kochkov2021machine} attempt to address some of these challenges by designing appropriate architectures and loss functions for learning {\em discretized operators} (i.e. maps between high-dimensional Euclidean spaces). Although these approaches can relax the requirement for paired input-output training data, they are limited by the resolution of their underlying mesh discretization, and, consequently, need modifications to their architecture for different resolutions/discretizations in order to achieve consistent convergence (if at all possible, as demonstrated in \cite{li2020neural}).

In this work, we aim to address the aforementioned challenges by exploring a simple, yet remarkably effective extension of the DeepONet framework \cite{lu2019deeponet}. Drawing motivation from physics-informed neural networks \cite{raissi2019physics}, we recognize that the outputs of a DeepONet model are differentiable with respect to their input coordinates, therefore allowing us to employ automatic differentiation \cite{griewank1989automatic, baydin2017automatic} to formulate an appropriate regularization mechanism for biasing the target output functions to satisfy the underlying PDE constraints. This yields a simple procedure for training {\em physics-informed DeepONet} models  even in the absence of any training data for the latent output functions, except for the appropriate IBCs of a given PDE system. By constraining the outputs of a DeepONet to approximately satisfy an underlying governing law, we observe significant improvements in predictive accuracy (up to 1-2 orders of magnitude reduction in predictive errors), enhanced generalization performance even for out-of-distribution prediction and extrapolation tasks, as well as enhanced data-efficiency (up to 100\% reduction in the number of examples required to train a DeepONet model). 
As such, we demonstrate how physics-informed DeepONet models can be used to solve parametric PDEs without any paired input-output observations; a setting for which existing approaches for operator learning in Banach spaces fall short. Moreover, a trained physics-informed DeepONet model can generate PDE solutions up to three orders of magnitude faster compared to traditional PDE solvers. Taken together, the computational infrastructure developed in this work can have broad technical impact in reducing computational costs and accelerating scientific modeling of complex non-linear, non-equilibrium processes across diverse applications including engineering design and control, Earth system science, and computational biology.

The remaining of this paper is structured as follows. In section \ref{sec: deeponet}, we provide an overview of the DeepONet framework put forth by Lu {\it et al.} \cite{lu2019deeponet}. Section \ref{sec: physics_informed_deeponet} provides a detailed discussion of our main technical contribution, namely the formulation of a physics-informed regularization mechanism for constraining the outputs of a DeepONet model to approximately satisfy an underlying PDE system. Our discussion is accompanied by an illustrative example that highlights the main advantages of the proposed approach in comparison to conventional DeepONets. Further, in section \ref{sec:results} we present a series of comprehensive numerical studies to assess the performance of the proposed physics-informed DeepONet framework. Finally, section \ref{sec:discussion} concludes with a discussion of our main findings, potential pitfalls, and shortcomings, as well as future research directions emanating from this study.

\section{Learning operators with DeepONets}
\label{sec: deeponet}

\begin{figure}
    \centering
    \includegraphics[width=0.8\textwidth]{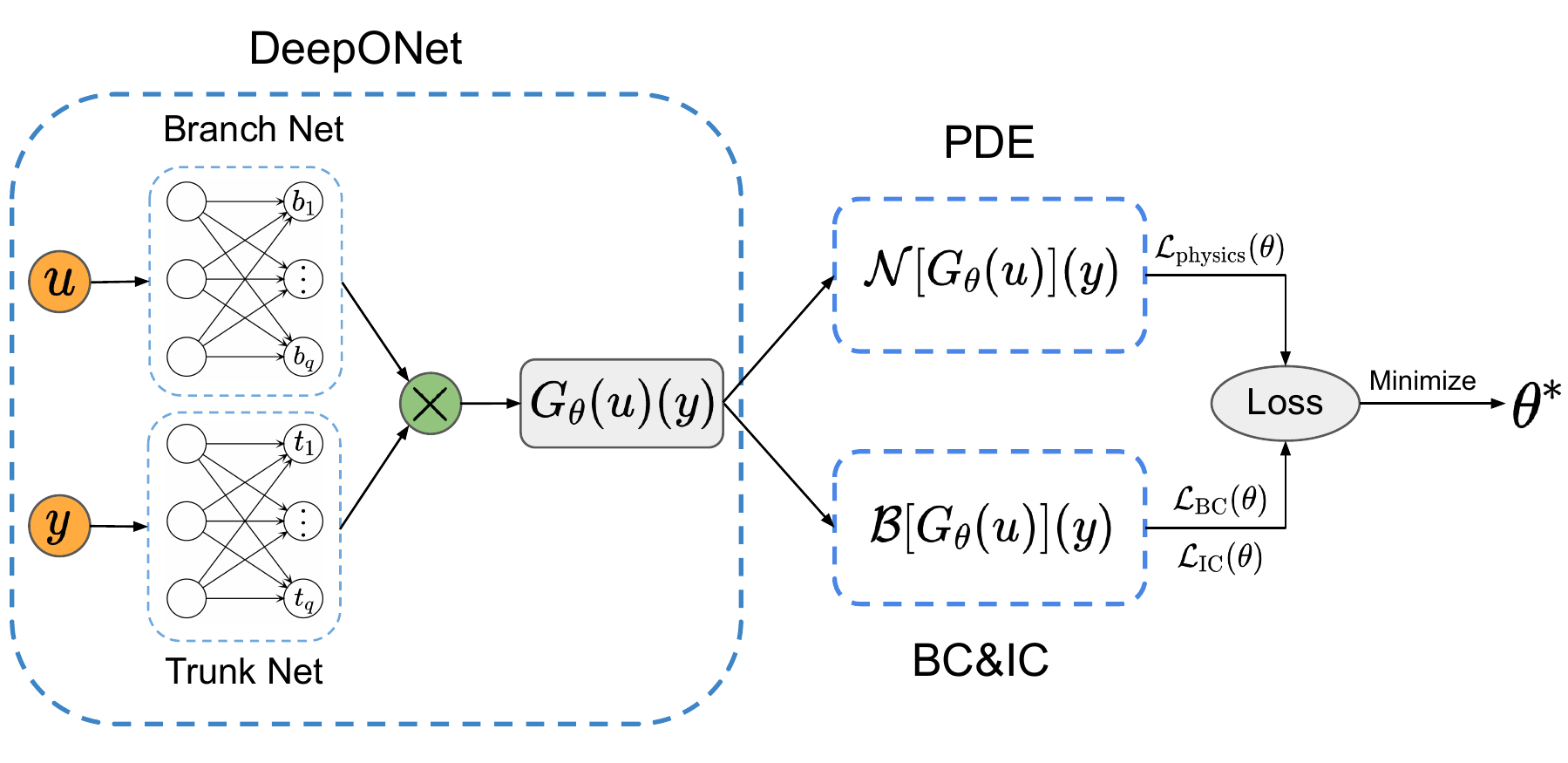}
    \caption{{\em Making DeepOnets physics-informed:}  The DeepONet architecture \cite{lu2019deeponet} consists of two sub-networks, the branch net for extracting latent  representations of input functions, and the trunk net for extracting latent representations of input coordinates at which the output functions are evaluated. A continuous and differentiable representation of the output functions is then obtained by merging the latent representations extracted by each sub-network via a dot product. Automatic differentiation can then be employed to formulate appropriate regularization mechanisms for biasing the DeepOnet outputs to satisfy a given system of PDEs.}
    \label{fig: deepOnet_architecture}
\end{figure}

In this section, we present a brief overview of the DeepONet model architecture \cite{lu2019deeponet} with a special focus on learning solution operators of parametric PDEs.  Here, the terminology "parametric PDEs" means that some parameters of a given PDE system are allowed to vary over a certain range. These input parameters may include, but are not limited to, the shape of the physical domain, the initial or boundary conditions, constant or variable coefficients (e.g. diffusion or reaction rates), source terms, etc. To describe such problems in their full generality, let $(\mathcal{U}, \mathcal{V}, \mathcal{S})$ be a triplet of Banach spaces and $\mathcal{N}: \mathcal{U} \times \mathcal{S} \rightarrow \mathcal{V}$ be a linear or nonlinear differential operator. We consider a parametric PDEs taking the form
\begin{align}
    \label{eq: parametric_PDE}
    \mathcal{N}(\bm{u}, \bm{s}) = 0,
\end{align}
where $\bm{u} \in \mathcal{U}$ denotes the parameters (i.e. input functions), and $\bm{s} \in \mathcal{S}$ is the corresponding unknown solutions of the PDE system. Specifically, we assume that, for any $\bm{u} \in \mathcal{U}$, there exists an unique solution $\bm{s} = \bm{s}(\bm{u}) \in \mathcal{U}$ to \ref{eq: parametric_PDE} (subject to appropriate initial and boundary conditions). Then, we can define the solution operator $G: \mathcal{A} \rightarrow \mathcal{U}$ as
\begin{align}
    G(\bm{u}) = \bm{s}(\bm{u})
\end{align}
Following the original formulation of Lu {\em et. al.} \cite{lu2019deeponet}, we represent the solution map $G$ by an unstacked DeepONet $G_{\bm{\theta}}$, where $\bm{\theta}$ denotes all trainable parameters of the DeepOnet network. As illustrated in Figure \ref{fig: deepOnet_architecture}, the unstacked DeepONet is composed of two separate neural networks referred to as the "branch net" and "trunk net", respectively. The branch net takes $\bm{u}$ as input and returns a features embedding $[b_1, b_2,\dots, b_q]^T \in \R^q$ as output, where $\bm{u} = [\bm{u}(\bm{x}_1), \bm{u}(\bm{x}_2), \dots, \bm{u}(\bm{x}_m) ]$ represents a function $\bm{u} \in \mathcal{U}$ evaluated at a collection of fixed locations $\{\bm{x}_i\}_{i=1}^m$.  The trunk net takes the continuous coordinates $\bm{y}$ as inputs, and outputs a features embedding $[t_1, t_2,\dots, t_q]^T \in \R^q$. To obtain the final output of the DeepONet, the outputs of the branch and trunk networks are merged together via a dot product. More specifically, a DeepONet $G_{\bm \theta}$ prediction of a function $\bm{u}$ evaluated at $\bm{y}$  can be expressed by
\begin{align}
    G_{\bm{\theta}}(\bm{u})(\bm{y}) = \sum_{k=1}^{q} \underbrace{b_{k}\left(\bm{u}\left(\bm{x}_{1}\right), \bm{u}\left(\bm{x}_{2}\right), \ldots, \bm{u}\left(\bm{x}_{m}\right)\right)}_{\text {branch }} \underbrace{t_{k}(\bm{y})}_{\text {trunk }},
\end{align}
where $\bm{\theta}$ denotes the collection of all trainable weight and bias parameters in the branch and trunk networks. These parameters can be optimized by minimizing the following mean square error loss
\begin{align}
\label{eq: loss_operator}
\begin{aligned}
    \mathcal{L}(\bm{\theta}) &= \frac{1}{NP} \sum_{i=1}^N \sum_{j=1}^P \left|G_{\bm{\theta}} (\bm{u}^{(i)})(\bm{y}^{(i)}_j)- G (\bm{u}^{(i)})(\bm{y}^{(i)}_j) \right|^2  \\
    &= \frac{1}{NP} \sum_{i=1}^N \sum_{j=1}^P \left| \sum_{k=1}^q b_k(\bm{u}^{(i)}(\bm{x}_1), \dots, \bm{u}^{(i)}(\bm{x}_m))t_k(\bm{y}^{(i)}_j)  - G (\bm{u}^{(i)})(\bm{y}^{(i)}_j) \right|^2,
\end{aligned}
\end{align}
where $\{ \bm{u}^{(i)} \}_{i=1}^N$ denotes $N$ separate input functions sampled from $\mathcal{U}$. For each  $\bm{u}^{(i)}$, $\{\bm{y}^{(i)}\}_{j=1}^P$ are $P$ locations in the domain of $G(\bm{u}^{(i)})$, and $G(\bm{u}^{(i)})(\bm{y}_j^{(i)})$ is the corresponding output data evaluated at $\bm{y}_j^{(i)}$ .  Contrary to the fixed sensor locations of $\{x_i\}_{i=1}^m$, we remark that the locations of $\{\bm{y}^{(i)}\}_{j=1}^P$ may vary for different $i$, thus allowing us to construct a continuous representation of the output functions $\bm{s} \in \mathcal{S}$.

\begin{remark}
In general, a DeepOnet training data-set is a triplet $[\bm{u}, \bm{y}, G(\bm{u})(\bm{y})]$ with following structure
\begin{align}
    [\bm{u}, \bm{y}, G(\bm{u})(\bm{y})]
    = \begin{bmatrix}
       \begin{bmatrix}
       \vdots \\
       \bm{u}^{(i)}(\bm{x}_1), \bm{u}^{(i)}(\bm{x}_2), \cdots , \bm{u}^{(i)}(\bm{x}_m)\\
        \bm{u}^{(i)}(\bm{x}_1), \bm{u}^{(i)}(\bm{x}_2), \cdots , \bm{u}^{(i)}(\bm{x}_m)\\
        \vdots\\
         \bm{u}^{(i)}(\bm{x}_1), \bm{u}^{(i)}(\bm{x}_2), \cdots , \bm{u}^{(i)}(\bm{x}_m)\\
       \vdots
       \end{bmatrix},
      \begin{bmatrix}
      \vdots\\
      \bm{y}^{(i)}_1\\
       \bm{y}^{(i)}_2\\
       \vdots\\
        \bm{y}^{(i)}_P\\
        \vdots
      \end{bmatrix},
      \begin{bmatrix}
      \vdots\\
      G(\bm{u}^{(i)})(\bm{y}^{(i)}_1)\\
      G(\bm{u}^{(i)})(\bm{y}^{(i)}_2)\\
      \vdots\\
      G(\bm{u}^{(i)})(\bm{y}^{(i)}_P)\\
      \vdots
      \end{bmatrix}
      \end{bmatrix}.
\end{align}
It is important to highlight that each input function $\bm{u}^{(i)} = [  \bm{u}^{(i)}(\bm{x}_1), \bm{u}^{(i)}(\bm{x}_2), \cdots , \bm{u}^{(i)}(\bm{x}_m)]$ repeats itself for $P$ times.
In other words, suppose that $\bm{u} \in \mathcal{U}, G(\bm{u}) \in \mathcal{V}$ are scalar-valued functions,   $\{\bm{u}^{(i)}\}_{i=1}^N$ are $N$  sample functions, and, for each sample $\bm{u}^{(i)}$, $G(\bm{u}^{(i)})$ is evaluated at $P$ different locations $\{\bm{y}^{(i)}\}_{j=1}^P \subset \R^d$, then the tensor dimensions constituting a DeepOnet training data-set $\bm{u}, \bm{y}, G(\bm{u})(\bm{y})$ are $(N \times P, m), (N \times P, d), (N \times P, 1)$ respectively.
\end{remark}



Although DeepONets \cite{lu2019deeponet} and their variants (e.g. DeepM\&Mnets \cite{cai2020deepm}) have demonstrated great potential in approximating operators and solving multi-physics and multi-scale problems, it is worth pointing out that the learned operator may not be consistent with the underlying  physical laws that generated the observed data (e.g., due to the finite capacity of neural networks or lack of sufficient training data). To illustrate this, let us consider a pedagogical example involving a simple initial value problem 
\begin{align}
    \label{eq: antideriv}
    \frac{d s(x)}{d x}=u(x), \quad x \in [0, 1],
\end{align}
with an initial condition $s(0) = 0$. Here, our goal is to learn the anti-derivative operator 
\begin{align}
    G: u(x) \longrightarrow s(x)=s(0)+\int_{0}^{x} u(t) d t,  \quad  x \in[0,1].
\end{align}
To generate a training data-set, we randomly sample 10,000 different functions $u$ from a zero-mean Gaussian process prior with an exponential quadratic kernel using a length scale of $l=0.2$ \cite{rasmussen2006gaussian}. We also obtain the corresponding 10,000 ODE solutions $s$ by integrating the ODE \ref{eq: antideriv} using an explicit Runge-Kutta method (RK45) \cite{iserles2009first}. For each observed pair of $(u, s)$, we choose $m=100$ sensors $\{x_i\}_{i=1}^m$ uniformly distributed on the time interval $[0,1]$ and randomly select $P=1$ observations of $s(\cdot)$ in $[0,1]$. To generate the test data-set, we repeat the same procedure with $m=100$ and $P=100$. The final test data-set contains $1,000$ different samples of random input functions $u$.

We represent the operator $G$ using the unstacked DeepONet $G_{\bm{\theta}}$ where both the branch net and the trunk net are two-layer fully-connected neural networks with $100$ neurons per hidden layer. Each network is equipped with ReLU activation functions. The network parameters can be trained by minimizing the following loss
\begin{align}
    \label{eq: antideriv_loss_operator}
       \mathcal{L}(\bm{\theta}) &= \frac{1}{N} \sum_{i=1}^N  \left|G_{\bm{\theta}} (\bm{u}^{(i)})(y^{(i)})- s^{(i)}(y^{(i)})  \right|^2, 
\end{align}
where $\bm{u}^{(i)} = [u^{(i)}(x_1), u^{(i)}(x_2), \dots, u^{(i)}(x_m)]$ represent the input function, and $s^{(i)}(y^{(i)})$ denotes the associated solution of equation \ref{eq: antideriv} evaluated at $y^{(i)}$. 

\begin{figure}
    \centering
    \includegraphics[width= 0.7\textwidth]{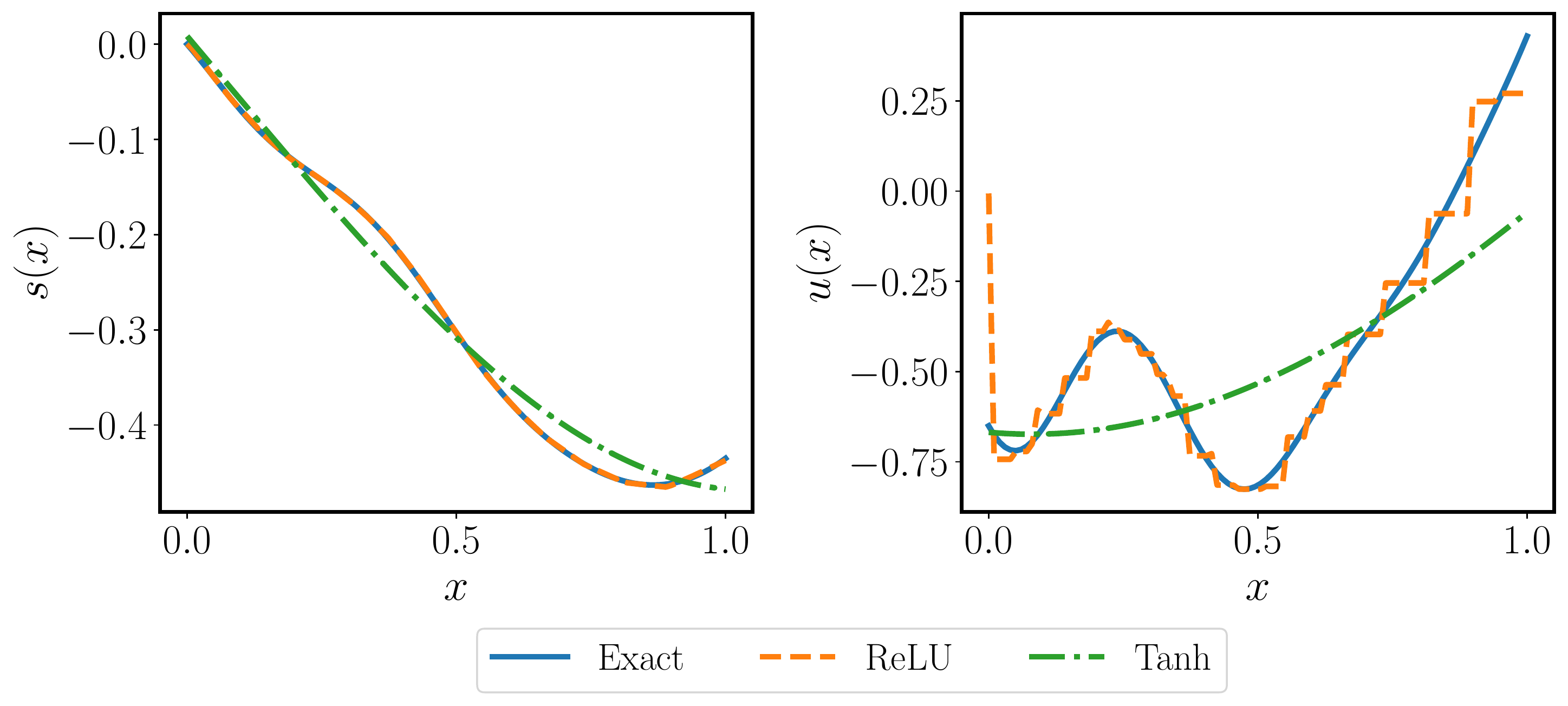}
    \caption{{\em Learning the anti-derivative operator:} Predicted solution $s(x)$ and residual $u(x)$ versus the ground truth for a representative input function. The results are obtained by training a conventional DeepONet model \cite{lu2019deeponet} equipped with different activation functions after 40,000 iterations of gradient descent using the Adam optimizer.}
    \label{fig: deeponet_antideriv_s_u}
\end{figure}

We train the DeepONet model by minimizing the above loss function via gradient descent using the Adam optimizer \cite{kingma2014adam}
for $40,000$ iterations. Note that the final output of DeepONet is a function of input coordinates $\bm{x}$. Thus, we can compute the residual $\frac{ds(x)}{dx}$ of the inferred solution $s(x)$ using automatic differentiation \cite{baydin2017automatic}, and compare the computed residual with $u(x)$
at the sensors $\{x_i\}_{i=1}^m$. Figure \ref{fig: deeponet_antideriv_s_u} shows the comparison of the predicted  $s(x)$ and $\frac{ds(x)}{dx}$ against the ground truth for one representative random sample from the test data-set. We can observe a good agreement between the predicted and the exact solution $s(x)$ when using ReLU activation functions. However, the predicted residual $\frac{ds(x)}{dx}$ seems to approximate $u(x)$ with step functions, which leads to a large approximation error.
One may postulate that this is due to the non-smoothness of ReLU activations. However, as shown in the same Figure \ref{fig: deeponet_antideriv_s_u}, similar poor predictions of both $u(x)$ and $s(x)$ are obtained by repeating the same process using a DeepONet equipped with tanh activations, under exactly the same hyper-parameter settings. Thus, despite the guarantee of universal approximation theorem  for operators \cite{chen1995universal}, it is possible that DeepONet models may not appropriately learn the correct solution operator in the sense that the predicted output functions are not compatible with the ground truth operator that generated the training data.

\section{Physics-informed DeepONets}
\label{sec: physics_informed_deeponet}

Physics-informed neural networks (PINNs) \cite{raissi2019physics} can seamlessly integrate the data measurements and physical governing laws by penalizing residuals of partial differential equations in the loss function of a neural network using automatic differentiation \cite{baydin2017automatic}. Motivated by PINNs and our findings in the previous section, we propose a novel model class referred to as "physics-informed" DeepONets that enables the DeepONet output functions to be consistent with physical constraints via minimizing the residual of the underlying governing laws in the same manner as PINNs. Specifically, we consider minimizing the following composite loss function
\begin{align}
    \label{eq: physics_informed_loss}
    \mathcal{L}(\bm{\theta}) = \mathcal{L}_{\text{operator}}(\bm{\theta}) +  \mathcal{L}_{\text{physics}}(\bm{\theta}),
\end{align}
where $ \mathcal{L}_{\text{operator}}(\bm{\theta})$ is defined exactly the same as in equation \ref{eq: loss_operator} and
\begin{align}
    \mathcal{L}_{\text{physics}}(\bm{\theta}) &= \frac{1}{NQm} \sum_{i=1}^N \sum_{j=1}^Q \sum_{k=1}^m \left|\mathcal{N}(u^{(i)}(\bm{x}_k), G_{\bm{\theta}}(\bm{u}^{(i)})(\bm{y}^{(i)}_j) ) \right|^2,
\end{align}
where $\{\bm{y}^{(i)}_j\}_{i=1}^Q$ denotes a set of collocation points that are randomly sampled from the domain of $G(\bm{u}^{(i)})$, and used to approximately enforce a set of given physical constraints, typically described by systems of PDEs.

To demonstrate the effectiveness of the proposed framework, let us revisit the numerical example shown in section \ref{sec: deeponet} and use the proposed physics-informed DeepONet architecture to learn the anti-derivate operator.  Specifically, we represent the operator $G$ by a DeepONet where both branch net and the trunk are 5-layer fully-connected neural network with 50 units per hidden layer. We also equip both networks with tanh activation functions.
For this particular example, $\mathcal{L}_{\text{operator}}$ is exactly the same as in equation \ref{eq: antideriv_loss_operator},  while the physics loss can be formulated as
\begin{align}
    \mathcal{L}_{\text{physics}}(\bm{\theta}) = \frac{1}{Nm} \sum_{i=1}^N \sum_{j=1}^m \left| \frac{dG_{\bm{\theta}}(\bm{u}^{(i)})(y)}{dy}\Big|_{y = x_j} - u^{(i)}(x_j)   \right|^2. 
\end{align}
In this example we consider $Q=m$ and $y^{(i)}_j = x_{j}$ for $j=1,2,\dots, Q$.

\begin{figure}
    \centering
    \includegraphics[width= 0.6\textwidth]{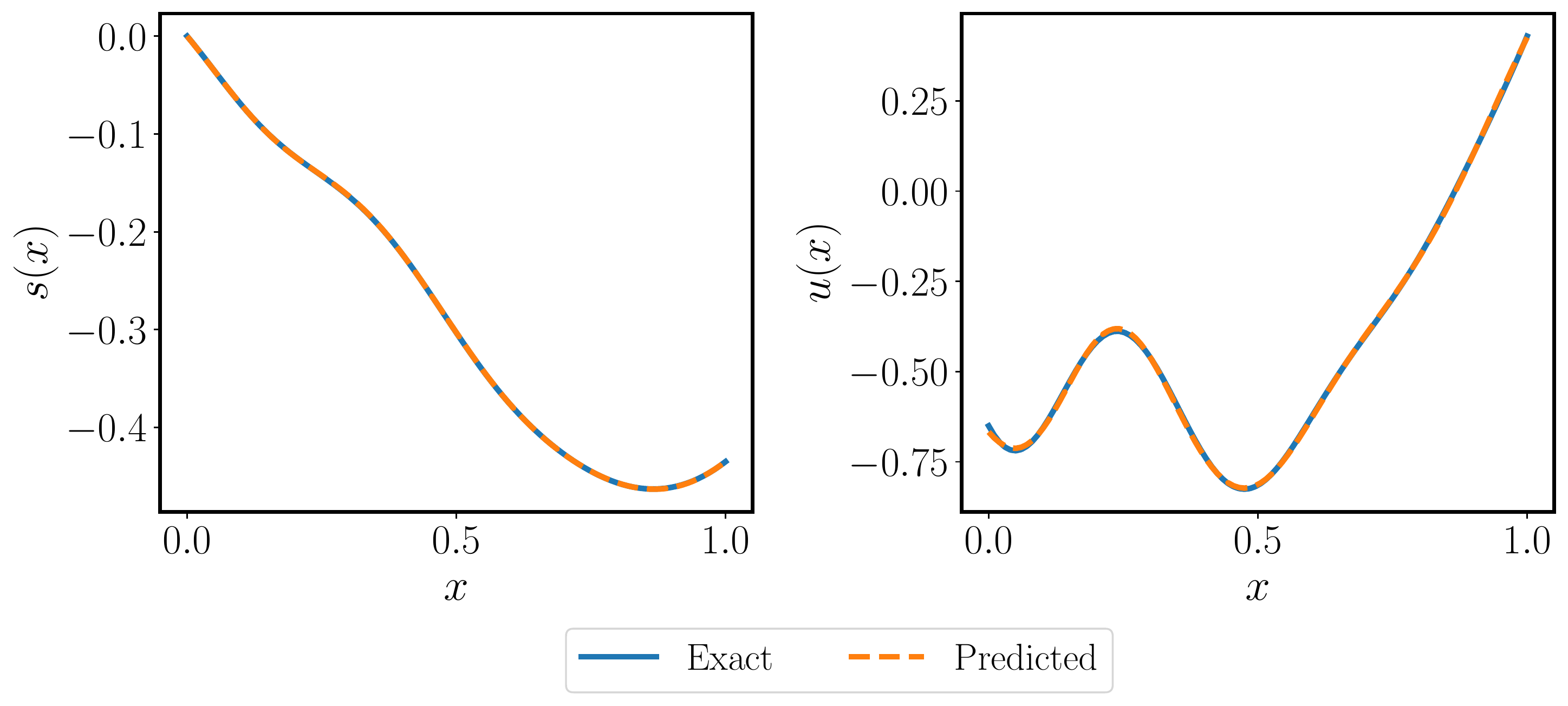}
    \caption{{\em Learning anti-derivative operator:} Exact solution and residual versus the predictions of a trained physics-informed DeepONet for the same input function as in Figure \ref{fig: deeponet_antideriv_s_u}.}
    \label{fig: physical_deeponet_antideriv_s_u}
\end{figure}

\begin{figure}
    \centering
    \includegraphics[width= 0.4\textwidth]{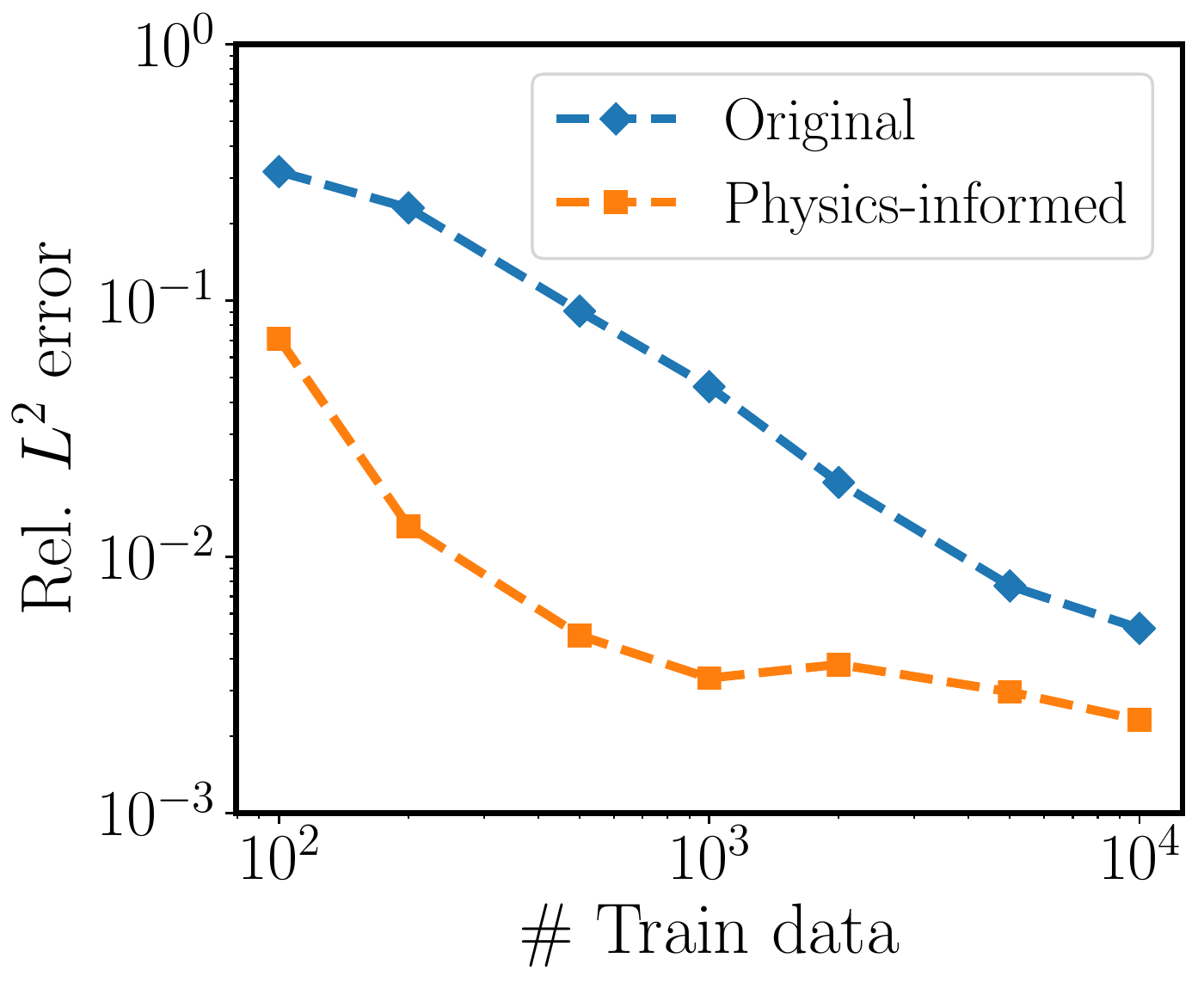}
    \caption{{\em Learning anti-derivative operator:} Mean of the relative $L^2$ prediction error of the original DeepONet \cite{lu2019deeponet} and the physics-informed DeepONet as a function of the number of $u$ samples.}
    \label{fig: antideriv_diff_train_data}
\end{figure}

Figure  \ref{fig: physical_deeponet_antideriv_s_u} presents the predicted $s(x)$ and $\frac{ds(x)}{dx}$ for the same random sample (see Figure \ref{fig: deeponet_antideriv_s_u}) by minimizing the loss function \ref{eq: physics_informed_loss} for 40,000 iterations of gradient descent using the Adam optimizer. Evidently, both predictions achieve an excellent agreement with the corresponding reference solutions. This can be further verified by the mean of relative $L^2$ error of the model predictions reported table \ref{tab: antideriv_l2_error}, from which we may conclude that the physics-informed DeepONet not only attains comparable accuracy to the original DeepONet, but also satisfies the underlying ODE constraint.  Another crucial finding is that physics-informed DeepONets are data-efficient and therefore effective in small data regime. To illustrate this, we train both a DeepONet and a  physics-informed DeepONet for different number of training data points (i.e, different number of samples $u$) and report the mean of the relative $L^2$ error of $s(x)$  over 1,000 realizations from the test data-set in Figure \ref{fig: antideriv_diff_train_data}. We  observe that conventional DeepONets require more than 10x training data to achieve the same accuracy as their physics-informed counterpart.

The remarkable success of DeepONets is based on the assumption that there exists enough data to train the model offline. However, high-fidelity numerical simulations are often computationally expensive and the volume of useful experimental data is generally limited or even intractable for many practical scenarios. Strikingly, as will be demonstrated below, the proposed physics-informed DeepONet is capable of learning the solution operator of parametric PDEs even without any paired input-output  data, except for appropriate initial or boundary conditions -- a setting for which  conventional DeepOnets \cite{lu2019deeponet} and other competing approaches for operator learning in Banach spaces \cite{cai2020deepm, lin2020operator, liu2021learning} fall short.



\begin{table}
\renewcommand{\arraystretch}{1.4}
    \centering
    \begin{tabular}{|c|c|c|}
\hline
 \diagbox{Model}{Relative $L^2$ error}  &  Relative $L^2$ error of $s$ & Relative $L^2$ error of $u$  \\ \hline
  DeepONet (ReLU)        &    $5.16e-03 \pm  4.58e-03$          &     $1.39e-01 \pm 5.58e-02$    \\ \hline
     DeepONet (Tanh)                    &         $1.89e-01 \pm 1.51e-01$        &  $6.14e-01 \pm 2.36e-01$           \\ \hline
Physics-informed DeepONet (Tanh)              &     $2.49e-03 \pm 2.74e-03$  &    $6.29e-03 \pm 3.65e-03$                           \\ \hline
\end{tabular}
    \caption{{\em Learning anti-derivative operator:} Mean and standard deviation of relative $L^2$ prediction errors of DeepONet and physics-informed DeepONet equipped with ReLU or Tanh activations over 1,000 examples in the test data-set.}
    \label{tab: antideriv_l2_error}
\end{table}





\section{Numerical results}\label{sec:results}

To demonstrate the effectiveness of physics-informed DeepONets, we   provide a series of comprehensive  numerical studies for solving various types of parametric PDEs. In most examples, we model random input functions $\bm{u}(\bm{x})$ using mean-zero Gaussian random fields (GRF) \cite{rasmussen2006gaussian} as
\begin{align*}
    \bm{u}(\bm{x}) \sim \mathcal{G} \mathcal{P}\left(0, k_{l}\left(\bm{x}_{1}, \bm{x}_{2}\right)\right),
\end{align*}
with an exponential quadratic covariance kernel $k_{l}\left(x_{1}, x_{2}\right)=\exp \left(-\left\|x_{1}-x_{2}\right\|^{2} / 2 l^{2}\right)$ with a length scale parameter $l >0$. The parameter $l$ will be used to control the complexity of the sampled input functions, and in general larger $l>0$ leads to smoother $\bm{u}$.

Throughout all benchmarks we will employ hyperbolic tangent activation functions (Tanh) and initialize the DeepOnet networks using the Glorot normal scheme \cite{glorot2010understanding}, unless otherwise stated. All networks are trained via  mini-batch stochastic gradient descent using the Adam optimizer \cite{kingma2014adam} with default settings.  Particularly, we set the batch size to be 10,000 and use exponential learning rate decay with a decay-rate of 0.9 every 1, 000 training iterations. 
The detailed hyper-parameters and computational cost for all examples are listed in Appendix \ref{sec: parameters} and \ref{sec: computational_cost}. The code and data accompanying this manuscript is publicly available \url{https://github.com/PredictiveIntelligenceLab/Physics-informed-DeepONets}.



\subsection{The anti-derivative operator}
\label{sec: Antiderivative}

To illustrate the capability of physics-informed DeepONets in solving parametric differential equations, let us again consider a pedagogical example involving the anti-derivative operator \ref{eq: antideriv} as discussed in section \ref{sec: deeponet}, \ref{sec: physics_informed_deeponet}, i.e.
\begin{align}
    &\frac{d s(x)}{d x}=u(x), \quad x \in [0, 1],\\
    &s(0) = 0.
\end{align}
Assuming that $u(x)$ is exactly known, we aim to learn the solution operator from $u(x)$ to the solution $s(x)$ without any paired input-output data. To this end, we represent the operator by a DeepONet $G_{\bm{\theta}}$ where both branch net and trunk net are 5-layer fully-connected neural networks with 50 neurons per hidden layer and equipped with tanh activations. The corresponding loss function is expressed as
\begin{align}
    \mathcal{L}(\bm{\theta}) &= \mathcal{L}_{\text{operator}}(\bm{\theta}) +  \mathcal{L}_{\text{physics}}(\bm{\theta}) \\
    &= \frac{1}{N} \sum_{i=1}^N  \left|G_{\bm{\theta}} (\bm{u}^{(i)})(0) \right|^2  + \frac{1}{NQ} \sum_{i=1}^N \sum_{j=1}^Q \left| \frac{dG_{\bm{\theta}}(\bm{u}^{(i)})(y)}{dy}\Big|_{y = x_j} - u^{(i)}(x_j)   \right|^2. 
\end{align}
Here, $\bm{u}^{(i)} = [u^{(i)}(x_1), u^{(i)}(x_2), \dots, u^{(i)}(x_m)]$, and we sample $N=10,000$ input functions $u(x)$ from a GRF with length scale $l=0.2$. Moreover, we take $Q = m = 100$ and
$\{x_j\}_{j=1}^Q$ are equi-spaced grid points in $[0,1]$. 
From the expression of the loss function, it is worth emphasizing  that all "training data" comes the measurements of $u(x)$, and the zero initial condition on $s(0)$ (i.e. no other observations of $s(x)$ are available). The test data-set is the same as discussed in section \ref{sec: deeponet}.

\begin{figure}
    \centering
    \includegraphics[width=0.8\textwidth]{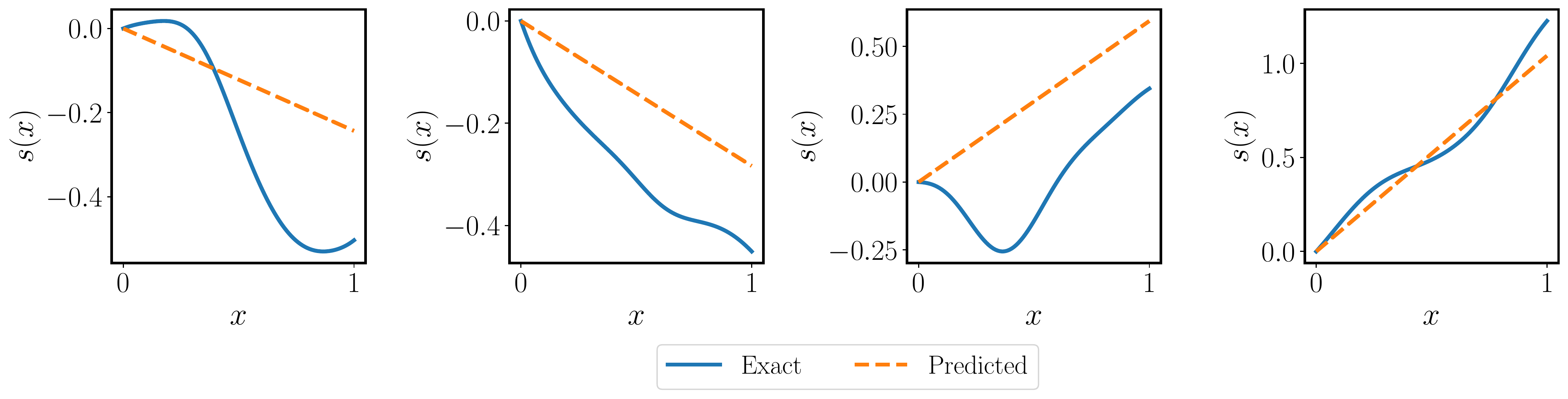}
    \caption{{\em Solving a 1D parametric ODE:} Exact solutions versus the predicted solutions of a trained DeepONet for four different input samples. We observe that the conventional DeepONet \cite{lu2019deeponet} learns a degenerate operator.}
    \label{fig: deeponet_ODE_relu_examples}
\end{figure}

We train the physics-informed DeepONet by minimizing the above loss function for 40,000 iterations via gradient descent using the Adam optimizer. Results for one representative input sample from the test data-set are presented in Figure \ref{fig: physical_deeponet_ODE_s_u}. Additional visualizations for different input samples are provided in Appendix Figure \ref{fig: physical_deeponet_ODE_s_u_examples}. As it can be seen, an excellent agreement can be achieved between the physics-informed DeepONet predictions and the ground truth. Furthermore, we investigate the performance of the original DeepONet \cite{lu2019deeponet} in solving this parametric ODE example. To this end,  we train a DeepONet by minimizing the loss function $\mathcal{L}_{\text{operator}}(\bm{\theta})$ under exactly the same hyper-parameter setting. Representative predicted solutions $s(x)$  for different input samples $u$ are shown in Figure \ref{fig: deeponet_ODE_relu_examples}. We observe that the conventional DeepONet learns a degenerate map that can fit the initial condition $s(0) = 0$, but returns erroneous predictions for all $x>0$. These observations can be further quantified in Table \ref{tab: ODE_l2_error}, which reports the mean and standard deviation of  the relative $L^2$ prediction error for the output functions $s$ and their corresponding ODE residual $u$ over 1,000 examples in the test data-set. Remarkably, the proposed physics-informed DeepONet is trained in the absence of any paired input-output data, but still obtains comparable accuracy to the results shown in Table \ref{tab: antideriv_l2_error}, where the model is trained with a large amount of paired input-output observations.

\begin{table}
\renewcommand{\arraystretch}{1.4}
    \centering
    \begin{tabular}{|c|c|c|}
\hline
 \diagbox{Model}{Relative $L^2$ error}  &  Relative $L^2$ error of $s$ & Relative $L^2$ error of $u$  \\ \hline
  DeepONet        &    $8.80e-01 \pm  4.72e-01$          &     $9.15e-01 \pm 1.86e-01$    \\ \hline
Physics-informed DeepONet               &     $3.25e-03 \pm 3.19e-03$  &    $6.97e-03 \pm 3.95e-03$   \\ \hline
\end{tabular}
    \caption{{Solving a 1D parametric ODE:} Mean and standard deviation of the relative $L^2$ prediction errors of a trained DeepONet and physics-informed DeepONet model over 1,000 examples  in the test data-set.}
    \label{tab: ODE_l2_error}
\end{table}

\begin{figure}
     \centering
     \begin{subfigure}[b]{0.6\textwidth}
         \centering
         \includegraphics[width=\textwidth]{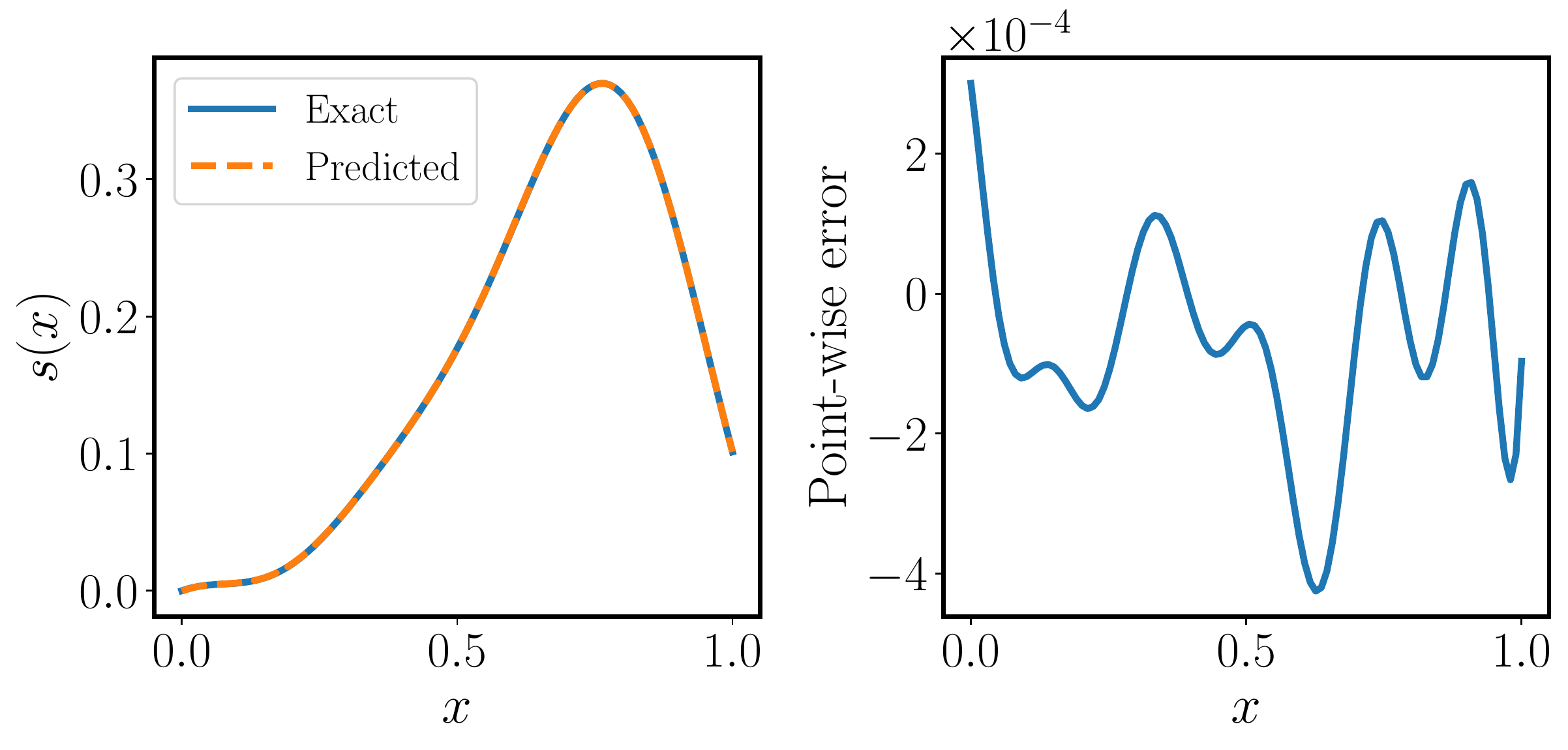}
         \caption{ }
         \label{fig: physical_deeponet_ODE_s}
     \end{subfigure}
     \begin{subfigure}[b]{0.6\textwidth}
         \centering
         \includegraphics[width=\textwidth]{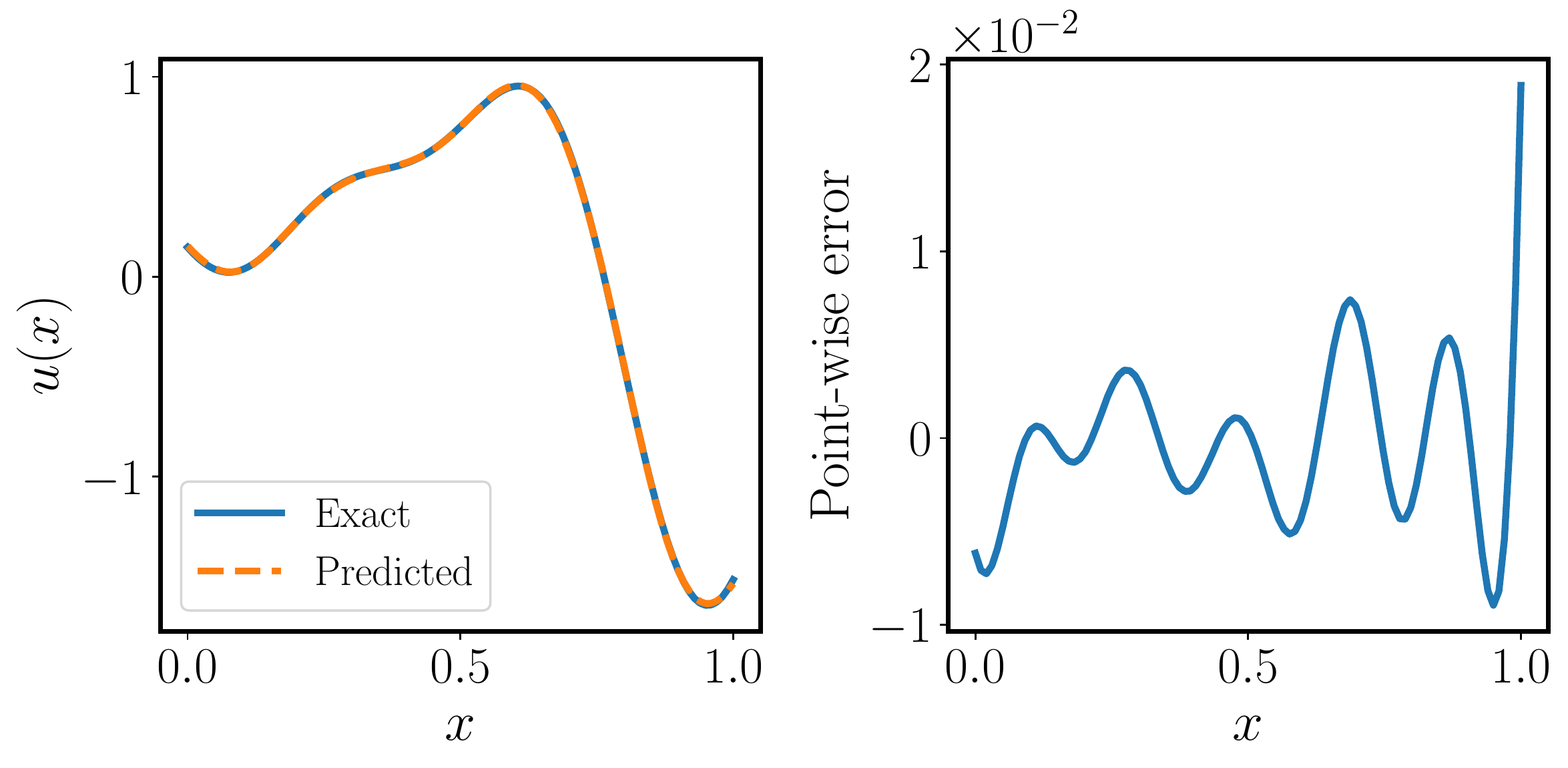}
         \caption{}
         \label{fig: physical_deeponet_ODE_u}
     \end{subfigure}
       \caption{{\em Solving a 1D parametric ODE:} (a)(b) Exact solution and residual versus the predictions of
       a trained physics-informed DeepONet for a representative input sample.}
        \label{fig: physical_deeponet_ODE_s_u}
\end{figure}

More impressively, below, we show that physics-informed DeepONets can accommodate extremely irregular input functions by using appropriate trunk network architectures. To illustrate this, we consider a GRF with a length scale $l=0.01$ as a prior on the input function space. We take $Q = m = 200$ and repeat the same data generation procedure as before. In this example, the training data-set contains $N = 10,000$ different $u$ samples, while the test data-set contains 1,000 realizations. 

Given that the input functions are sampled from a GRF with a relatively small length scale,  the associated solutions are expected to exhibit high frequencies. Therefore, we represent the solution operator by a DeepONet with Fourier feature embeddings \cite{tancik2020fourier}, which are able to learn high-frequency components more effectively. Generally,
A random Fourier mapping $\gamma$ is defined as
\begin{align}
    \gamma(\bm{v})= \begin{bmatrix}
    \cos (\bm{B v} ) \\
    \sin (\bm{Bv} )
    \end{bmatrix},
\end{align}
where each entry in $\bm{B} \in \R^{m \times d}$
is sampled from a Gaussian distribution $\mathcal{N}(0, \sigma^2)$ and $\sigma > 0$ is a user-specified hyper-parameter. 
Then, a Fourier feature network \cite{tancik2020fourier} can be simply constructed using a random Fourier features mapping $\gamma$ as a coordinate embedding of the inputs, followed by a conventional fully-connected neural network.

In particular, we encode the input functions by a branch net that is a 5-layer fully-connected neural network with 200 neurons per hidden layer. In addition,  we apply a Fourier feature embedding \cite{tancik2020fourier} initialized with $\sigma = 50$ to the input coordinates $\bm{y}$ before passing the embedded inputs through a trunk network with the same architecture as the branch net. Then we train the physics-informed DeepONet for 300,000 iterations of gradient descent using the Adam optimizer. As can be seen in Figure \ref{fig: physical_deeponet_ODE_s_u}, the predicted solutions $s(x)$ and their corresponding ODE residuals $u(x)$ obtained by the  physics-informed DeepONet with Fourier feature networks are in excellent agreement with the growth truth. Moreover, the results of training the same physics-informed DeepONet without Fourier feature embeddings are presented in Appendix Figure \ref{fig: physical_deeponet_MLP_s_u_examples}. One may observe that using a conventional fully-connected trunk networks cannot accurately capture the high-frequency oscillations, leading to a large prediction error. These observations can be further quantified in  Table \ref{tab: irregular_ODE_l2_error},  which summarizes the mean and standard derivation of the relative $L^2$ prediction error of trained physics-informed DeepONets constructed with different network architectures. Although here we have illustrated that an appropriate network architecture plays a prominent role in the performance of DeepONets, a comprehensive investigation of DeepONet architectures is beyond the scope of the present study and will be investigated in  future work.

It is also worth pointing out that the trained physics-informed DeepONets is even capable of yielding accurate predictions for out-of-distribution test data. To illustrate this, we create a test data-set by sampling input functions from a GRF with a larger length-scale of $l=0.2$ (recall that the training data for this case is generated using $l=0.01$). The corresponding relative $L^2$ prediction error averaged over $1,000$ test examples is measured as $7.12e-03$. Some visualizations of the model predictions for this out-of-distribution prediction task  are shown in the Appendix, Figure \ref{fig: physical_deeponet_FF_02_s_u_examples}.  

\begin{figure}
     \centering
     \begin{subfigure}[b]{0.6\textwidth}
         \centering
         \includegraphics[width=\textwidth]{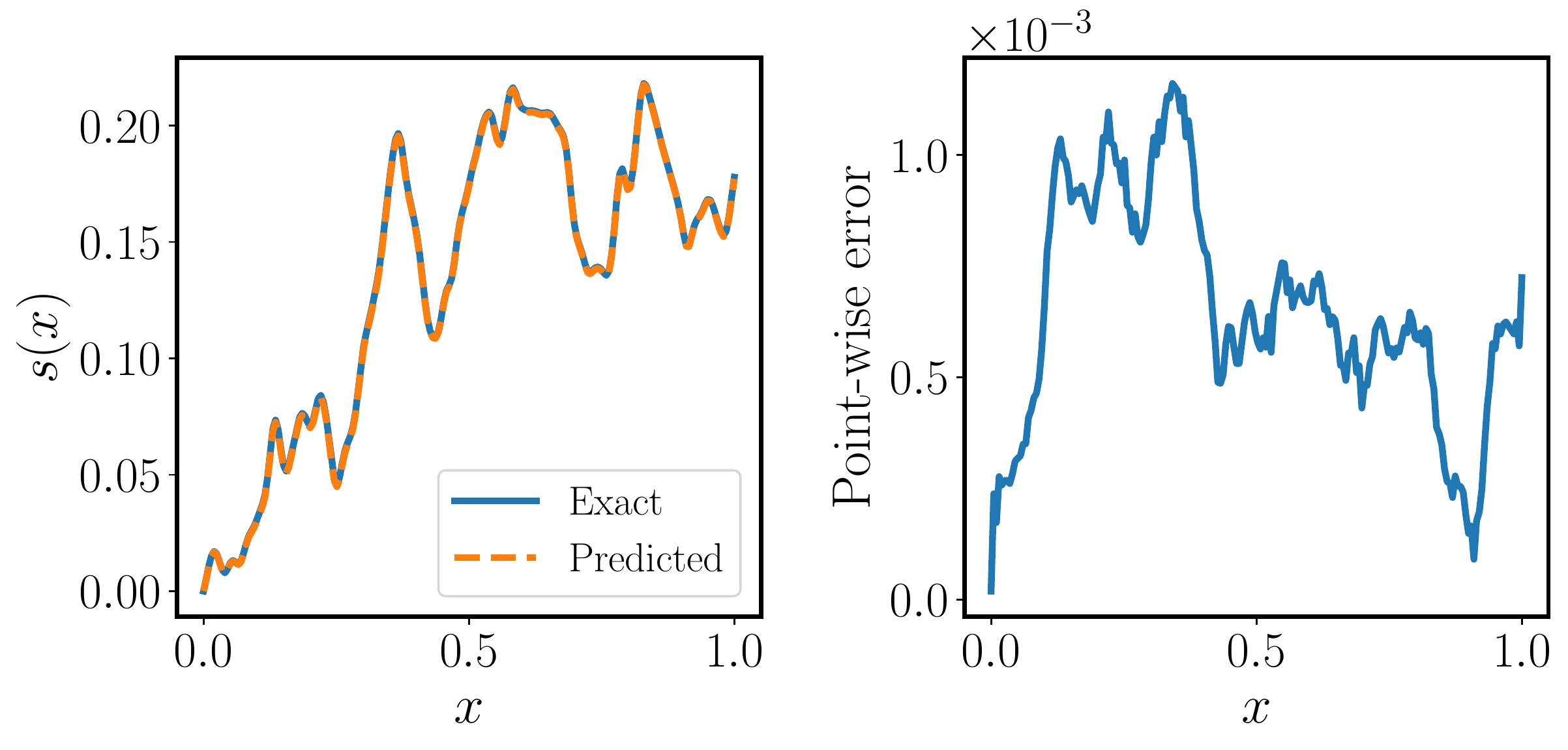}
         \caption{ }
         \label{fig: physical_deeponet_FF_ODE_s}
     \end{subfigure}
     \begin{subfigure}[b]{0.6\textwidth}
         \centering
         \includegraphics[width=\textwidth]{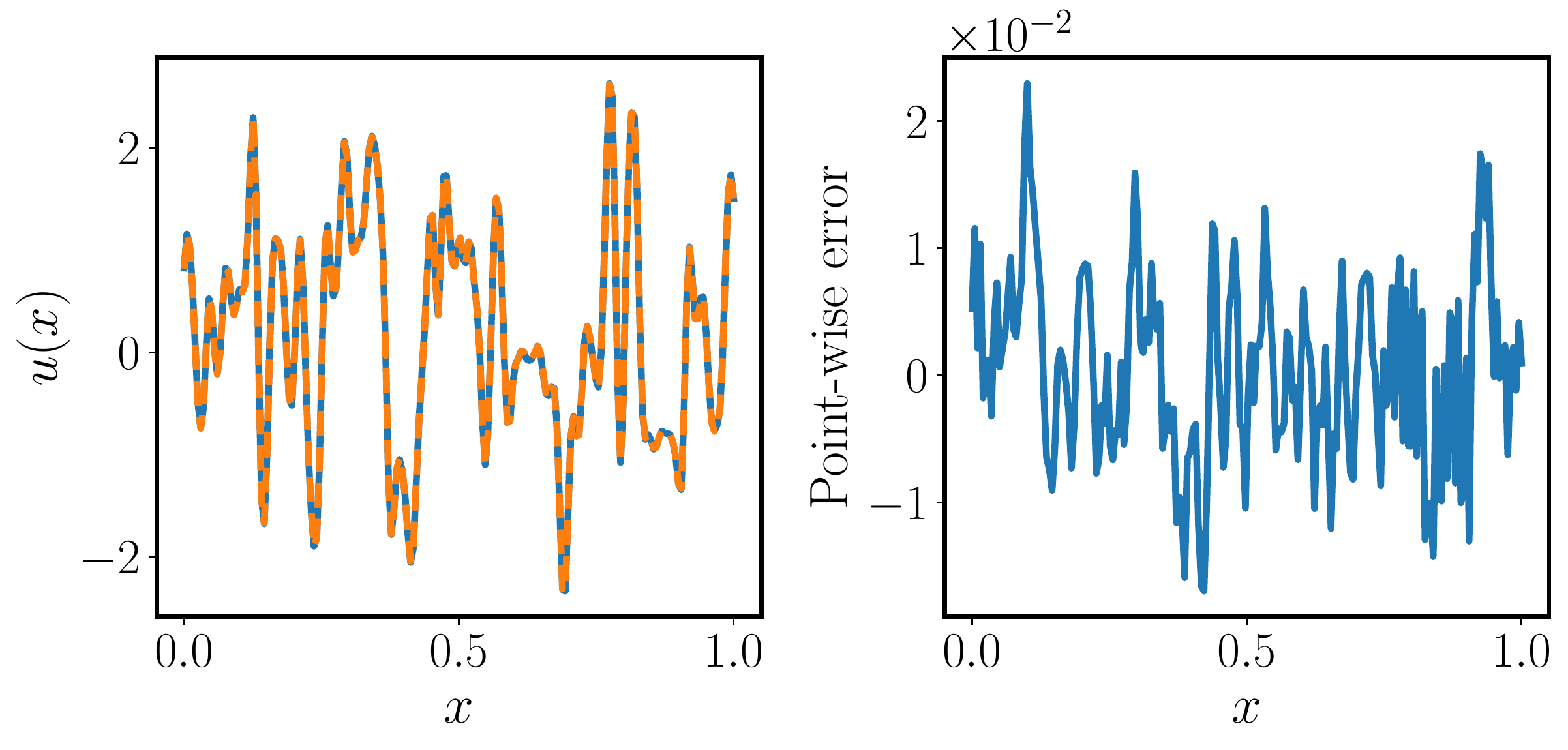}
         \caption{}
         \label{fig: physical_deeponet_FF_ODE_u}
     \end{subfigure}
         \caption{{\em Solving a 1D parametric ODE with irregular input functions:} (a)(b) Exact solutions and corresponding ODE residuals versus the predictions of a trained physics-informed DeepONet with Fourier feature embeddings (for a representative input function sampled from a GRF with length scale $l=0.01$). Additional visualizations are provided in Appendix Figure \ref{fig: physical_deeponet_FF_s_u_examples}.}
        \label{fig: physical_deeponet_FF_ODE_s_u}
\end{figure}

\begin{table}
\renewcommand{\arraystretch}{1.4}
    \centering
    \begin{tabular}{|c|c|c|}
\hline
 \diagbox{Architecture}{Relative $L^2$ error}  &  Relative $L^2$ error of $s$ & Relative $L^2$ error of $u$  \\ \hline
  Fully-connected network        &    $3.48e-1 \pm  2.34e-1$          &     $6.81e-1 \pm 6.31e-2$    \\ \hline
Fourier feature network             &     $8.45e-3\pm 6.65e-3$  &    $8.25e-3 \pm 1.54e-3$   \\ \hline
\end{tabular}
    \caption{{\em Solving a 1D parametric ODE with irregular input functions:} Mean and standard deviation of the relative  $L^2$ prediction errors of physics-informed DeepONet represented by different network architectures over 1,000 examples in the test data-set.}
    \label{tab: irregular_ODE_l2_error}
\end{table}

\subsection{Diffusion-reaction systems}
\label{sec: DR}

Our next example involves an implicit operator described by a nonlinear diffusion-reaction PDE with a source term $u(x)$
\begin{align}
    \frac{\partial s}{\partial t}=D \frac{\partial^{2} s}{\partial x^{2}}+k s^{2}+u(x), \quad (x,t) \in (0,1] \times (0,1],
\end{align}
with the zero initial and boundary conditions, where $D=0.01$ is the diffusion coefficient and $k=0.01$ is the reaction rate. We aim to learn the solution operator for mapping source terms $u(x)$ to the corresponding PDE solutions $s(x)$ using a physics-informed DeepONet. 

We approximate the operator by a physics-informed DeepONet architecture $G_{\bm{\theta}}$, where the branch and trunk networks are two separate 5-layer fully-connected neural networks with 50 neurons per hidden layer. For a given input function $\bm{u}^{(i)}$,  we define the corresponding PDE residual as
\begin{align}
    R_{\bm{\theta}}^{(i)}(x, t) = \frac{d G_{\bm{\theta}}(\bm{u}^{(i)})(x,t)}{dt} - D\frac{d^2 G_{\bm{\theta}}(\bm{u}^{(i)})(x,t)}{dx^2} - k [G_{\bm{\theta}}(\bm{u}^{(i)})(x,t)]^2,
\end{align}
where  $\{\bm{u}^{(i)}\}_{i=1}^N = 
\{[u^{(i)}(x_1), u^{(i)}(x_2), \dots, u^{(i)}(x_m)]\}_{i=1}^N$ represents the input functions, and $\{x_i\}_{i=1}^m$ is a collection of equi-spaced sensor locations in $[0,1]$.
The parameters of the physics-informed DeepONet can be trained by minimizing the loss function 
\begin{align}
    \label{eq: DR_loss}
    \mathcal{L}(\bm{\theta}) &= \mathcal{L}_{\text{operator}}(\bm{\theta}) + \mathcal{L}_{\text{physics}}(\bm{\theta}) \\
                             &= \frac{1}{NP} \sum_{i=1}^N \sum_{j=1}^P \left|G_{\bm{\theta}} (\bm{u}^{(i)})(x^{(i)}_{u,j}, t^{(i)}_{u,j}) \right|^2  + \frac{1}{NQ} \sum_{i=1}^N \sum_{j=1}^Q \left|  R_{\bm{\theta}}^{(i)}(x_{r,j}^{(i)}, t^{(i)}_{r,j})  - u^{(i)}(x_{r, j}^{(i)})   \right|^2. 
\end{align}
Here, for each $\bm{u}^{(i)}$,
$\{(x^{(i)}_{u,j}, t^{(i)}_{u,j}\}_{j=1}^P$ are uniformly sampled points from the boundary of $[0,1] \times [0,1]$ (excluding $t=1$), while $\{(x_{r,j}^{(i)}, t^{(i)}_{r,j})\}_{j=1}^Q$ is a set of collocation points satisfying $x^{(i)}_{r,j} = x_j$, and $\{ t^{(i)}_{r,j})\}_{j=1}^Q$ are uniformly sampled in $[0,1]$. Consequently, $\mathcal{L}_{\text{operator}}(\bm{\theta})$ enforces the zero initial and boundary conditions, and $\mathcal{L}_{\text{physics}}(\bm{\theta})$ penalizes the parametric PDE residual at the $Q$ collocation  points.  In this example, we set $P = Q = 100$ and randomly sample $N = 10,000$ input functions $u(x)$ from a GRF with length scale $l=0.2$. 
To generate the test data-set, we sample $N = 1,000$ input functions $u(x)$ from the same GRF and solve the diffusion-reaction system using a second-order implicit finite difference method on a $100 \times 100$ equi-spaced grid \cite{iserles2009first}. Hence, the test data-set will contain $1,000$ realizations evaluated on a $100 \times 100$ uniform grid.

\begin{figure}
    \centering
    \includegraphics[width=0.8\textwidth]{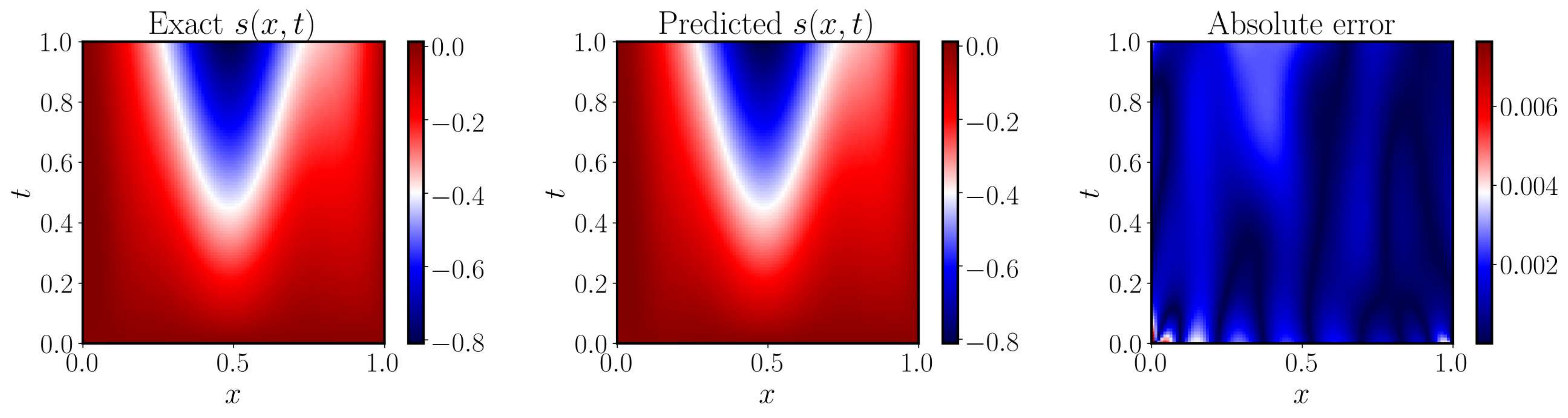}
   \caption{{\em Solving a parametric diffusion-reaction system:} Exact solution versus the prediction of a trained physics-informed DeepONet for a representative example in the test data-set.}
    \label{fig: physical_deeponet_DR_1}
\end{figure}

\begin{figure}
    \centering
    \includegraphics[width=0.4\textwidth]{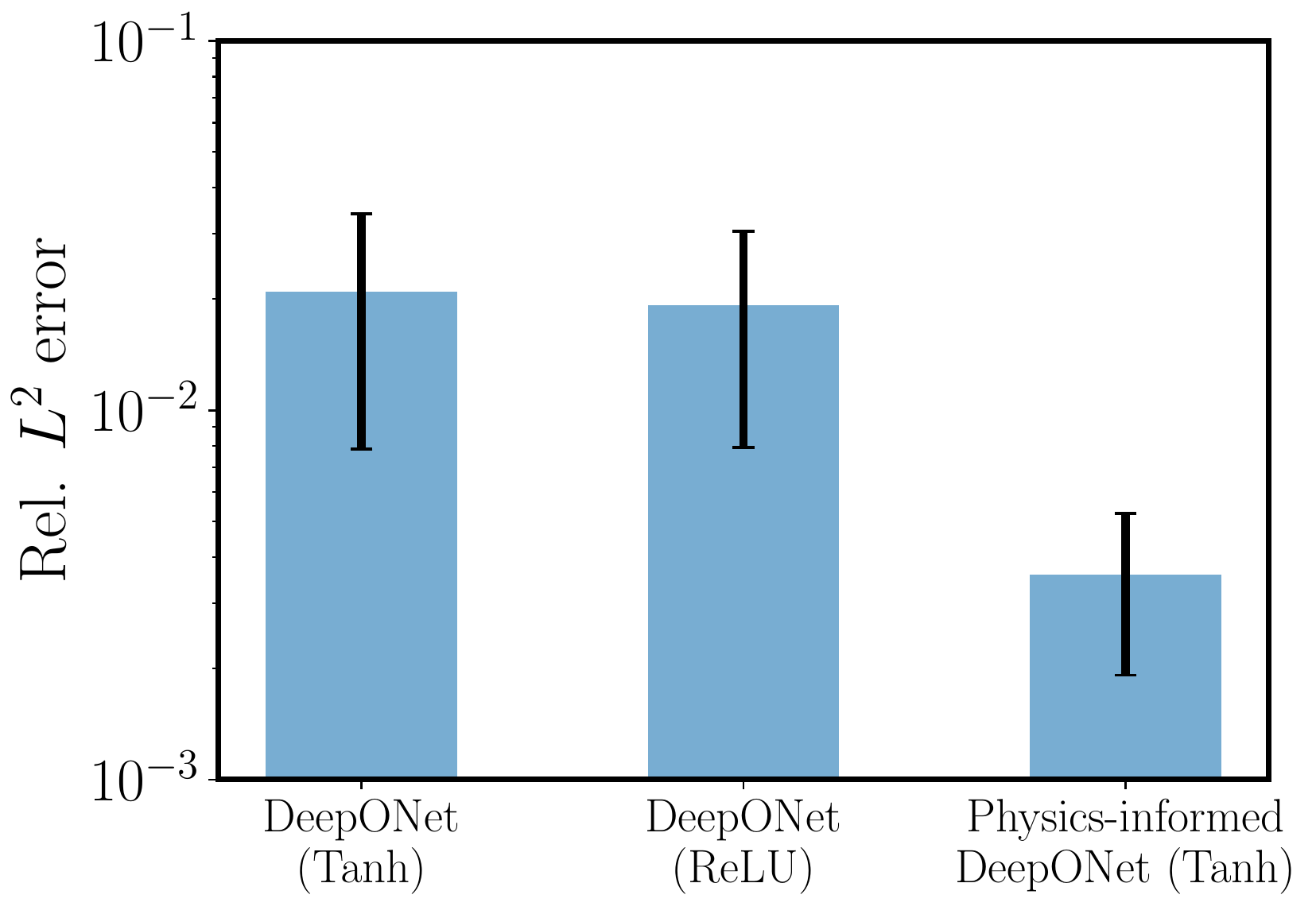}
    \caption{{\em Solving a parametric diffusion-reaction system:} Mean and standard deviation of the relative $L^2$ prediction error of a trained DeepONet {\em (with paired input-output training data)} and a  physics-informed DeepONet {\em (without paired input-output training data)} over $1,000$ examples in the test data-set. The mean and standard deviation of the relative $L^2$ prediction are $\sim 1.92\% \pm 1.12\%$ (DeepOnet) and $ \sim 0.45\% \pm 0.16\%$ (physics-informed DeepOnet), respectively. 
    Remarkably, the physics-informed DeepONet yields a $\sim 80\%$ improvement in prediction accuracy with a 100\% reduction in the data-set size required for training.}
    \label{fig: DR_l2_error}
\end{figure}

We train the physics-informed DeepONet by minimizing the loss function \ref{eq: DR_loss} for $120,000$ iterations of gradient descent using the Adam optimizer with default settings. Figure \ref{fig: physical_deeponet_DR_1} shows the comparison between the predicted and the exact solution for a random test input sample.  More visualizations for different input samples can be found in Appendix Figure \ref{fig: physical_deeponet_DR_2}. We observe that the physics-informed DeepONet predictions achieve an excellent agreement with the corresponding reference solutions. Furthermore, we investigate the performance of a conventional DeepONet model \cite{lu2019deeponet} in the case where some training data are available. Specifically, we still use the same $10,000$ input functions sampled before and for each $u$, and we randomly select $P=100$ solution measurements out of the associated reference numerical solutions on the $100\times 100$ grid. Then, we train the conventional DeepONet model under exactly the same hyper-parameter settings. The mean and standard deviation of relative $L^2$ errors of trained DeepONet and physics-informed DeepONet over the test data-set are visualized in Figure \ref{fig: DR_l2_error}. 
The average relative $L^2$ error of DeepONet and physics-informed DeepONet are $\sim 1.92\%$ and $\sim 0.45\%$, respectively. 
Remarkably, in contrast to the conventional DeepONet that is trained on paired input-output measurements, the proposed physics-informed DeepONet can yield much more accurate predictions even without any paired training data (except for the initial and boundary conditions).  In our experience, predictive accuracy can be generally improved by using a larger batch size during training. A study of the effect of batch size for training physics-informed DeepONets can be found in Appendix \ref{appendix: DR_batch_size}.

\subsection{Burgers' equation}
\label{sec: Burger}

To highlight the ability of the proposed framework to handle  nonlinearity in the governing PDEs, we consider the 1D Burgers' equation benchmark investigated in Li {\em et. al.} \cite{li2020fourier}
\begin{align}
    \label{eq: Burger_eq}
    &\frac{ds}{dt} + s \frac{ds}{dx} - \nu \frac{d^2 s}{dx^2} = 0, \quad (x, t) \in (0,1) \times (0,1], \\
    &s(x,0) = u(x), \quad x \in (0,1),
\end{align}
with periodic boundary conditions
\begin{align}
    & s(0, t) = s(1,t), \\
    & \frac{ds}{dx}(0,t) = \frac{ds}{dx}(1,t),
\end{align}
where $t \in (0,1)$, the viscosity is set to $\nu = 0.01$, and the initial condition $u(x)$ is generated from a GRF $\sim \mathcal{N}\left(0,  25^2(-\Delta+5^2 I)^{-4}\right)$ satisfying the periodic boundary conditions. 
Our goal here is to use the proposed physics-informed DeepONet model to learn the solution operator mapping initial conditions $u(x)$ to the full spatio-temporal solution $s(x,t)$ of the 1D Burgers' equation. 

Suppose that the solution operator is approximated by a physics-informed DeepONet $G_{\bm{\theta}}$. For a specific input function $\bm{u}^{(i)}$, the PDE residual is defined by
\begin{align}
    R_{\bm{\theta}}^{(i)}(x, t) = \frac{d G_{\bm{\theta}}(\bm{u}^{(i)})(x,t)}{dt} + G_{\bm{\theta}}(\bm{u}^{(i)})(x,t) \frac{d G_{\bm{\theta}}(\bm{u}^{(i)})(x,t)}{dx} 
    - \nu \frac{d^2 G_{\bm{\theta}}(\bm{u}^{(i)})(x,t)}{dx^2},
\end{align}
where $\bm{u}^{(i)}$ denotes the input function evaluated a collection of fixed sensors $\{x_i\}_{i=1}^m$ that are uniformly spaced  in $[0, 1]$. 
Then, the physics-informed loss function is given by
\begin{align}
    \label{eq: Burger_loss}
    \mathcal{L}(\bm{\theta}) = \mathcal{L}_{\text{IC}}(\bm{\theta}) + \mathcal{L}_{\text{BC}}(\bm{\theta}) + \mathcal{L}_{\text{physics}}(\bm{\theta}),
\end{align}
where 
\begin{align}
     \mathcal{L}_{\text{IC}}(\bm{\theta}) &= \frac{1}{NP}\sum_{i=1}^N \sum_{j=1}^{P} \left|G_{\bm{\theta}}(\bm{u}^{(i)})(x^{(i)}_{ic, j}, 0 ) - u^{(i)}(x^{(i)}_{ic, j})  \right|^2 \\
      \mathcal{L}_{\text{BC}}(\bm{\theta}) &= \frac{1}{NP}\sum_{i=1}^N \sum_{j=1}^{P} 
     \left|G_{\bm{\theta}}(\bm{u}^{(i)})(0, t^{(i)}_{bc,j} ) - G_{\bm{\theta}}(\bm{u}^{(i)})(1, t^{(i)}_{bc,j} )\right|^2 
    \\
    &+ \frac{1}{NP}\sum_{i=1}^N \sum_{j=1}^{P} \left|\frac{d G_{\bm{\theta}}(\bm{u}^{(i)})(x,t) }{dx}\bigg|_{(0, t^{(i)}_{bc,j})} - \frac{d G_{\bm{\theta}}(\bm{u}^{(i)})(x,t) }{dx}\bigg|_{(1, t^{(i)}_{bc,j})} \right|^2 \\
    \mathcal{L}_{\text{physics}}(\bm{\theta}) &= \frac{1}{NQ}\sum_{i=1}^N \sum_{j=1}^{Q} 
    \left|    R_{\bm{\theta}}^{(i)}(x^{(i)}_{r,j}, t^{(i)}_{r,j})   \right|^2.
\end{align}
Here, for every input sample $\bm{u}^{(i)}$, 
$x_{ic,j}^{(i)} = x_j$ and 
$ \{(0, t_{ic,j}^{(i)})\}_{j=1}^P, \{(1, t_{ic,j}^{(i)})\}_{j=1}^P$ and $\{( x_{r,j}^{(i)}, t_{r,j}^{(i)})\}_{j=1}^Q$ are randomly sampled in the computational domain for enforcing the initial and boundary conditions and the PDE residual, respectively. In this example,  we take $P = m = 100, Q = 2,500$.
 
To obtain a set of training and test data, we randomly sample 2,000 input functions from a GRF $\sim \mathcal{N}\left(0,  25^2(-\Delta+5^2 I)^{-4}\right)$, and select a subset of $N = 1,000$ samples as training data. For each sample $u$,  
we solve the Burgers equation \ref{eq: Burger_eq} using conventional spectral methods. Specifically, assuming  periodic boundary conditions, we start from a given initial
condition $s(x,0) = u(x), x\in [0,1]$ and integrate the equation \ref{eq: Burger_eq} up to the final time $t=1$. Synthetic test data for this example are generated using
the Chebfun package \cite{driscoll2014chebfun} with a spectral Fourier discretization and a fourth-order stiff time-stepping scheme (ETDRK4) \cite{cox2002exponential} with time-step size $10^{-4}$.  Temporal snapshots of the solution are saved every $\Delta t = 0.01$ to give us 101 snapshots in total. Consequently, the test data-set contains 1,000 realizations evaluated at a $100 \times 101$ spatio-temporal grid.

\begin{figure}
     \centering
          \begin{subfigure}[b]{0.25\textwidth}
         \centering
         \includegraphics[width=\textwidth]{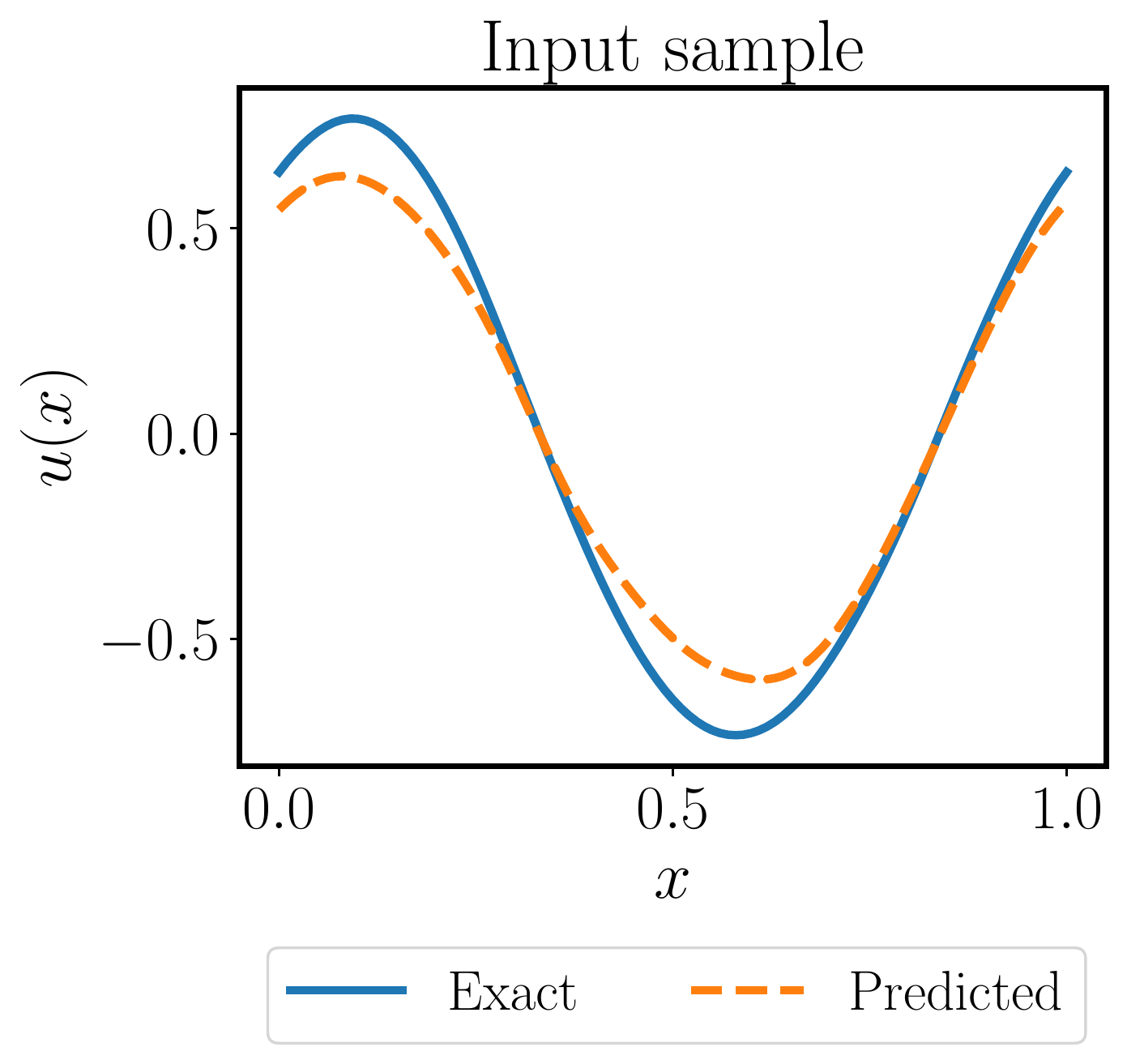}
         \caption{ }
         \label{fig: physical_deeponet_Burger_MLP_u}
     \end{subfigure}
     \begin{subfigure}[b]{0.25\textwidth}
         \centering
         \includegraphics[width=\textwidth]{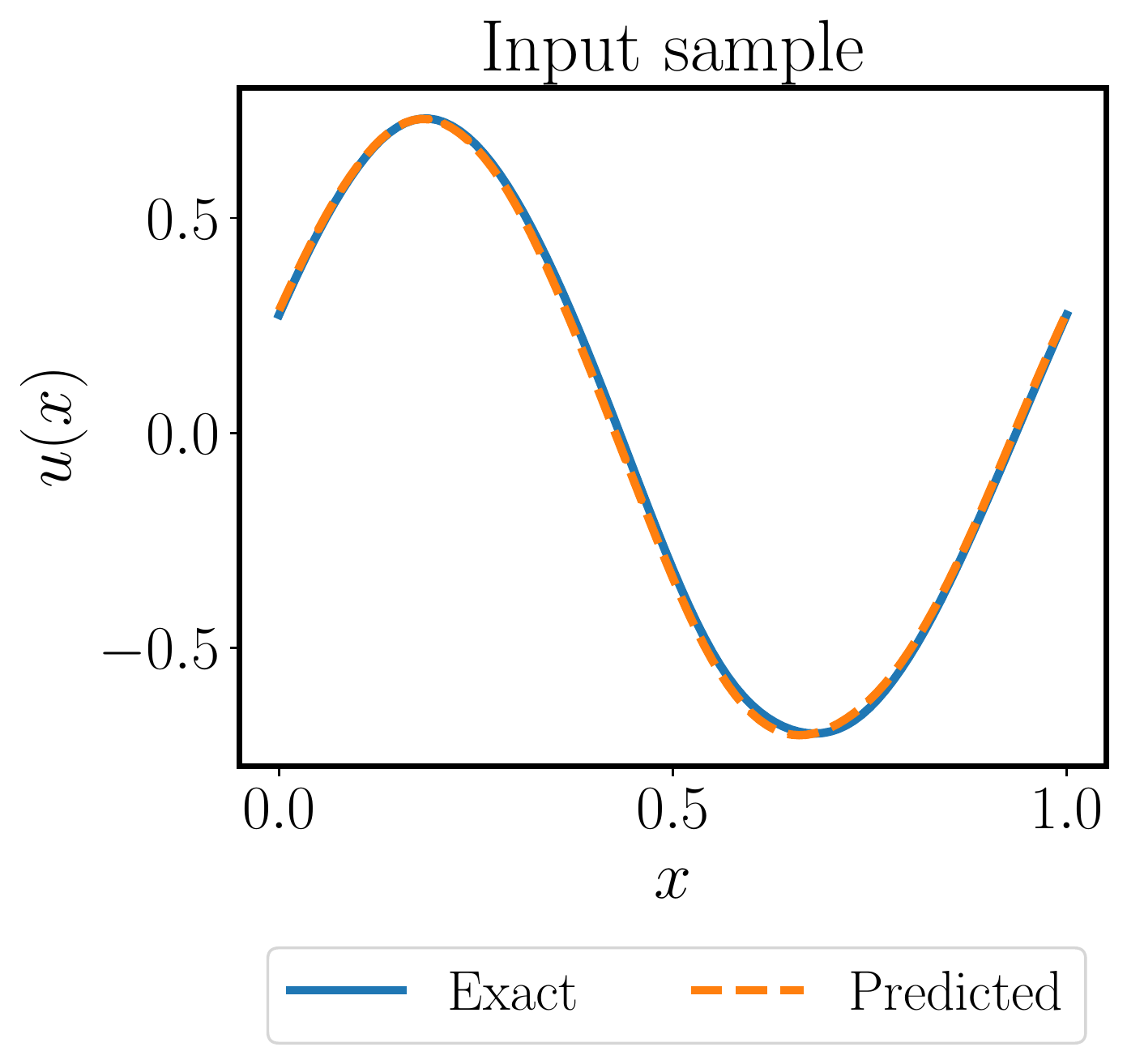}
         \caption{ }
         \label{fig: physical_deeponet_Burger_ADGM_u}
     \end{subfigure}
          \begin{subfigure}[b]{0.8\textwidth}
         \centering
         \includegraphics[width=\textwidth]{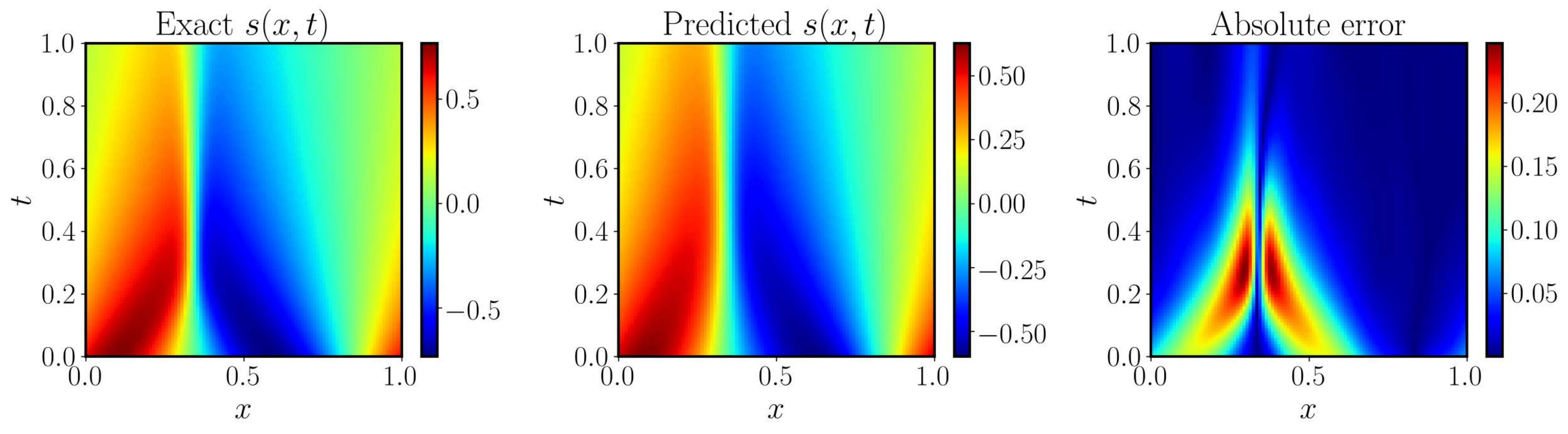}
         \caption{}
         \label{fig: physical_deeponet_Burger_MLP_s}
     \end{subfigure}
     \begin{subfigure}[b]{0.8\textwidth}
         \centering
         \includegraphics[width=\textwidth]{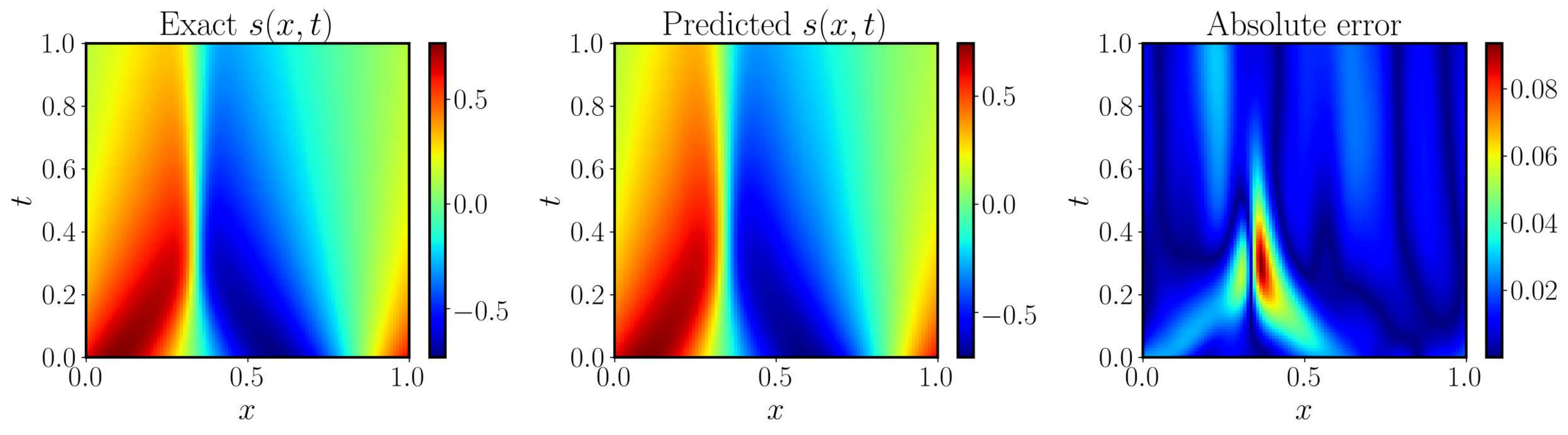}
         \caption{}
         \label{fig: physical_deeponet_Burger_ADGM_s}
     \end{subfigure}
        \caption{{\em Solving a parametric Burgers' equation:} (a)(c) Exact solution and the initial condition versus the predictions of a trained physics-informed DeepONet with a conventional MLP architecture. The resulting relative $L^2$ error of the predicted solution is $17.10e-01$. (b)(d) Exact solution and the initial condition versus the predictions of a trained physics-informed DeepONet with a modified MLP architecture \cite{wang2020understanding} and $\lambda=20$. The resulting relative $L^2$ error of the predicted solution is reduced to $3.31e-02$.}
        \label{fig: physical_deeponet_Burger_ADGM}
\end{figure}

\begin{figure}
    \centering
    \begin{subfigure}[b]{0.3\textwidth}
         \centering
         \includegraphics[width=\textwidth]{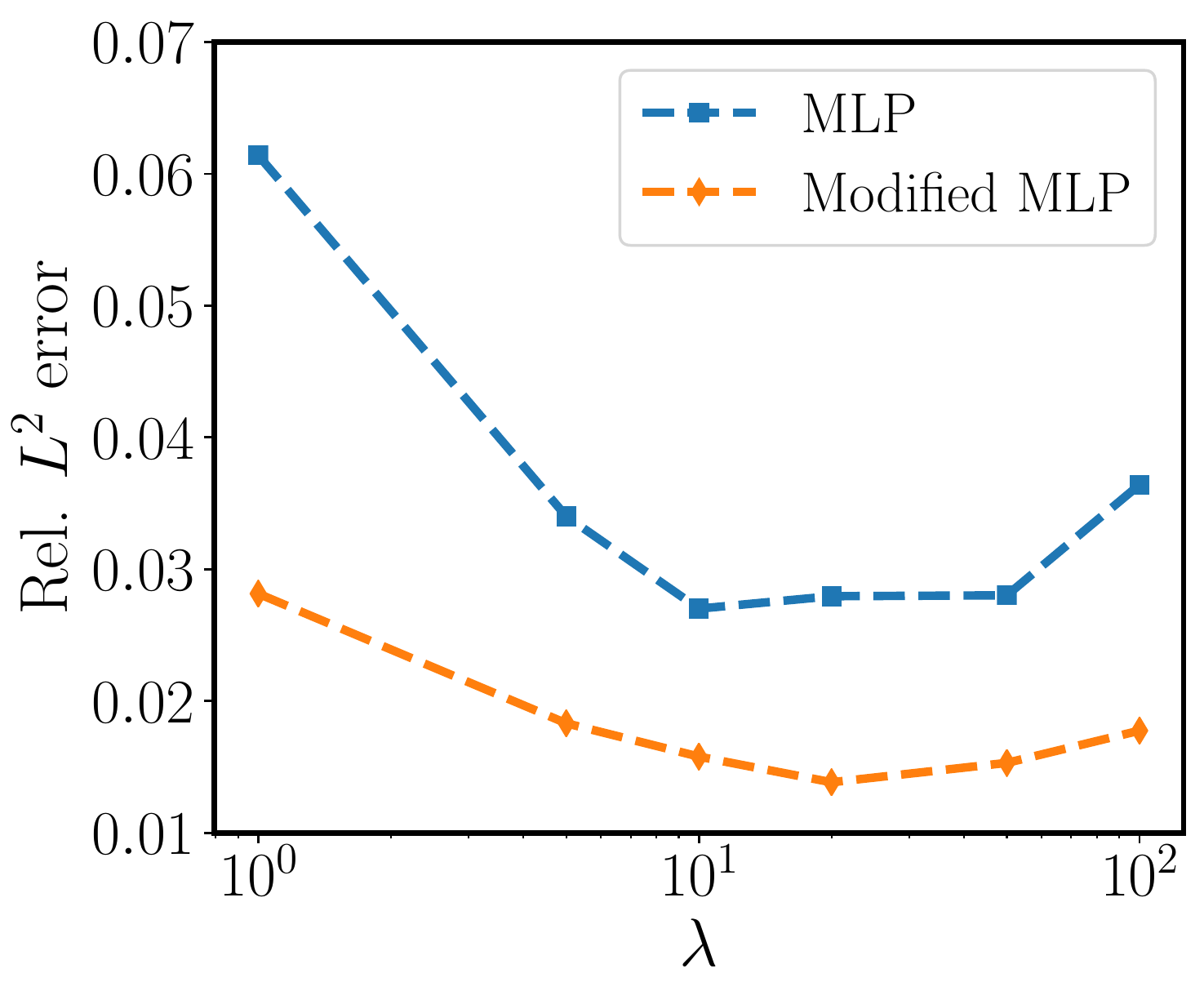}
         \caption{ }
         \label{fig: Burger_l2_error}
     \end{subfigure}
     \begin{subfigure}[b]{0.5\textwidth}
         \centering
         \includegraphics[width=\textwidth]{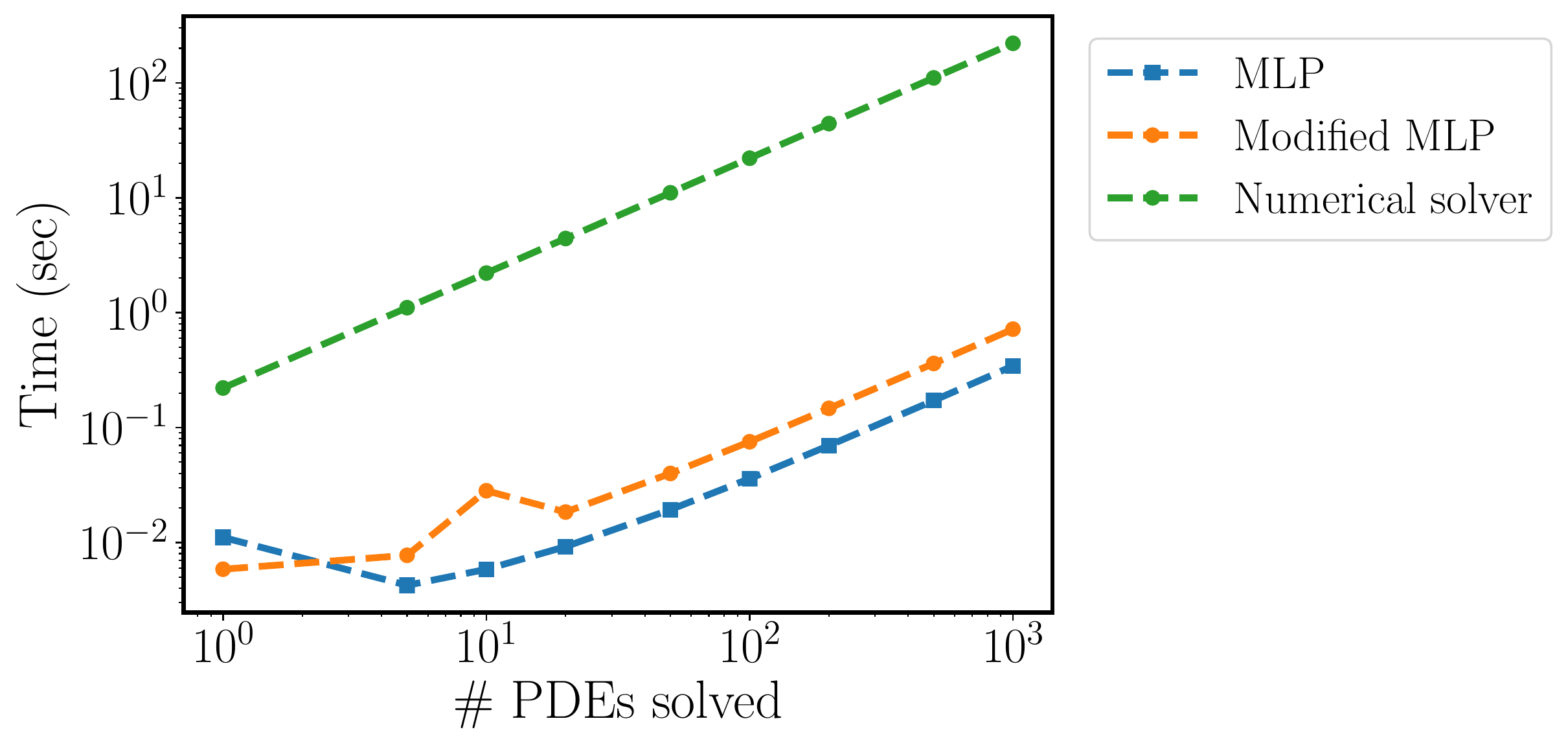}
         \caption{ }
         \label{fig: Burger_infer_time}
     \end{subfigure}
    \caption{{\em Solving a parametric Burgers' equation:} (a) The average relative $L^2$ error of training physics-informed DeepONets with standard or modified MLPs for different $\lambda \in \{1, 5, 10, 20, 50, 100\}$ over 1,000 examples in the test data-set. The smallest errors for standard and modified MLPs are $2.80e-02$ and  $1.38e-02$, respectively. (b) Computational cost (sec) for performing inference with a trained physics-informed DeepONet model (conventional or modified MLP architecture), as well as corresponding timing for solving a PDE with a conventional spectral solver \cite{driscoll2014chebfun}. Strikingly, a trained physics informed DeepOnet model can predict the solution of $\mathcal{O}(10^3)$ time-dependent PDEs in a fraction of a second -- up to three orders of magnitude faster compared to a conventional PDE solver. Reported timings are obtained on a single NVIDIA V100 GPU.}
    \label{fig: Burger}
\end{figure}

We employ two separate 7-layer fully-connected neural networks to represent the branch net and the trunk net, respectively. 
Each network is equipped with Tanh activation functions and has $100$ units per hidden layer. The physics-informed DeepONet is trained by minimizing the loss function \ref{eq: Burger_loss} via gradient descent using the Adam optimizer for $200,000$ iterations. Figure \ref{fig: physical_deeponet_Burger_MLP_s} shows the predicted solution of a trained physics-informed DeepONet for the worst sample in the test data-set, with a resulting relative $L^2$ error of $1.71e-01$. 
Moreover, a discrepancy between the exact and the predicted initial condition $u(x)$ can be observed in Figure \ref{fig: physical_deeponet_Burger_MLP_u}. This indicates that the physics-informed DeepONet cannot accurately reconstruct the initial condition, which results in a large prediction error of the full solution. To enforce the initial condition and improve the performance of physics-informed DeepONet, we consider assigning  weights to $\mathcal{L}_{IC}(\bm{\theta})$ and 
use a more powerful network architecture as the backbone of the branch net and the trunk net. Specifically, we modify the loss function \ref{eq: Burger_loss} as
\begin{align}
    \label{eq: Burger_modified_loss}
    \mathcal{L}(\bm{\theta}) = \lambda \mathcal{L}_{\text{IC}}(\bm{\theta}) + \mathcal{L}_{\text{BC}}(\bm{\theta}) + \mathcal{L}_{\text{physics}}(\bm{\theta}),
\end{align}
where $\lambda$ is a hyper-parameter that aims to balance the interplay of different terms in the loss function. Moreover,  we employ a simple modified fully-connected neural network proposed by Wang {\em et. al} \cite{wang2020understanding}, which has been  empirically proven to outperform conventional multi-layer percerptron (MLP) networks. More details can be found in Appendix \ref{Appendix: Burger}. We train the physics-informed DeepONet using standard and modified MLP networks by minimizing the modified loss function \ref{eq: Burger_modified_loss} for different $\lambda \in \{1, 5, 10, 20, 50, 100\}$ under exactly the same hyper-parameter setting.  The resulting average relative $L^2$ prediction errors are summarized in Figure \ref{fig: Burger_l2_error}.  Compared against  conventional MLPs, the modified MLP architecture is capable of consistently yielding much better prediction accuracy, which can be further improved by assigning appropriate weights in the loss function. Among all these hyper-parameters, the smallest test error $\sim 1.38\%$ is obtained for the modified fully-connected neural network with $\lambda = 20$. It is worth noting that the physics-informed DeepONet achieves the comparable accuracy compared to Fourier operator methods \cite{li2020fourier}, albeit the latter is trained on a large corpus of paired input-output data. Furthermore, visualizations corresponding to the worst example in the test data-set are shown in Figure \ref{fig: physical_deeponet_Burger_ADGM_u} and Figure \ref{fig: physical_deeponet_Burger_ADGM_s}, respectively. One can see that model predictions achieve a good agreement against the reference solutions, with a the relative $L^2$ error being reduced to $3.30\%$.

Here, we must also emphasize that a trained physics-informed DeepONet model can rapidly predict the entire spatio-temporal solution of a given Burgers equation in $\sim$10ms. Inference with physics-informed DeepONets is trivially parallelizable, allowing for the solution of $\mathcal{O}(10^3)$ PDEs in a fraction of a second, yielding up to three orders of magnitude in speed up compared to a conventional spectral solver  \cite{driscoll2014chebfun}, see Appendix Figure \ref{fig: infer_computational_cost}.

\subsection{Eikonal equation}
\label{sec: Eikonal}

Our last example aims to highlight the capability of the proposed physics-informed DeepONet to handle different type of input functions. To this end, let us consider a two-dimensional Eikonal equation of the form
\begin{align}
\label{eq: Eikonal}
\begin{aligned}
        &\|\nabla s(\bm{x})\|_2 = 1,  \\
    &  s(\bm{x}) = 0, \quad \bm{x} \in \partial \Omega,
\end{aligned}
\end{align}
where $\bm{x} = (x,y) \in \R^2$ denotes 2D spatial coordinates, $\Omega$ is an open domain with a piece-wise smooth boundary   $\partial \Omega$. A solution to the above equation is a signed distance function quantifying the distance of a point in $\Omega$ to the boundary $\partial \Omega$, i.e
\begin{align*}
    s(\bm{x}) = \left\{\begin{array}{ll}
d(\bm{x}, \partial \Omega) & \text { if } \bm{x} \in \Omega, \\
-d(\bm{x}, \partial \Omega) & \text { if } \bm{x} \in \Omega^{c},
\end{array}\right.
\end{align*}
where $d(\cdot, \cdot)$ is the distance function defined by
\begin{align}
    d(\bm{x}, \partial \Omega):=\inf _{\bm{y} \in \partial \Omega} d(\bm{x},\bm{y}).
\end{align}
Sign distance functions (SDFs) have recently sparked increased  interest in the computer vision and graphics communities as a tool for shape representation learning \cite{park2019deepsdf}. This is because SDFs can continuously represent abstract shapes or surfaces implicitly as their zero-level-set, yielding high quality shape representations, interpolation and completion from partial and noisy input data \cite{park2019deepsdf}.

In this example, we seek to learn the solution map from a well-behaved closed curve $\Gamma$ to its associated signed distance function, i.e the solution of the Eikonal equation defined in equation \ref{eq: Eikonal}. To this end, we use a DeepONet $G_{\bm{\theta}}$ to represent the unknown operator. This allows us to  define the PDE residual 
\begin{align}
    R_{\bm{\theta}}^{(i)}(x,y) = \|\nabla G_{\bm{\theta}}(\bm{\Gamma}^{(i)})(x,y) \|_2 = \left\| \sqrt{  \left( \frac{d G_{\bm{\theta}}(\bm{\Gamma}^{(i)})(x,y)}{d x} \right)^2 +  \left( \frac{d G_{\bm{\theta}}(\bm{\Gamma}^{(i)})(x,y)}{d y} \right)^2  } \right\|_2
\end{align}
Here, $\bm{\Gamma}^{(i)} = [(x^{(i)}_1, y^{(i)}_1), (x^{(i)}_2, y^{(i)}_2), \dots, (x^{(i)}_m, y^{(i)}_m)]$ denotes a parametrized curve evaluated at a set of fixed sensor locations $\{(x_j^{(i)}, y_j^{(i)})\}_{j=1}^m$.
Then, a physics-informed DeepONet can be trained by minimizing the following loss function
\begin{align}
     \label{eq: Eikonal_loss}
    \mathcal{L}(\bm{\theta}) &= \mathcal{L}_{\text{BC}}(\bm{\theta}) + \mathcal{L}_{\text{physics}}(\bm{\theta}) \\
                             &= \frac{1}{Nm} \sum_{i=1}^N \sum_{j=1}^m \left|G_{\bm{\theta}} (\bm{\Gamma}^{(i)})(x^{(i)}_{j}, y^{(i)}_{j}) \right|^2  + \frac{1}{NQ} \sum_{i=1}^N \sum_{j=1}^Q \left|  R_{\bm{\theta}}^{(i)}(x_{r,j}^{(i)}, y^{(i)}_{r,j})  - 1  \right|^2,
\end{align}
where $\mathcal{L}_{\text{BC}}(\bm{\theta})$ and $\mathcal{L}_{\text{physics}}(\bm{\theta})$ are used to impose the zero boundary condition and the PDE residual, respectively. Moreover, for each input curve $\bm{\Gamma}^{(i)}$, $\{(x_{r,j}^{(i)}, y^{(i)}_{r,j})\}_{j=1}^Q$ are uniformly sampled in the given computational domain. Unlike the previous parametric PDE examples, it is worth noting that the "input functions" of this example are actually defining a variable computational domain. In the following we consider two case studies corresponding to different families of curves: parametric circles and airfoils.

\subsubsection{Case I: Parametric circles}

We start with a simple case corresponding to circular boundaries $\partial \Omega$ centered at the origin, each having a different radius. In this case, the corresponding signed distance function that solves equation \ref{eq: Eikonal} can be analytically derived. For example, suppose that $\Gamma$ is a circle with radius $r$, then the signed distance function is given by
\begin{align}
    s(x,y) = \sqrt{x^2 + y^2} - r
\end{align}
To generate a set of training data, we randomly choose $N = 1,000$ circles with radii sampled from a uniform distribution. Then, for each input circle $\Gamma^{(i)}$ with radius $r^{(i)}$, we have $\{(x_{j}^{(i)}, y^{(i)}_{j})\}_{j=1}^Q   =  \{(r^{(i)}\cos\theta_j, r^{(i)} \sin\theta_j)\}_{j=1}^Q$, where $\{\theta_j\}_{j=1}^Q$ are evenly spaced in $[0, 2\pi]$. Here, we consider a computational domain $D = [-2, 2] \times [-2, 2]$ and we set $m =100, Q = 1,000$. 

\begin{figure}
    \centering
    \includegraphics[width=0.8\textwidth]{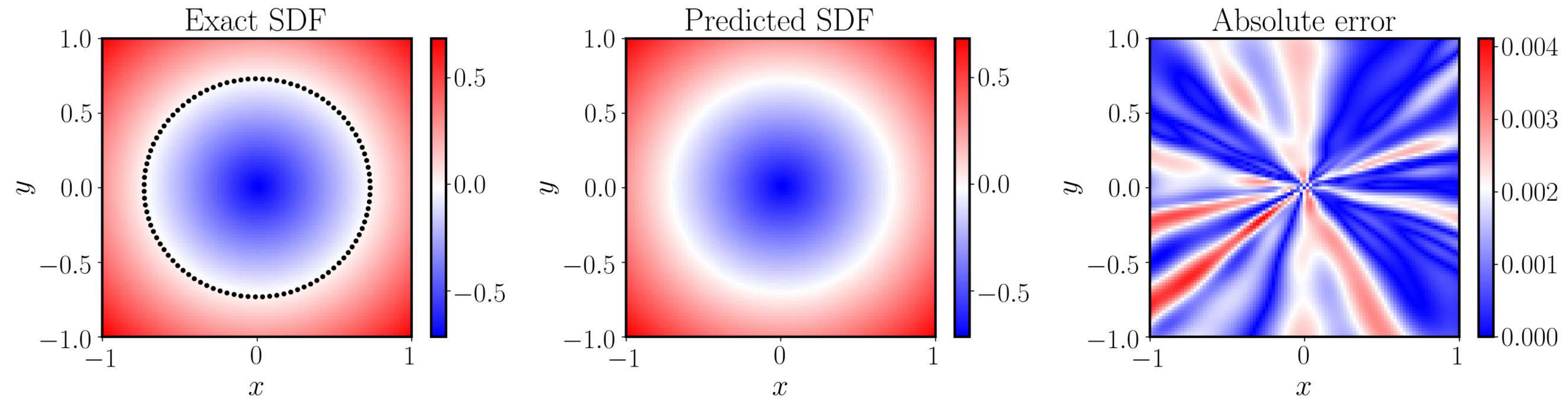}
    \caption{{\em Solving a parametric Eikonal equation (circles):} Exact solutions versus the predicted solutions of a trained physics-informed DeepONet for a representative input sample. The black dots represent the location of sensors on the circular boundary.}
    \label{fig: physical_deeponet_eikonal_circle_sdf}
\end{figure}

The branch net and the trunk networks are two separate 6-layer fully-connected neural network with $50$ neurons per hidden layer. We train the physics-informed DeepONet by minimizing the loss function \ref{eq: Eikonal_loss} for $80,000$ iterations of gradient descent using the Adam optimizer. As shown in Figure \ref{fig: physical_deeponet_eikonal_circle_sdf}, an excellent agreement can be achieved between the exact and the predicted signed distance functions for a representative example in the test data-set. The  relative $L^2$ prediction error averaged over 1,000 examples in the test data-set is $4.22e-03$.  More model predictions for different input samples can be found in Appendix Figure \ref{fig: physical_deeponet_eikonal_circle_sdf_examples}.

\subsubsection{Case II: Parametric airfoils}

Next, we consider a more complex case where the parameterized curves correspond to airfoils of different shapes. To obtain a set of training and test data, we use the UIUC Airfoil Data Site \cite{selig1996uiuc} which contains a total of 1,552 airfoil geometries.
We use the first 1,000 shapes as training data, and the rest are included in the test data-set. Without loss of generality, we normalize the airfoil shapes to have zero mean and unit variance. 

\begin{figure}
    \centering
    \includegraphics[width=0.8\textwidth]{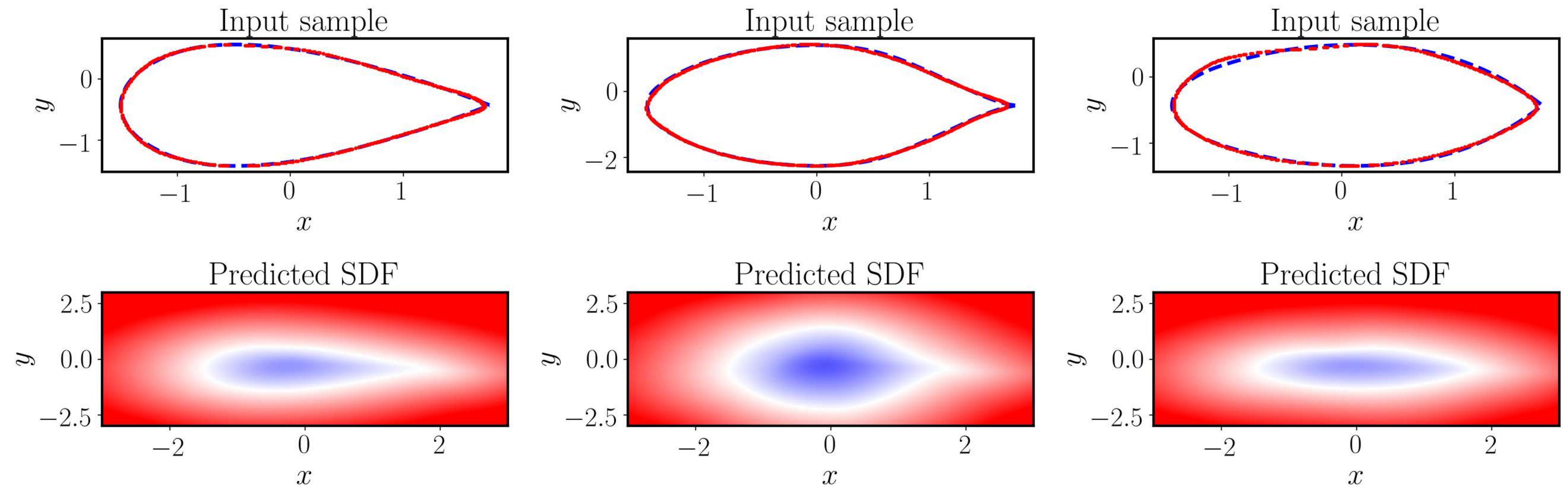}
    \caption{{\em Solving a parametric Eikonal equation (airfoils):} {\em Top}: Exact airfoil geometry versus the zero-level-set obtained from the predicted signed distance function for three different input examples in the test data-set. {\em Bottom}: Predicted signed distance function of a trained physics-informed DeepONet for three different airfoil geometries in the test data-set. }
    \label{fig: physical_deeponet_eikonal_airfoil_examples}
\end{figure}

In this example, the computation domain is the unit square $[-3, 3] \times [-3, 3]$ and the branch net and the trunk net are two separate 6-layer fully-connected neural networks with $100$ neurons per hidden layer. Both two networks are equipped with ELU activation functions. We train the physics-informed DeepONet for $120,000$ iterations of gradient descent using the Adam optimizer. To evaluate the performance of the trained model, we visualize the zero-level set of the learned signed distance function and compare it with the exact airfoil geometry.  As shown in Figure \ref{fig: physical_deeponet_eikonal_airfoil_examples}, the zero-level-sets achieve a good agreement with the exact airfoil geometries. From the results of these two case studies, one may conclude that
the proposed framework is capable of achieving a relatively accurate approximation of the exact signed distance function.


\section{Summary and Discussion}\label{sec:discussion}

This paper presents physics-informed DeepONets; a new model class for nonlinear operator learning in infinite-dimensional Banach spaces. We illustrate how automatic differentiation can be leveraged to bias the outputs of DeepONets towards physically-consistent predictions for systems whose evolution can be described by systems of differential equations. By doing so, we observe significant improvements in predictive accuracy (up to 1-2 orders of magnitude reduction in predictive errors), enhanced generalization performance, as well as enhanced data-efficiency (up to 100\% reduction in the number of examples required to train a DeepONet model). Strikingly, the proposed framework can be employed to solve parametric PDEs in an unsupervised manner, i.e. without any paired input-output observations. A series of comprehensive numerical studies demonstrate not only significant improvements in terms of predictive accuracy, but also a remarkable reduction number of training data required to train a DeepONet model.

Despite the encouraging results presented here, numerous questions remain open and require further investigation. Motivated by the  successful application of Fourier feature networks \cite{tancik2020fourier} in section \ref{sec: Antiderivative}, it is natural to ask: For a given parametric PDE, what is the optimal features embedding or network architecture for physics-informed DeepONets?  Recently, Wang {\em et. al.} \cite{wang2020eigenvector} proposed a multi-scale Fourier feature network to tackle PDEs with multi-scale behavior. Such an architecture may be potentially used as the backbone of the physics-informed DeepONet to learn multi-scale operators and solve multi-scale parametric PDEs.  Another question arises from the possibility of achieving improved performance by assigning weights in the physics-informed DeepOnet loss function, as discussed in section \ref{sec: Burger}. It has been shown that these weights play an important role in enhancing the trainability of constrained neural networks  \cite{wang2020understanding,wang2020and,mcclenny2020self}.
Therefore, it is natural to ask: What are the appropriate weights to use for training physics-informed DeepONets? How to design effective algorithms for accelerating training and ensuring accuracy and robustness in the predicted outputs? We believe that addressing these questions will not only enhances the performance of physics-informed DeepONets, but pave a new way for modeling and simulating complex, non-linear and multi-scale physical systems across diverse applications in science and engineering.

\section*{Author Contributions}
SW and PP conceptualized the research and designed the numerical studies. SW and HW implemented the methods and conducted the numerical experiments. PP provided funding and supervised all aspects of this work. All authors contributed in writing the manuscript.

\section*{Acknowledgements}
This work received support from DOE grant DE-SC0019116, AFOSR grant FA9550-20-1-0060, and DOE-ARPA grant DE-AR0001201. 

\bibliographystyle{unsrt}
\bibliography{references}


\appendix
\input{appendix}

\end{document}

%% file: appendix.tex
\clearpage
\section{Nomenclature}
Table \ref{tab: Notations} summarizes the main symbols and notation used in this work.

\begin{table}[h]
\renewcommand{\arraystretch}{1.4}
    \centering
    \begin{tabular}{ll} 
    \Xhline{3\arrayrulewidth} 
    Notation     & Description \\
    \Xhline{3\arrayrulewidth} 
        $\bm{u}(\cdot)$ & an input function \\  
        $\bm{s}(\cdot)$ & a solution to a parametric PDE \\
        $G$         & an operator \\
        $G_{\bm{\theta}}$  &  an unstacked DeepONet representation of the operator $G$ \\
        $\bm{\theta}$ &  all trainable parameters of a DeepONet \\
        $\{\bm{x}_i\}_{i=1}^m$  & $m$ sensor points where input functions $\bm{u}(\bm{x})$ are evaluated\\
        $[u(\bm{x}_1), u(\bm{x}_2), \dots, u(\bm{x}_m)]$ & an input of the branch net, representing the input function $u$ \\ 
        $\bm{y}$          & an input of the trunk net, a point in the domain of $G(u)$ \\
        N                & number of input samples in the training data-set \\
        M               &  number of locations for evaluating the input functions $u$        \\
        P               &  number of locations for evaluating the output functions $G(u)$        \\
        Q                & number of collocation points for evaluating the PDE residual \\
        GRF             &  a Gaussian random field                   \\
        SDF             & a signed distance function \\
        $l$              & length scale of a Gaussian random field   \\
        $\mathcal{L}_{\text{operator}}(\bm{\theta})$ & a loss function to fit available observations \\
    $\mathcal{L}_{\text{physics}}(\bm{\theta})$ & a loss function to fit the underlying physical laws \\
    \Xhline{3\arrayrulewidth}
    \end{tabular}
    \caption{{\em Nomenclature}: Summary of the main symbols and notation used in this work.}
    \label{tab: Notations}
\end{table}

\clearpage
\section{Hyper-parameter settings}

\label{sec: parameters}
In all examples considered in this work, the branch net and the trunk net are equipped with  hyperbolic tangent activation functions (Tanh), except for the  Eikonal benchmark (airfoils), where ELU activations were employed. Physics-informed DeepONet models are trained via mini-batch gradient descent with a batch-size of 10,000 using the Adam optimizer \cite{kingma2014adam} with default settings and exponential learning rate decay with a decay-rate of 0.9 every 1,000 iterations. In this work, we tuned these hyper-parameters manually, without attempting to find the absolute best hyper-parameter setting. This process can be automated in the future leveraging effective techniques for meta-learning and hyper-parameter optimization \cite{finn2017model}.

\begin{table}[h]
\renewcommand{\arraystretch}{1.4}
    \centering
    \begin{tabular}{c|ccccccc}
        \Xhline{3\arrayrulewidth}
      Case   & Input function space & \# Sensors m &  \#u Train & P & Q & \# u Test & Iterations  \\
      \hline
     Anti-derivative operator & $l=0.2$ & 100 & 10,000 & 1 & 100 & 1,000  & $4 \times 10^{4}$ \\
      1D ODE (regular input) & $l=0.2$ & 100 & 10,000 & 1 & 100 & 1,000 & $4 \times 10^{4}$\\
    1D ODE (irregular input) & $l=0.01$ & 200 & 10,000 & 1 & 200 & 1,000 & $3 \times 10^{5}$\\
     Diffusion-reaction & $l=0.2$ & 100 & 10,000 & 100 & 100 & 1,000 & $1.2 \times 10^{5}$\\
    Burgers & $ - $ & 100 & 1,000 & 100 & 2,500 & 1,000  &  $2 \times 10^{5}$\\\
        Eikonal (circles) & $ - $ & 100 & 1,000 & 100 & 1,000 & 1,000 & $8 \times 10^{4}$ \\
         Eikonal (airfoils) & $ - $ & 250 & 1,000 & 250 & 1,000 & 500 & $1.2 \times 10^{5}$ \\
    \Xhline{3\arrayrulewidth}
    \end{tabular}
    \caption{Default hyper-parameter settings for each benchmark employed in this work  (unless otherwise stated).}
    \label{tab: parameters_case}
\end{table}

\begin{table}[h]
\renewcommand{\arraystretch}{1.4}
    \centering
    \begin{tabular}{c|cccccc}
        \Xhline{3\arrayrulewidth}
        Case   &   Trunk depth &  Trunk width & Branch depth & Branch width  \\
        \hline
        Anti-derivative operator & 50 & 5  &   50 & 5  \\
         1D ODE (regular input) & 50 & 5  &   50 & 5  \\
        1D ODE (irregular input) & 200 & 5  &   200 & 5  \\
         Diffusion-reaction  & 50 & 5  &   50 & 5  \\
        Burger & 100 & 7  &   100 & 7  \\
        Eikonal (circles) & 50 & 6  &   50 & 6  \\
        Eikonal (airfoils) & 100 & 7  & 100 & 7  \\
           \Xhline{3\arrayrulewidth}
    \end{tabular}
    \caption{Physics-informed DeepONet architectures for each benchmark employed in this work (unless otherwise stated). }
    \label{tab: Physics_informed_DeepONet_size}
\end{table}

\begin{table}[h]
\renewcommand{\arraystretch}{1.4}
    \centering
    \begin{tabular}{c|cccccc}
        \Xhline{3\arrayrulewidth}
        Case   &   Trunk depth &  Trunk width & Branch depth & Branch width  \\
        \hline
        Antiderivative operator & 100 & 3  &  100 & 3  \\
         1D ODE (regular input) & 50 & 5  &   50 & 5  \\
        1D ODE (irregular input) & 50 & 5  &   50 & 5  \\
         Diffusion-reaction  & 50 & 5  &   50 & 5  \\
           \Xhline{3\arrayrulewidth}
    \end{tabular}
    \caption{Conventional DeepONet \cite{lu2019deeponet}  architectures for each corresponding benchmark (unless otherwise stated).  }
    \label{tab: DeepONet_size}
\end{table}

\clearpage
\section{Computational cost}

\label{sec: computational_cost}

{\bf Training:} Table \ref{sec: computational_cost} summarizes the computational cost  (hours) of training DeepONet and physics-informed DeepONet models with different network architectures. 
The size of different models as well as network architectures are listed table \ref{tab: DeepONet_size} and \ref{tab: Physics_informed_DeepONet_size}, respectively. All networks are trained using a single V100 card. It can be observed that training a physics-informed DeepONet
model is generally slower than  training a conventional DeepONet. This is expected as physics-informed DeepONets require to compute the PDE residual via automatic differentiation, yielding a lager computational graph, and, therefore, a higher computational cost.

\begin{table}[h]
\renewcommand{\arraystretch}{1.4}
    \centering
    \begin{tabular}{c|c| c}
     \Xhline{3\arrayrulewidth}
       Case  &  Model (Architecture) &  Training time (hours)  \\
       \hline
      \multirow{2}{*}{ Anti-derivative operator} & DeepONet  & 0.03
      \\ &Physics-informed DeepONet  & 0.15    \\
      \hline
    \multirow{2}{*}{ 1D ODE (regular input)} & DeepONet  & 0.03
      \\ &Physics-informed DeepONet  & 0.15   \\
      \hline
        \multirow{2}{*}{ 1D ODE (irregular input)} & Physics-informed DeepONet (MLP)  & 1.61
      \\ &Physics-informed DeepONet (FF)  & 1.37      \\
      \hline
    \multirow{2}{*}{ Diffusion-reaction } & DeepONet   & 1.13
      \\ &Physics-informed DeepONet  &  2.27     \\
      \hline
          \multirow{2}{*}{ Burgers } & Physics-informed DeepONet (MLP)   & 7.61
      \\ &Physics-informed DeepONet (Modified MLP) & 9.25     \\
      \hline
      Eikonal (circle) & Physics-informed DeepONet  &  0.76 \\
      \hline
      Eikonal (airfoil) & Physics-informed DeepONet  &   0.38 \\
    \Xhline{3\arrayrulewidth}
    \end{tabular}
    \caption{Computational cost (hours) for training DeepONet and physics-informed DeepONet models across the different becnhmarks and architectures employed in this work. Reported timings are obtained on a single NVIDIA V100 GPU.}
    \label{tab: computational_cost}
\end{table}

\begin{figure}[h]
    \centering
    \includegraphics[width=0.6\textwidth]{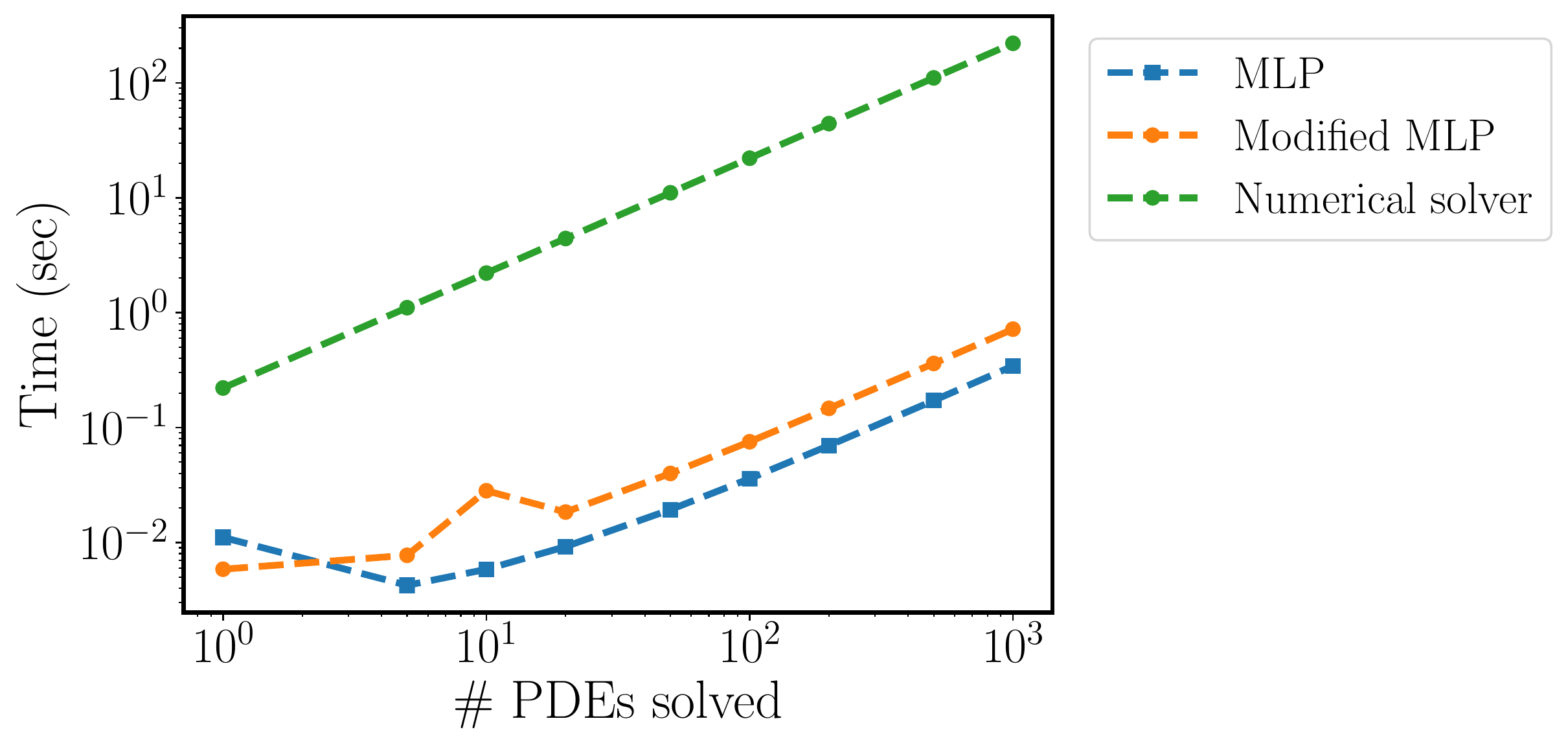}
    \caption{{Solving a parametric Burgers' equation:} Computational cost (sec) for performing inference with a trained physics-informed DeepONet model (conventional or modified MLP architecture), as well as corresponding timing for solving a PDE with a conventional spectral solver \cite{driscoll2014chebfun}. Strikingly, a trained physics informed DeepOnet model can predict the solution of $\mathcal{O}(10^3)$ time-dependent PDEs in a fraction of a second -- up to three orders of magnitude faster compared to a conventional PDE solver. Reported timings are obtained on a single NVIDIA V100 GPU.}
    \label{fig: infer_computational_cost}
\end{figure}

{\bf Inference:} A trained physics-informed DeepONet model can rapidly predict the entire spatio-temporal solution of the Burgers equation in $\sim$10ms. Inference with DeepONets is trivially parallelizable, allowing for the solution of $\mathcal{O}(10^3)$ PDEs in a fraction of a second, yielding up to three orders of magnitude in speed up compared to a traditional spectral solver \cite{driscoll2014chebfun} (see Figure \ref{fig: infer_computational_cost}).

\clearpage
\section{Anti-derivative}


\begin{figure}[h]
    \centering
   \begin{subfigure}[b]{0.4\textwidth}
         \centering
         \includegraphics[width=\textwidth]{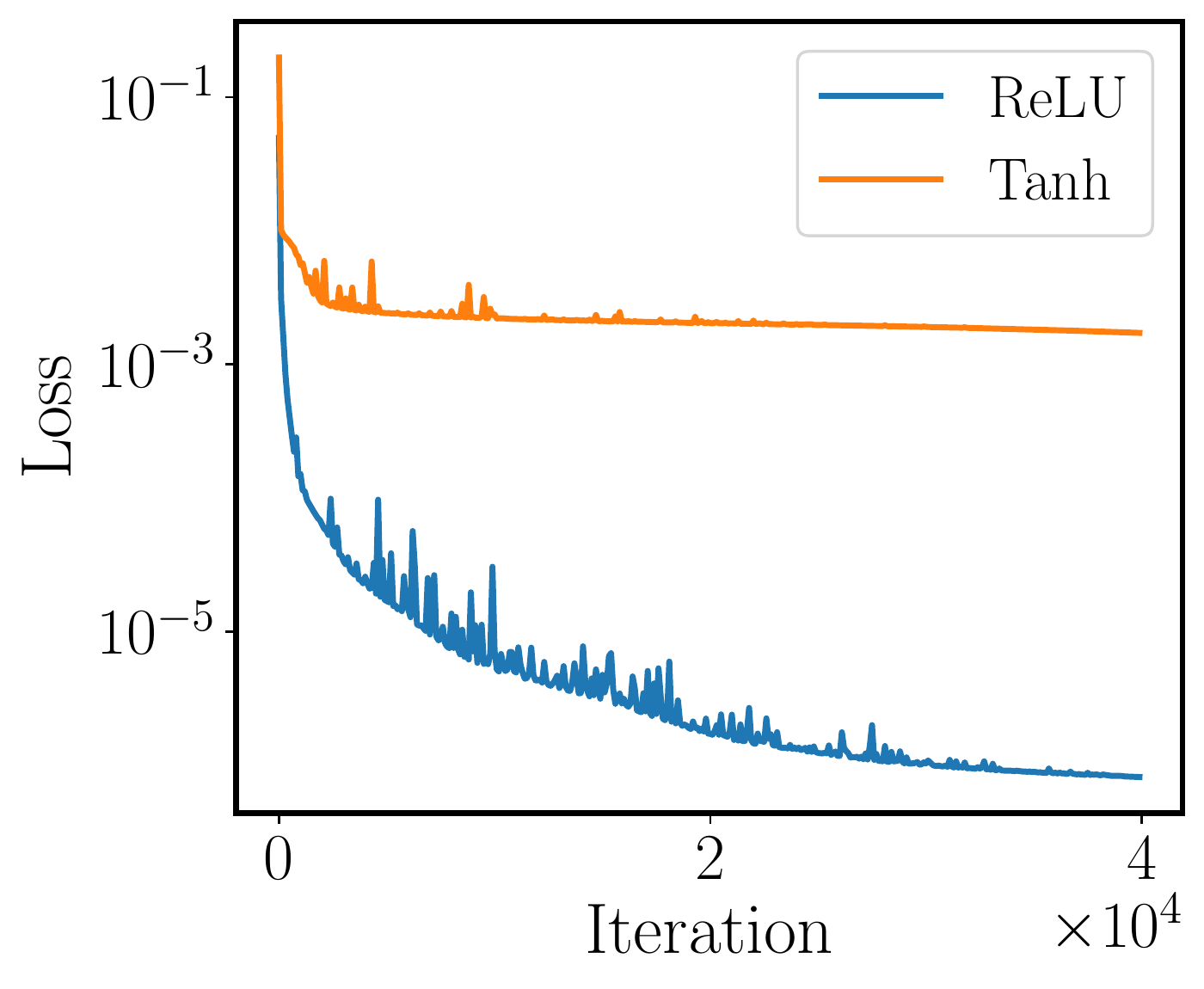}
         \caption{ }
         \label{fig: deeponet_antideriv_loss}
     \end{subfigure}
     \begin{subfigure}[b]{0.4\textwidth}
         \centering
         \includegraphics[width=\textwidth]{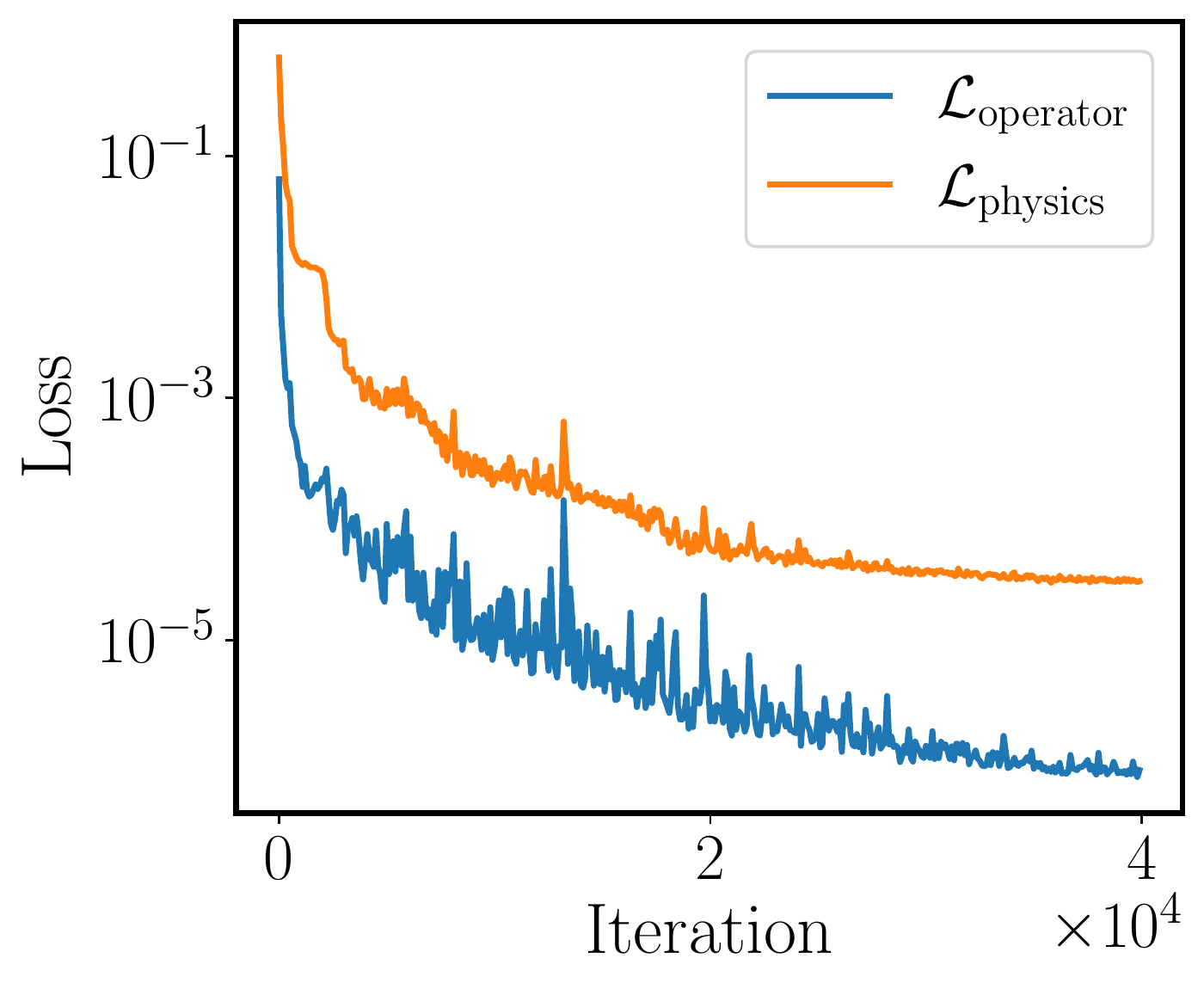}
        \caption{}
         \label{fig: physical_deeponet_antideriv_loss}
     \end{subfigure}
   \caption{{\em Learning an anti-derivative operator:}  (a) Training loss convergence of a conventional DeepONet model equipped with different activation functions for 40,000 iterations of gradient descent using the Adam optimizer. (b) Training loss convergence of a physics-informed DeepONet equipped with Tanh activations for 40,000 iterations of gradient descent using the Adam optimizer.}
    \label{fig: antideriv_loss}
\end{figure}

\begin{figure}[h]
    \centering
    \includegraphics[width=0.4\textwidth]{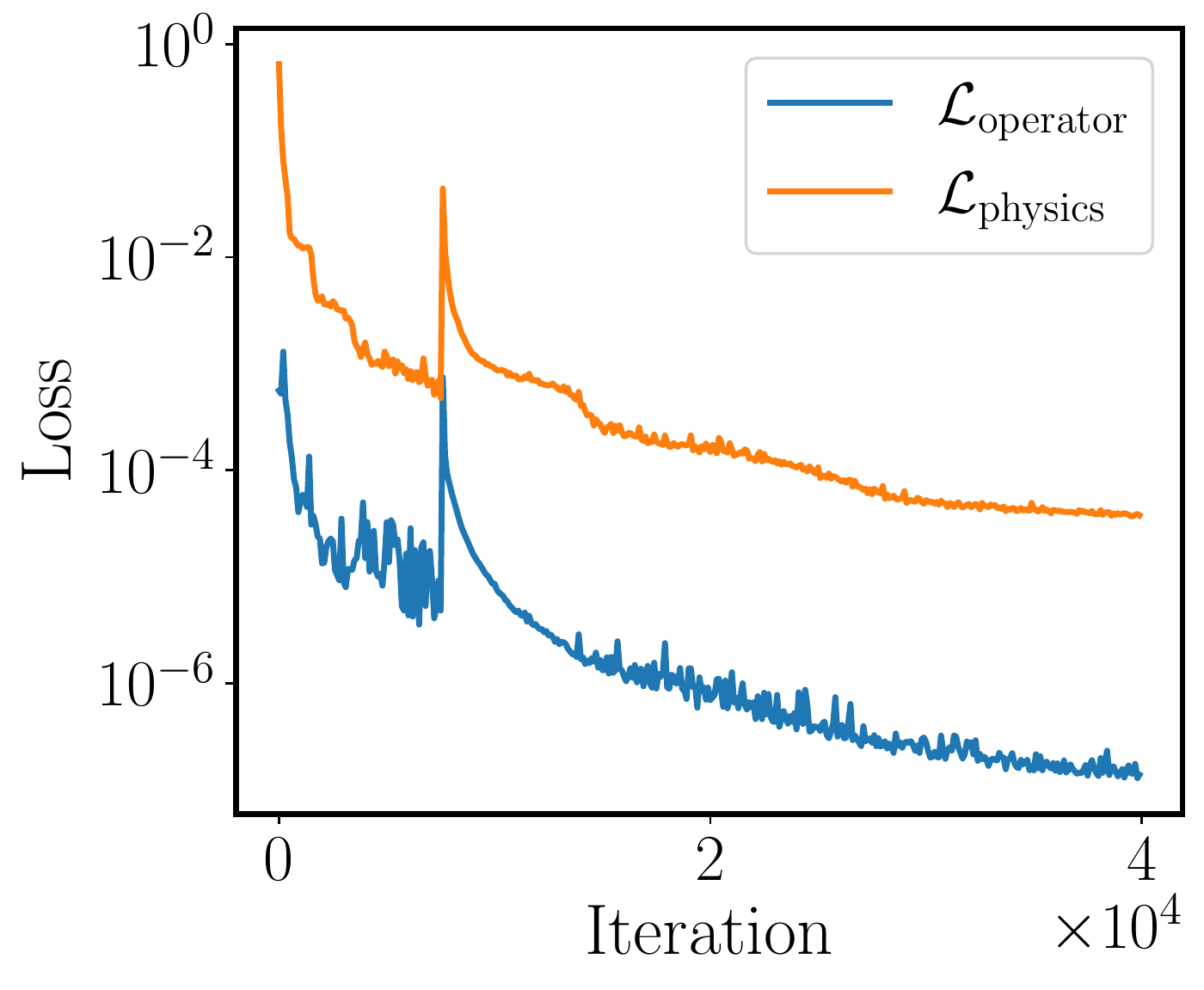}
    \label{fig: physical_deeponet_ODE_loss}
     \caption{{\em Solving a 1D parametric ODE:}  Training loss convergence of a physics-informed DeepONet for 40,000 iterations of gradient descent using the Adam optimizer without any paired input-output data, except the initial condition. }
\end{figure}

\begin{figure}[h]
     \centering
     \begin{subfigure}[b]{0.8\textwidth}
         \centering
         \includegraphics[width=\textwidth]{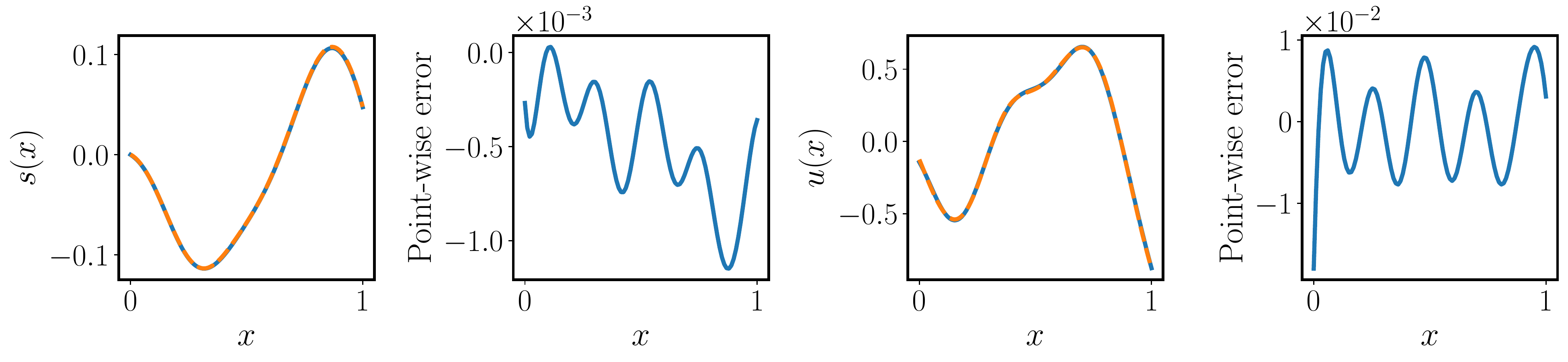}
         \caption{ }
         \label{fig: physical_deeponet_ODE_s_u_1}
     \end{subfigure}
     \begin{subfigure}[b]{0.8\textwidth}
         \centering
         \includegraphics[width=\textwidth]{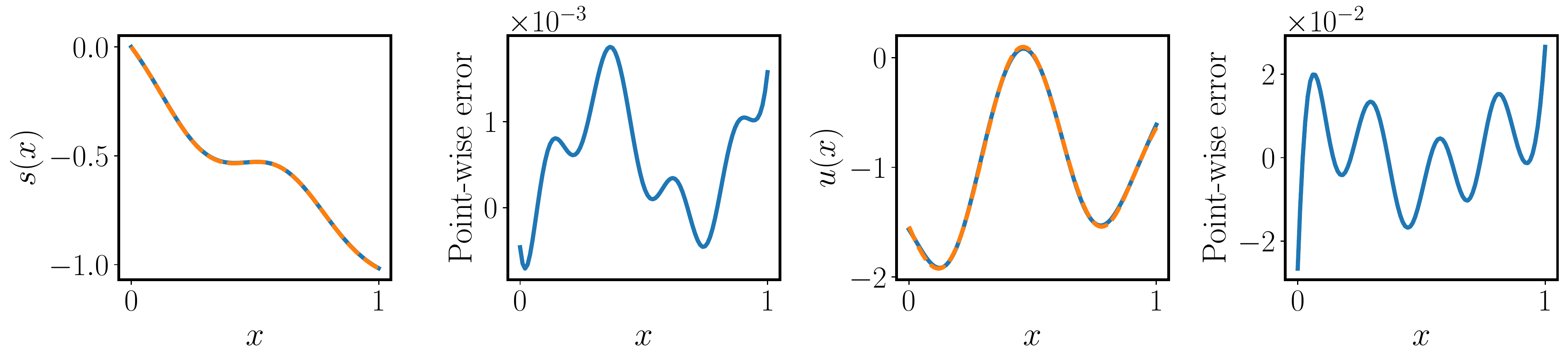}
         \caption{}
         \label{fig: physical_deeponet_ODE_s_u_2}
     \end{subfigure}
          \begin{subfigure}[b]{0.8\textwidth}
         \centering
         \includegraphics[width=\textwidth]{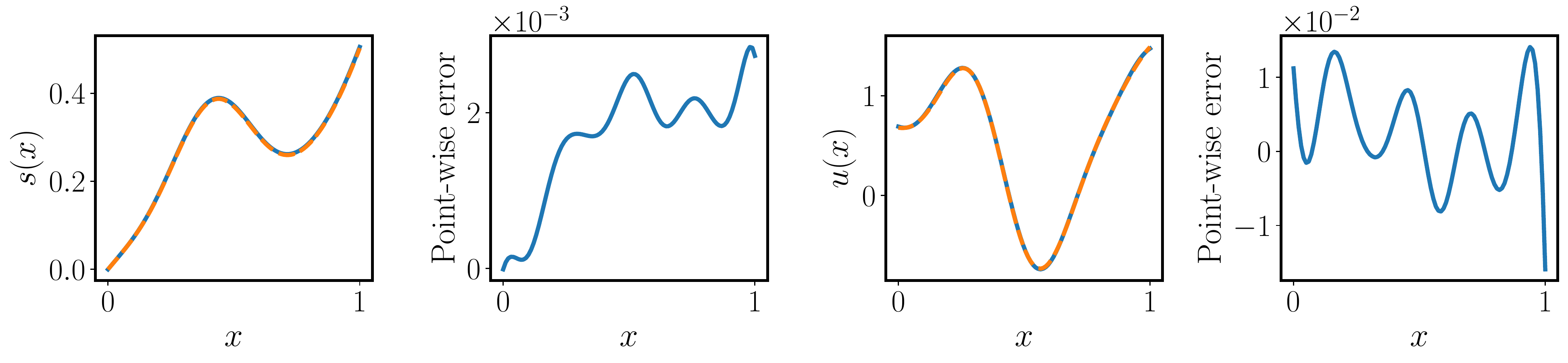}
         \caption{}
         \label{fig: physical_deeponet_ODE_s_u_3}
     \end{subfigure}
         \caption{{\em Solving a 1D parametric ODE:} Predicted solutions $s(x)$ and corresponding ODE residuals $u(x)$ for  a trained physics-informed DeepONet, across three different examples in the test data-set. }
        \label{fig: physical_deeponet_ODE_s_u_examples}
\end{figure}

\begin{figure}[h]
    \centering
     \begin{subfigure}[b]{0.4\textwidth}
         \centering
         \includegraphics[width=\textwidth]{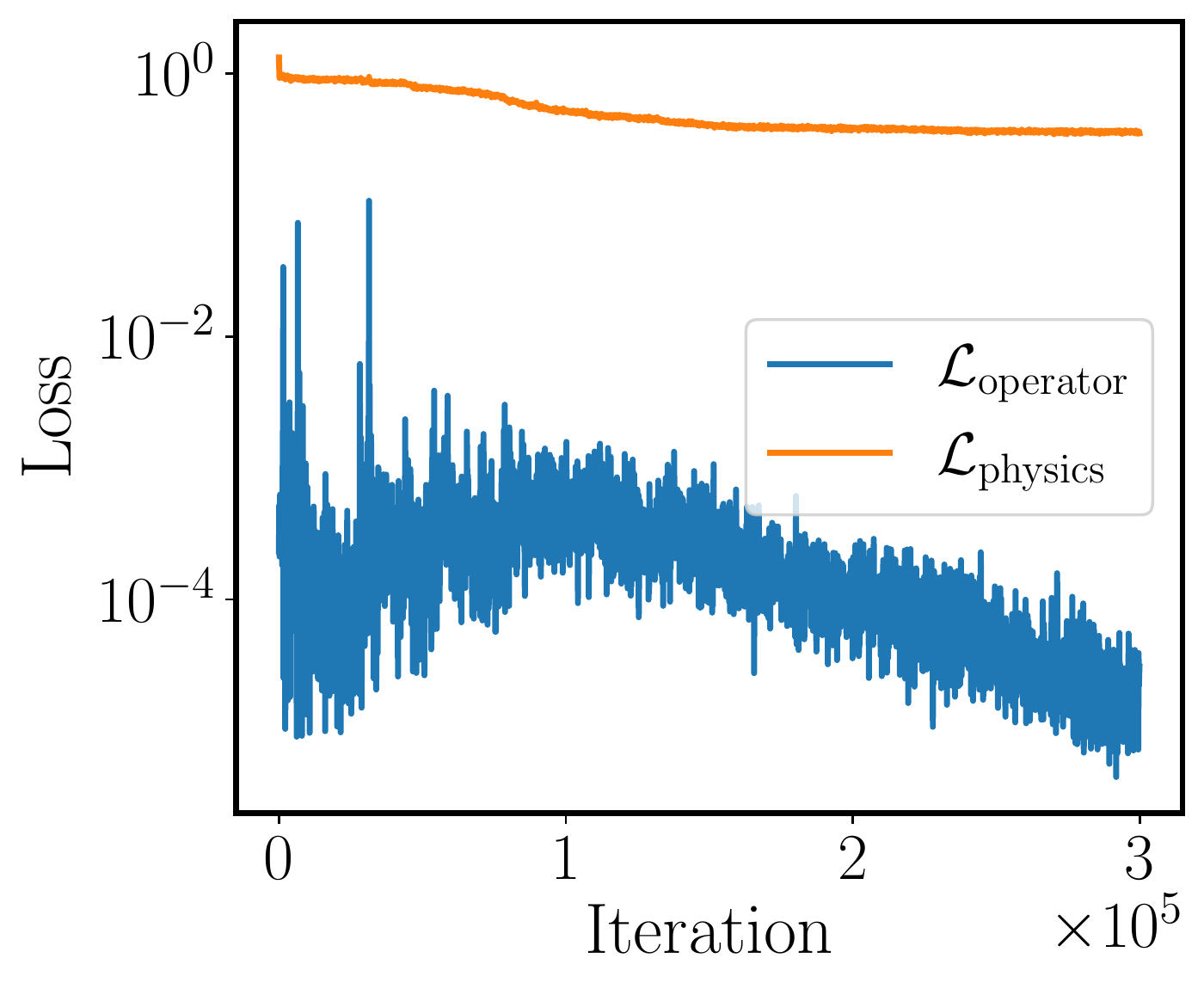}
         \caption{ }
         \label{fig: physical_deeponet_MLP_ODE_loss}
     \end{subfigure}
      \begin{subfigure}[b]{0.4\textwidth}
         \centering
         \includegraphics[width=\textwidth]{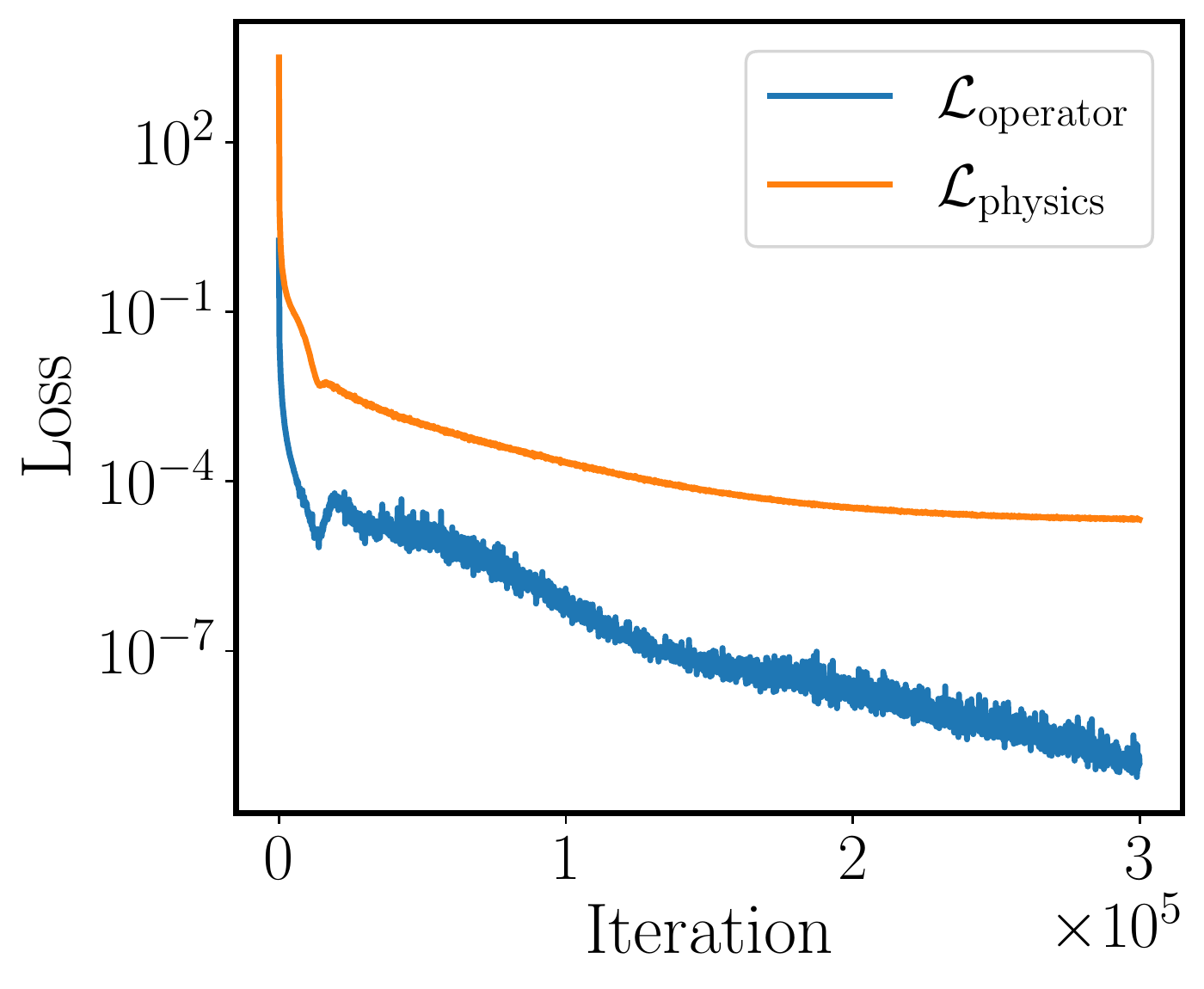}
         \caption{ }
         \label{fig: physical_deeponet_FF_ODE_loss}
     \end{subfigure}
       \caption{{\em Solving a 1D parametric ODE with irregular input functions:} (a)(b) Training loss convergence of a  physics-informed DeepONets using a conventional fully-connected neural network, and a Fourier feature network, respectively, for 300,000 iterations of gradient descent using the Adam optimizer. }
    \label{fig: physical_deeponet_MLP_FF_ODE_loss}
\end{figure}

\begin{figure}[h]
     \centering
     \begin{subfigure}[b]{0.4\textwidth}
         \centering
         \includegraphics[width=\textwidth]{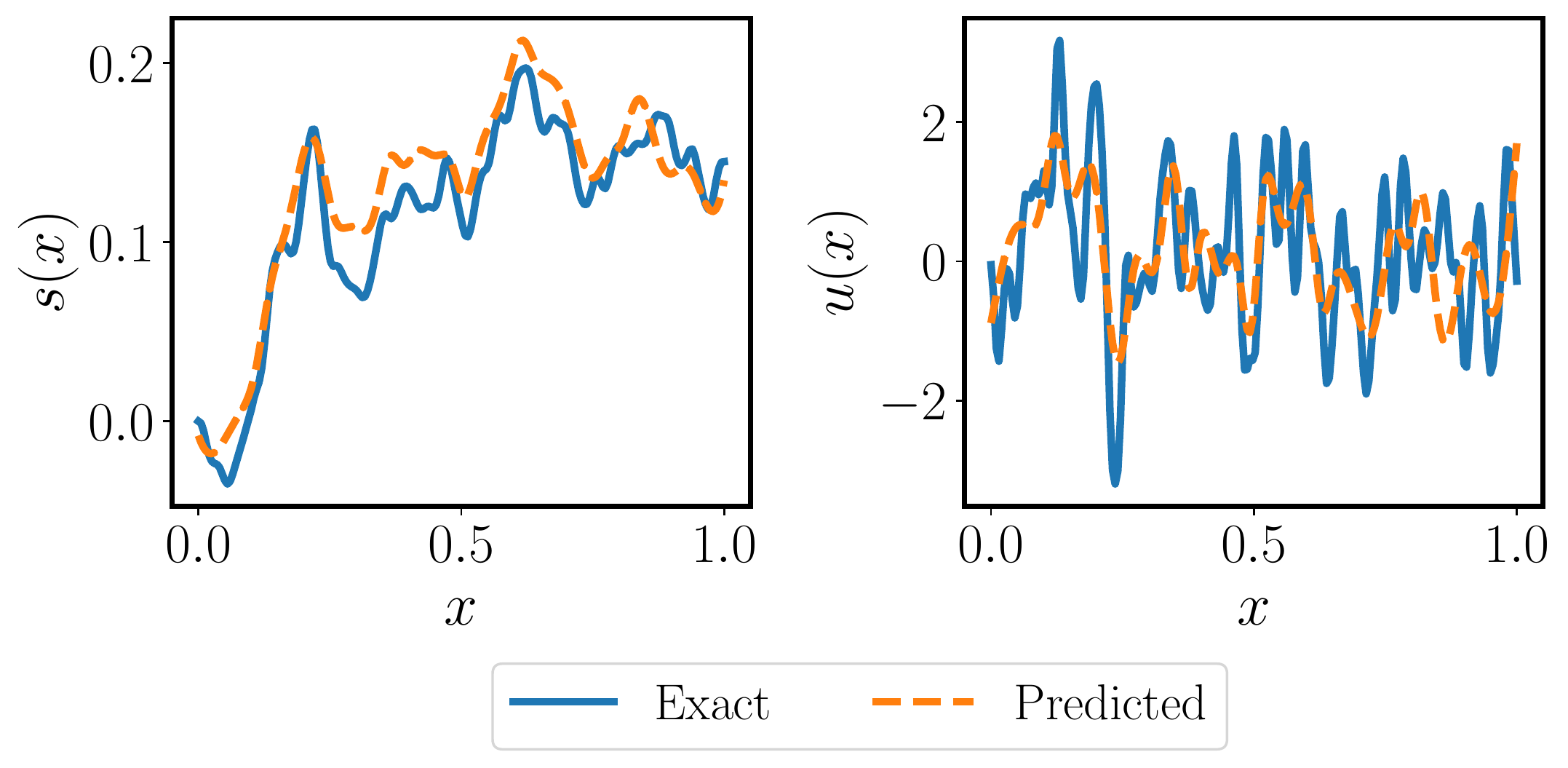}
         \caption{ }
         \label{fig: physical_deeponet_MLP_s_u_1}
     \end{subfigure}
     \begin{subfigure}[b]{0.4\textwidth}
         \centering
         \includegraphics[width=\textwidth]{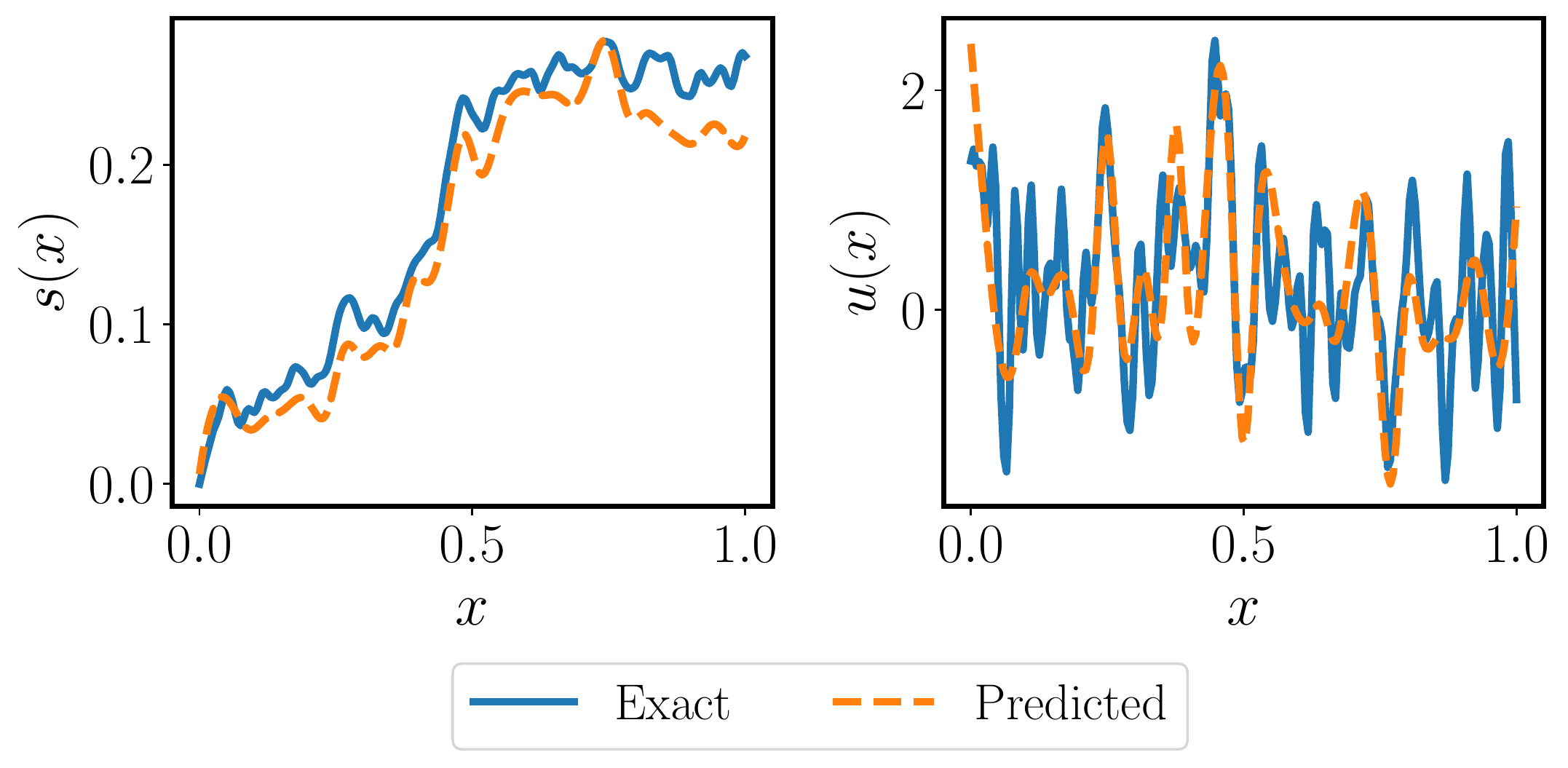}
         \caption{}
         \label{fig: physical_deeponet_MLP_s_u_2}
     \end{subfigure}
          \begin{subfigure}[b]{0.4\textwidth}
         \centering
         \includegraphics[width=\textwidth]{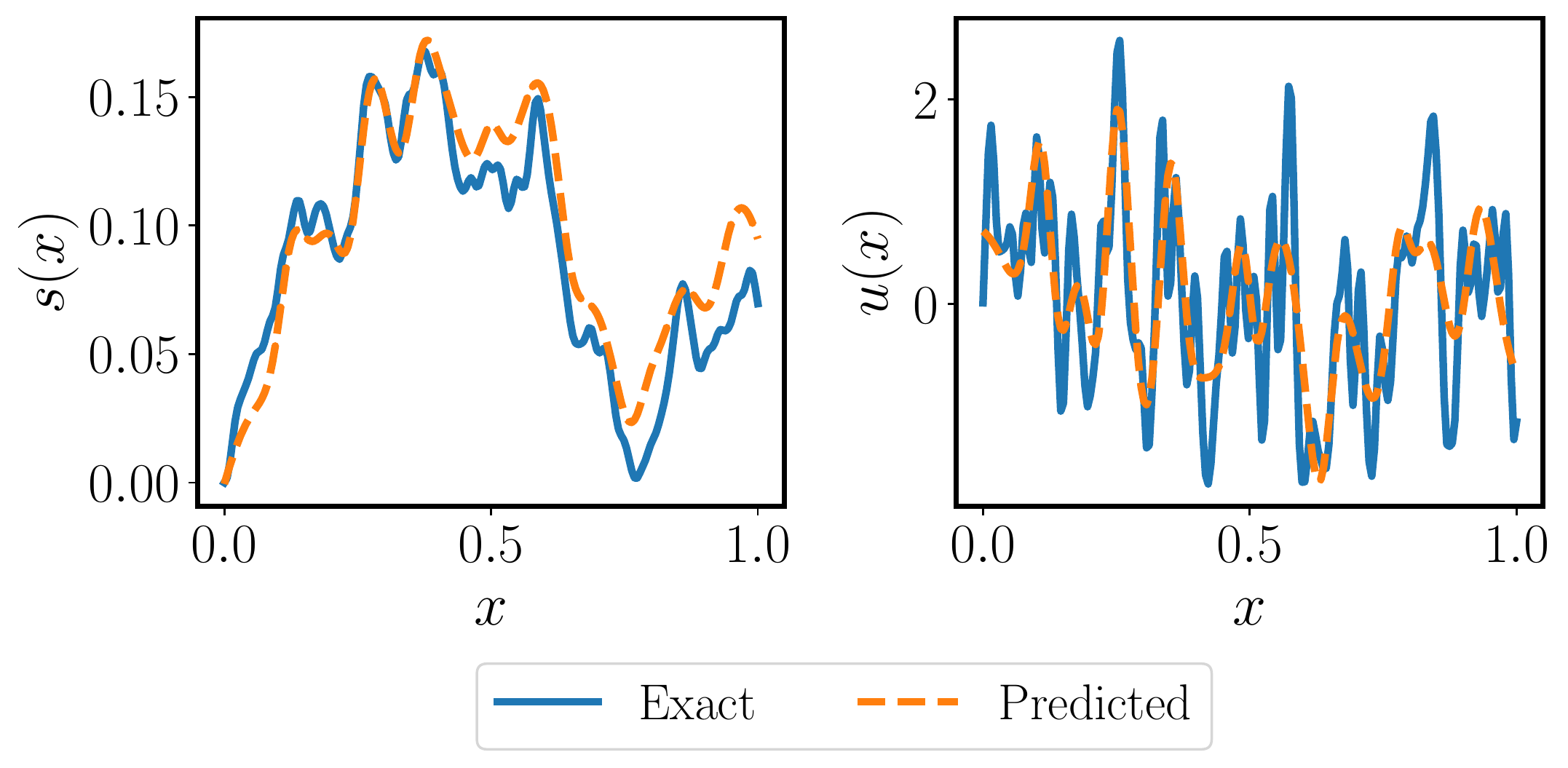}
         \caption{}
         \label{fig: physical_deeponet_MLP_s_u_3}
     \end{subfigure}
       \begin{subfigure}[b]{0.4\textwidth}
         \centering
         \includegraphics[width=\textwidth]{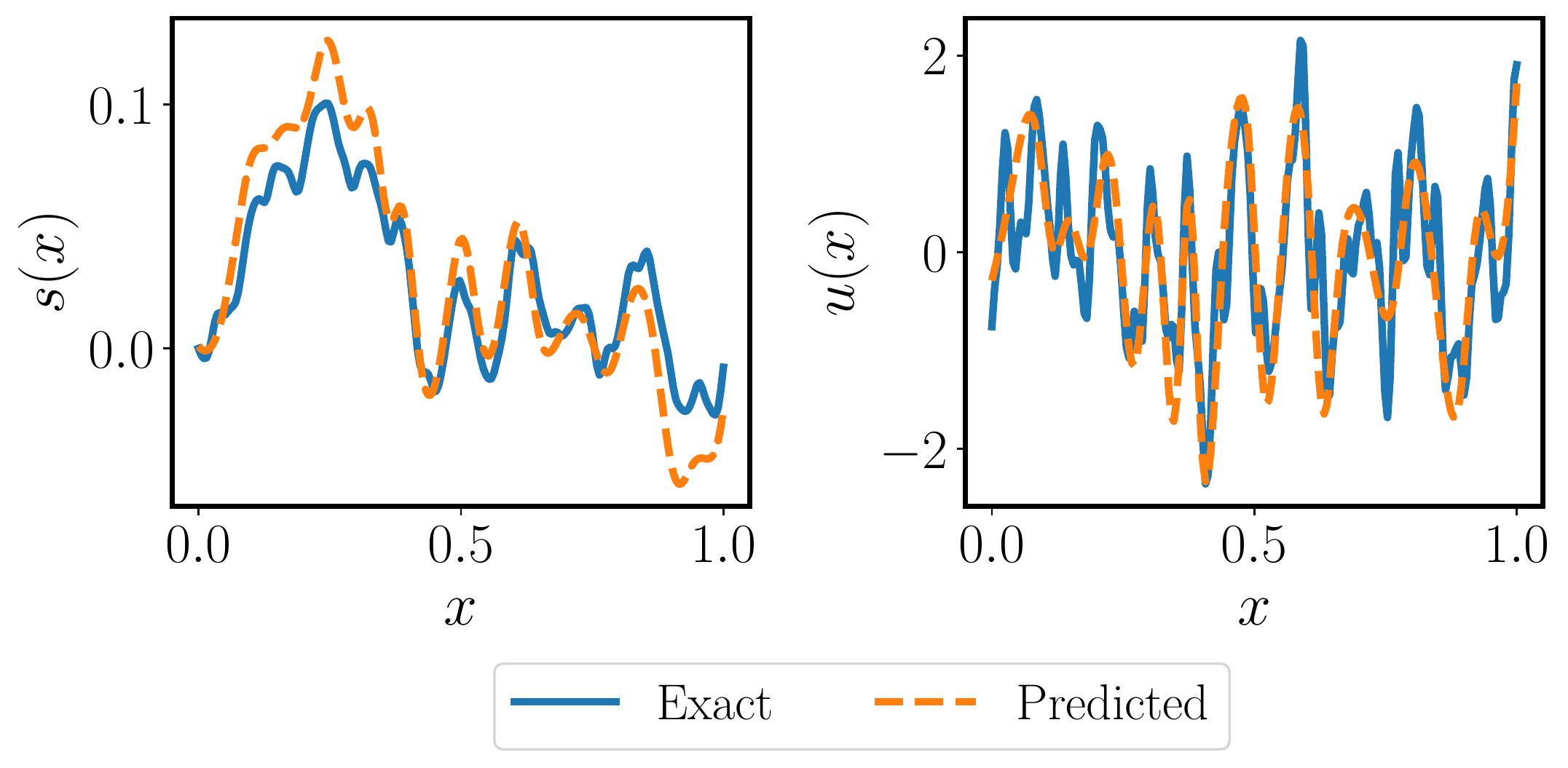}
         \caption{}
         \label{fig: physical_deeponet_MLP_s_u_4}
     \end{subfigure}
         \caption{{\em Solving a 1D parametric ODE with irregular input functions:} Predicted solutions $s(x)$ and corresponding ODE residuals $u(x)$ for a trained physics-informed DeepONet with a conventional fully-connected architecture, across four different examples in the test data-set. }
        \label{fig: physical_deeponet_MLP_s_u_examples}
\end{figure}

\begin{figure}[h]
     \centering
     \begin{subfigure}[b]{0.4\textwidth}
         \centering
         \includegraphics[width=\textwidth]{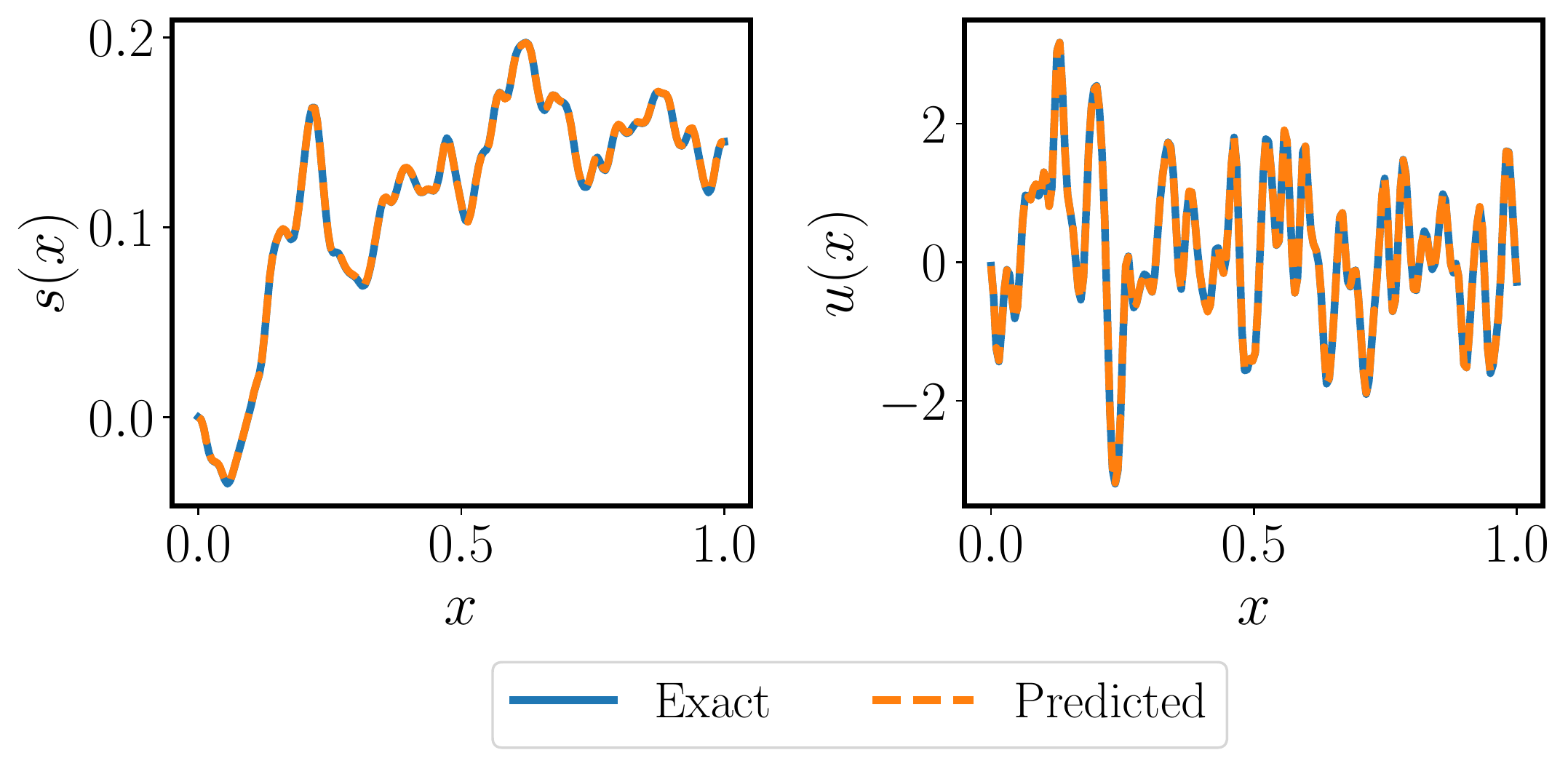}
         \caption{ }
         \label{fig: physical_deeponet_FF_s_u_1}
     \end{subfigure}
     \begin{subfigure}[b]{0.4\textwidth}
         \centering
         \includegraphics[width=\textwidth]{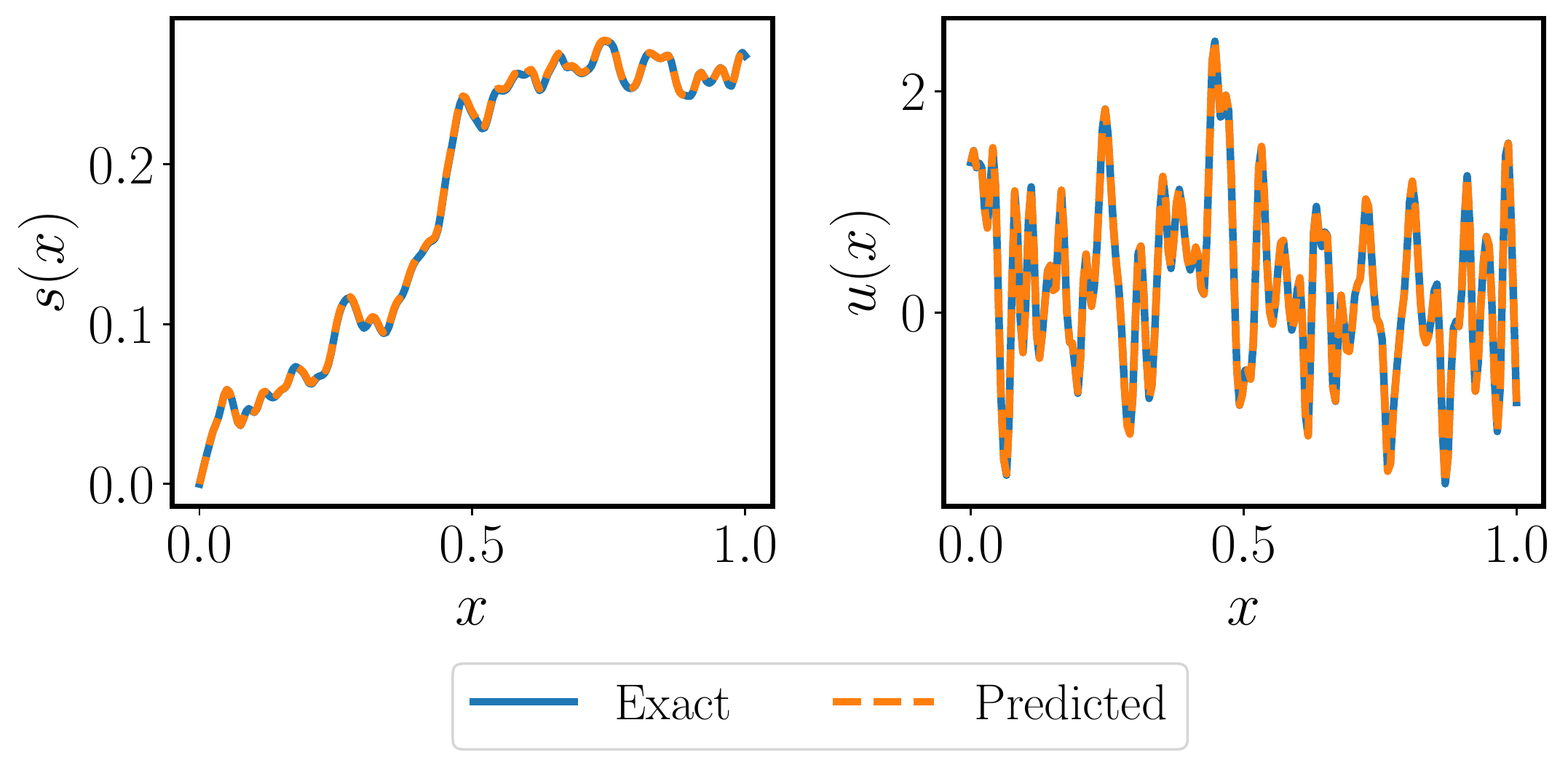}
         \caption{}
         \label{fig: physical_deeponet_FF_s_u_2}
     \end{subfigure}
          \begin{subfigure}[b]{0.4\textwidth}
         \centering
         \includegraphics[width=\textwidth]{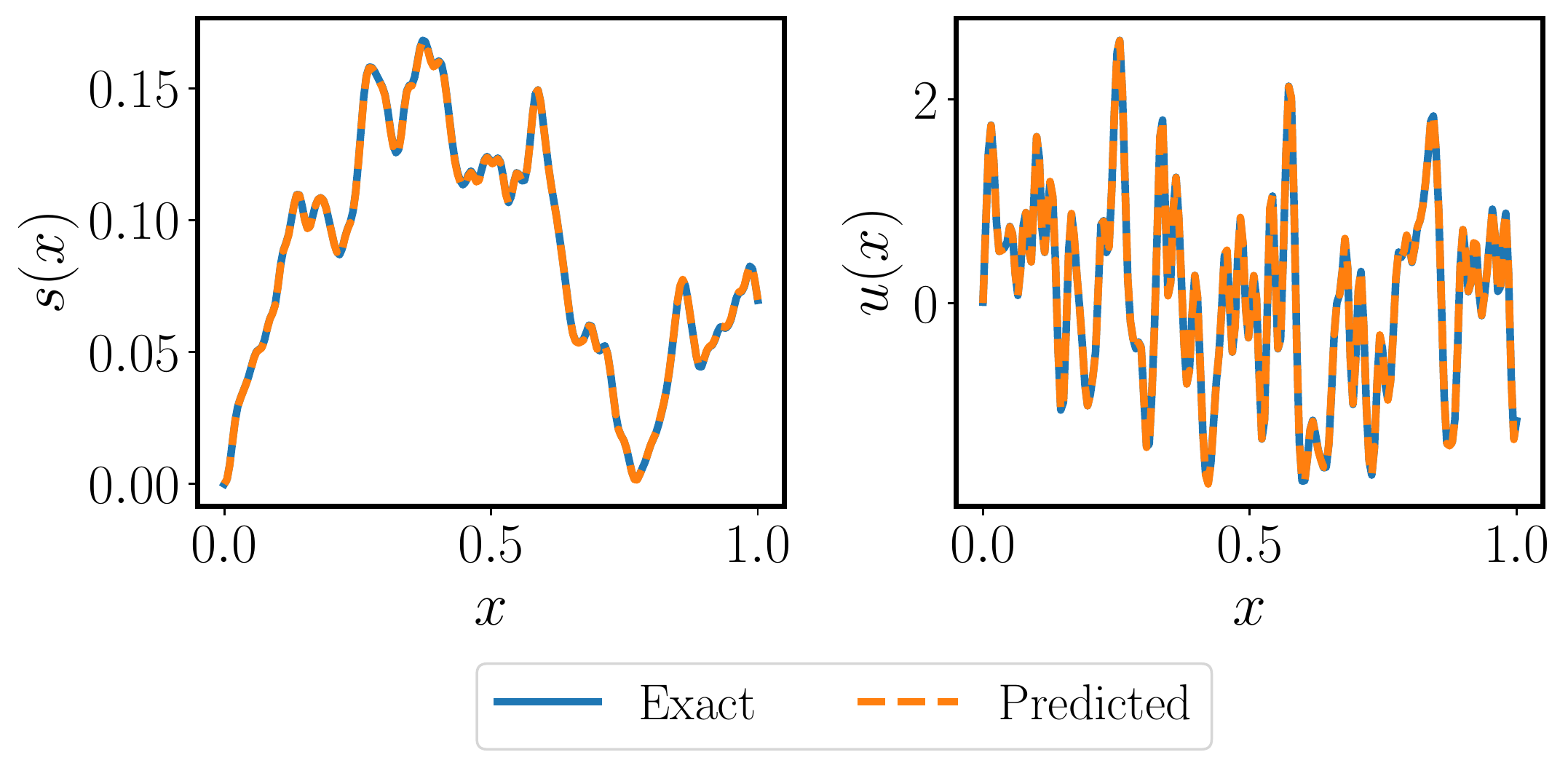}
         \caption{}
         \label{fig: physical_deeponet_FF_s_u_3}
     \end{subfigure}
       \begin{subfigure}[b]{0.4\textwidth}
         \centering
         \includegraphics[width=\textwidth]{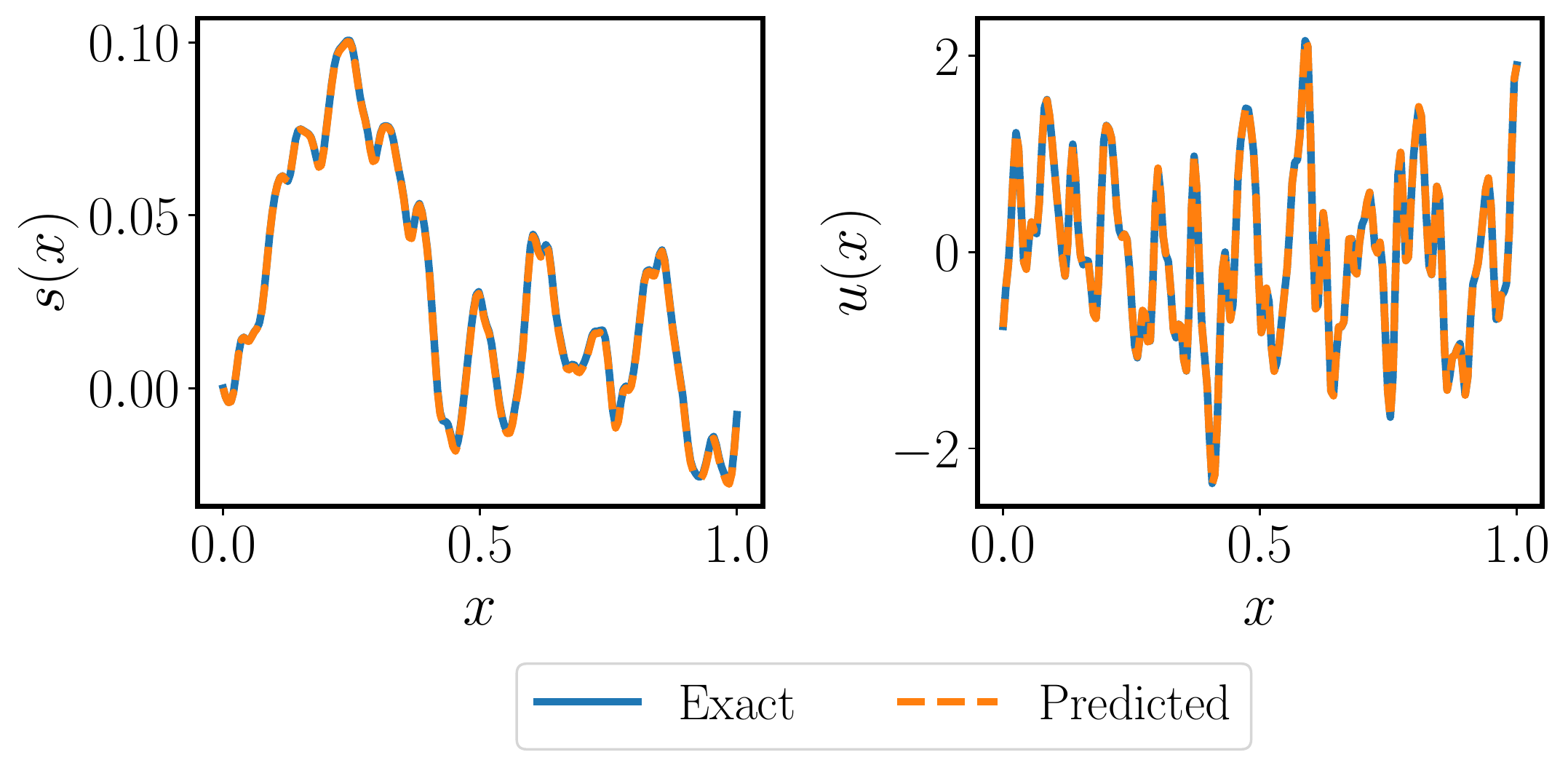}
         \caption{}
         \label{fig: physical_deeponet_FF_s_u_4}
     \end{subfigure}
         \caption{{\em Solving a 1D parametric ODE with irregular input functions:} Predicted solutions $s(x)$ and corresponding ODE residuals $u(x)$ for a trained physics-informed DeepONet with a with Fourier feature  architecture, across four different examples in the test data-set.}
        \label{fig: physical_deeponet_FF_s_u_examples}
\end{figure}

\begin{figure}[h]
     \centering
     \begin{subfigure}[b]{0.4\textwidth}
         \centering
         \includegraphics[width=\textwidth]{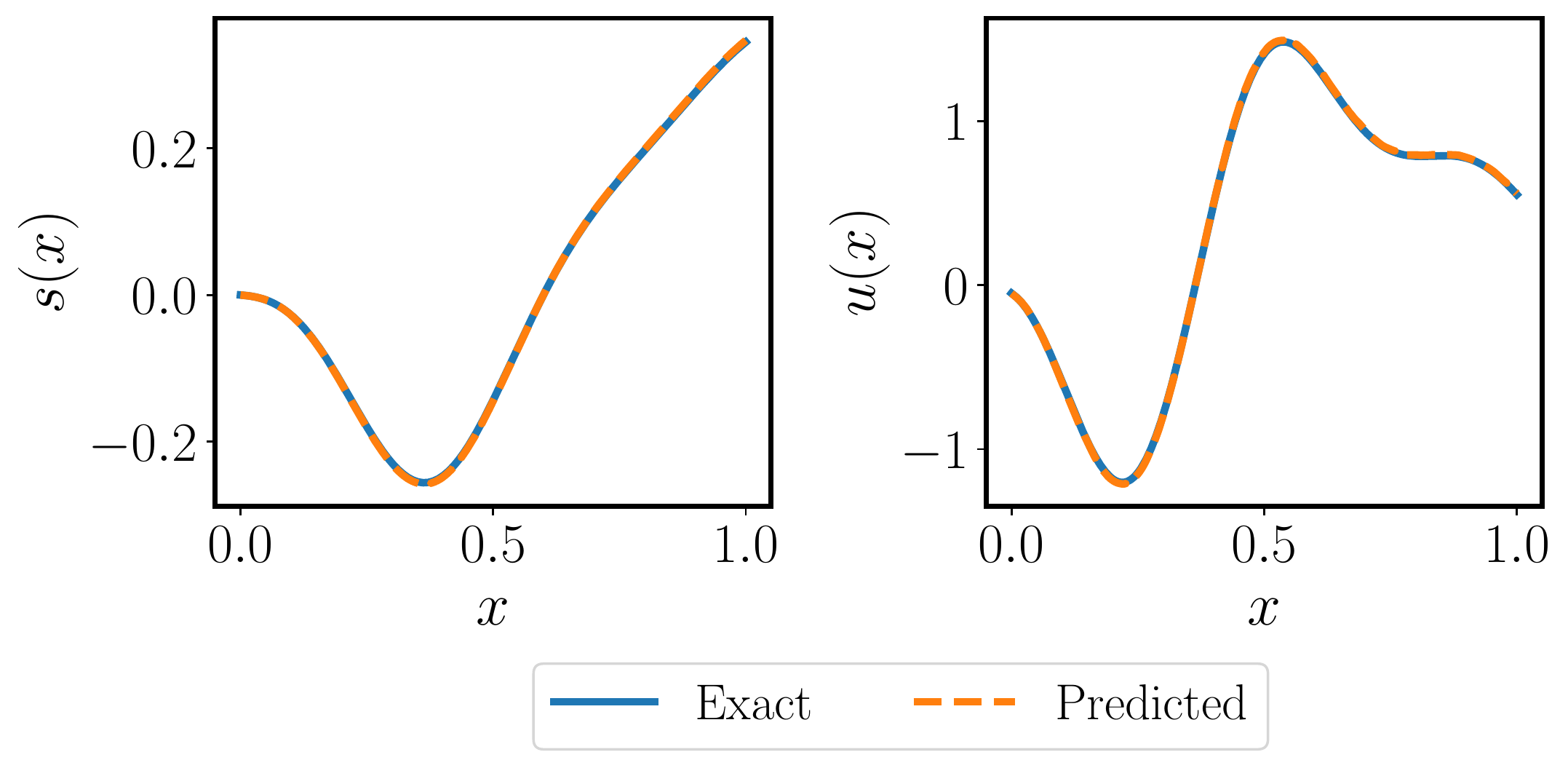}
         \caption{ }
         \label{fig: physical_deeponet_FF_02_s_u_1}
     \end{subfigure}
     \begin{subfigure}[b]{0.4\textwidth}
         \centering
         \includegraphics[width=\textwidth]{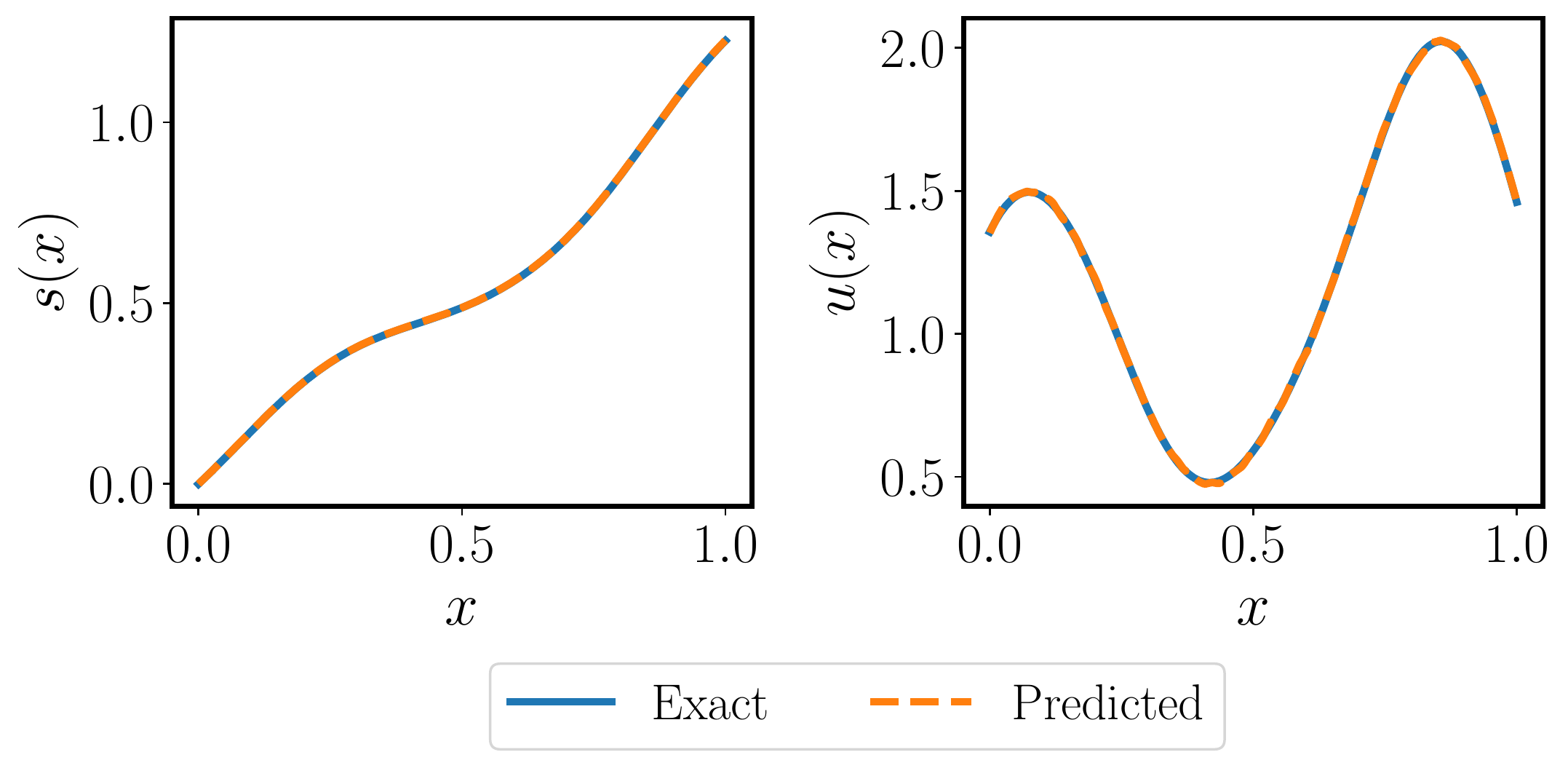}
         \caption{}
         \label{fig: physical_deeponet_FF_02_s_u_2}
     \end{subfigure}
          \begin{subfigure}[b]{0.4\textwidth}
         \centering
         \includegraphics[width=\textwidth]{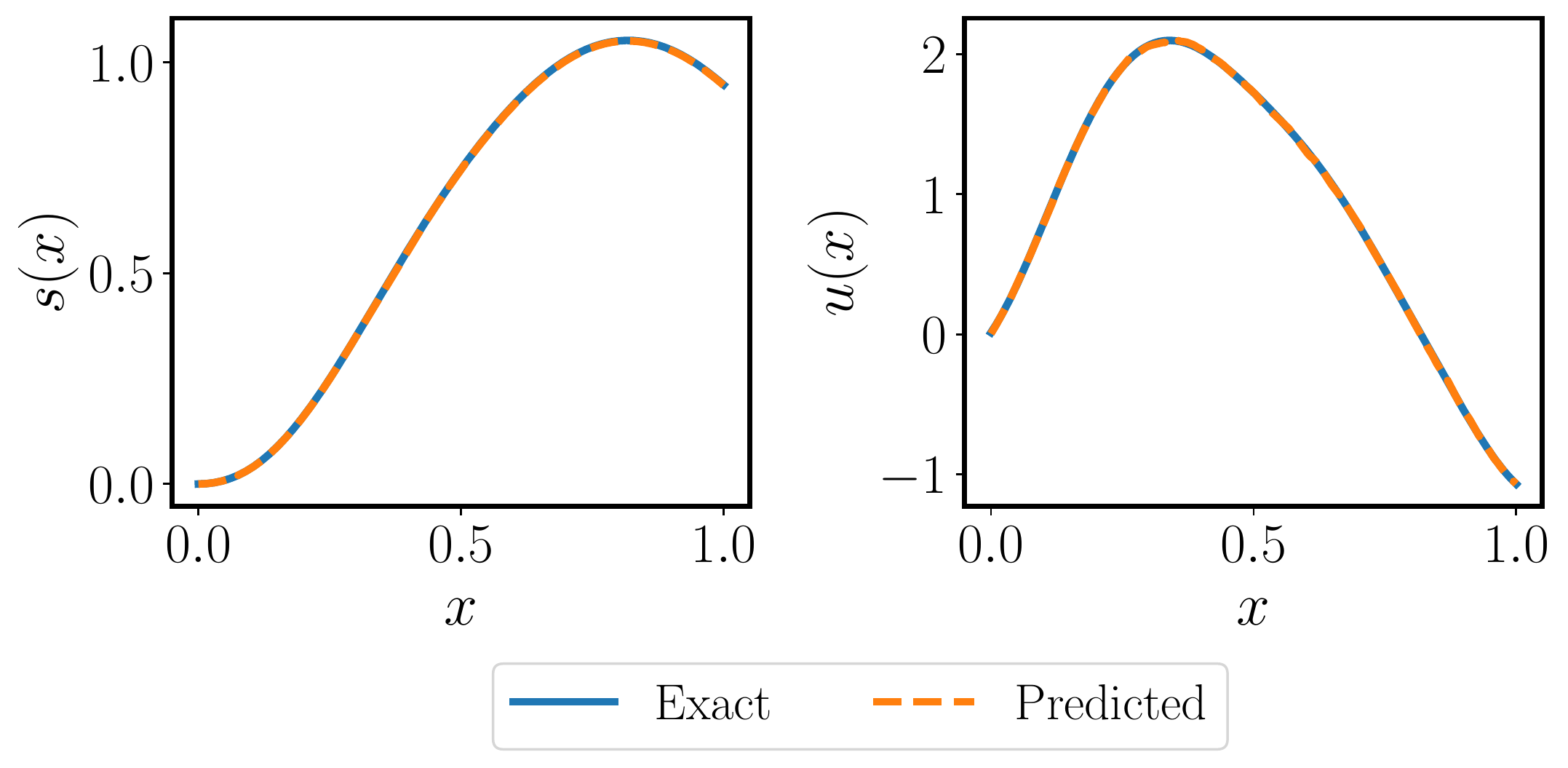}
         \caption{}
         \label{fig: physical_deeponet_FF_02_s_u_3}
     \end{subfigure}
       \begin{subfigure}[b]{0.4\textwidth}
         \centering
         \includegraphics[width=\textwidth]{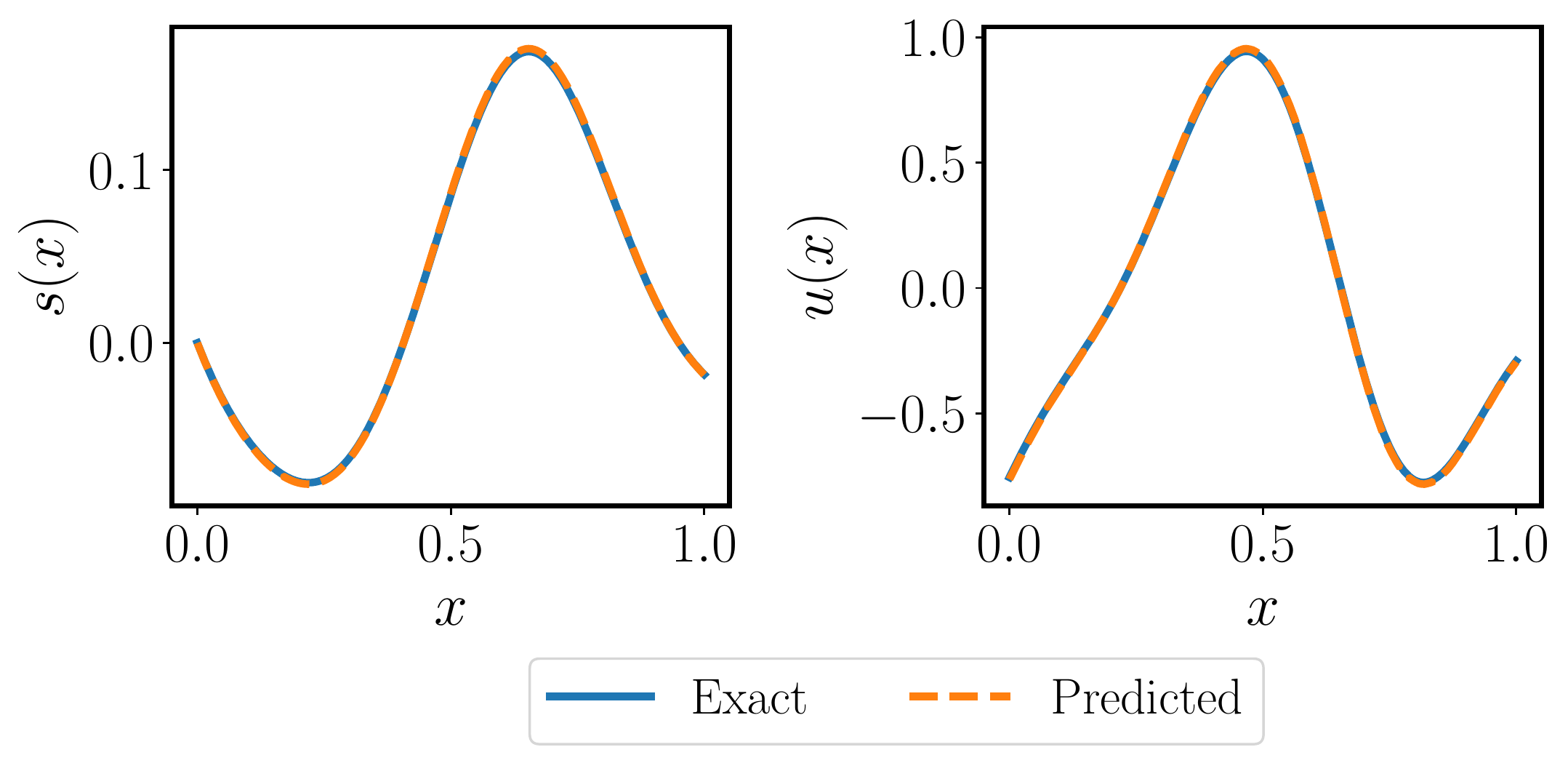}
         \caption{}
         \label{fig: physical_deeponet_FF_02_s_u_4}
     \end{subfigure}
         \caption{{\em Solving a 1D parametric ODE with irregular input functions:}   Predicted solutions $s(x)$ and corresponding ODE residuals $u(x)$ for a trained physics-informed DeepONet with a with Fourier feature  architecture, across four different out-of-distribution  examples sampled from a GRF with a length scale $l=0.2$ (recall that the training data for this case is generated using $l=0.01$).}
        \label{fig: physical_deeponet_FF_02_s_u_examples}
\end{figure}

\clearpage
\section{Diffusion-reaction system}
\label{appendix: DR}

\begin{figure}[h]
    \centering
   \begin{subfigure}[b]{0.4\textwidth}
         \centering
         \includegraphics[width=\textwidth]{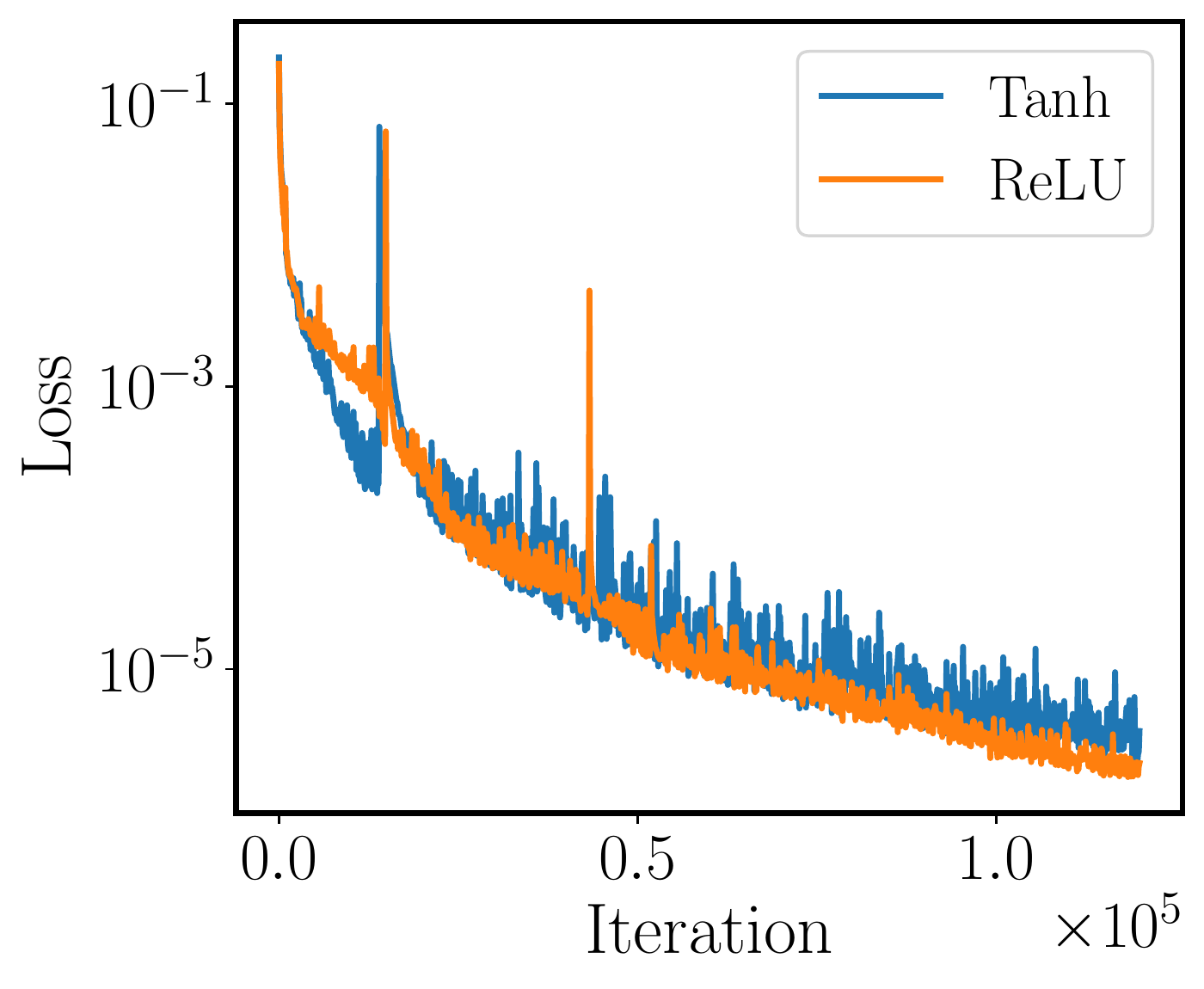}
         \caption{ }
         \label{fig: deeponet_DR_Losses}
     \end{subfigure}
     \begin{subfigure}[b]{0.4\textwidth}
         \centering
         \includegraphics[width=\textwidth]{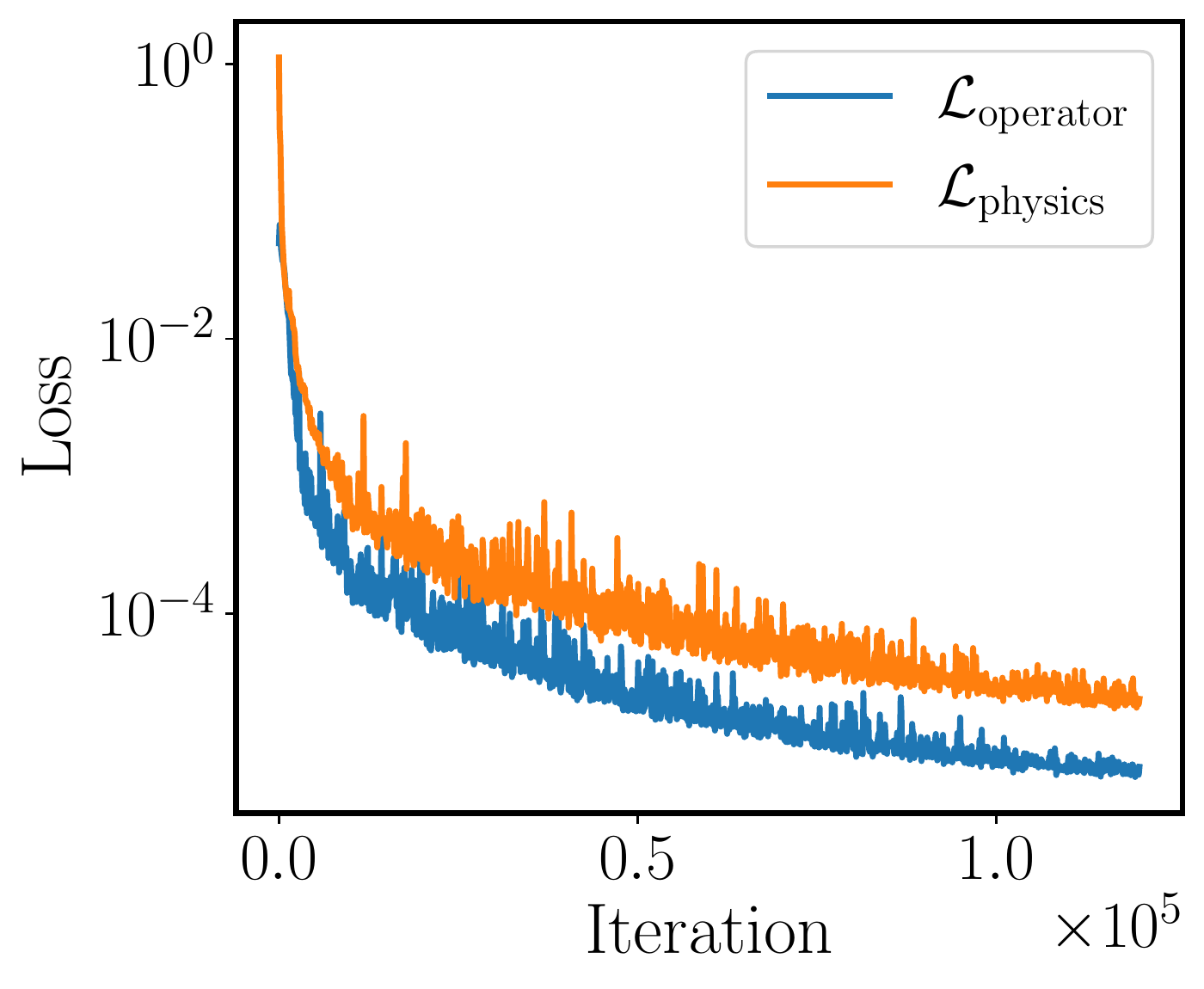}
        \caption{}
         \label{fig: physical_deeponet_DR_Losses}
     \end{subfigure}
   \caption{{\em Solving  a parametric diffusion-reaction system:} (a) Training loss convergence of a DeepONet equipped with different activations for 120,000 iterations of gradient descent using the Adam optimizer (with paired input-output training data). (b) Training loss convergence of a  physics-informed DeepONet equipped with Tanh activations for 120,000 iterations of gradient descent using the Adam optimizer (without paired input-output training data). }
    \label{fig: DR_Losses}
\end{figure}

\begin{figure}[h]
    \centering
    \includegraphics[width=0.8\textwidth]{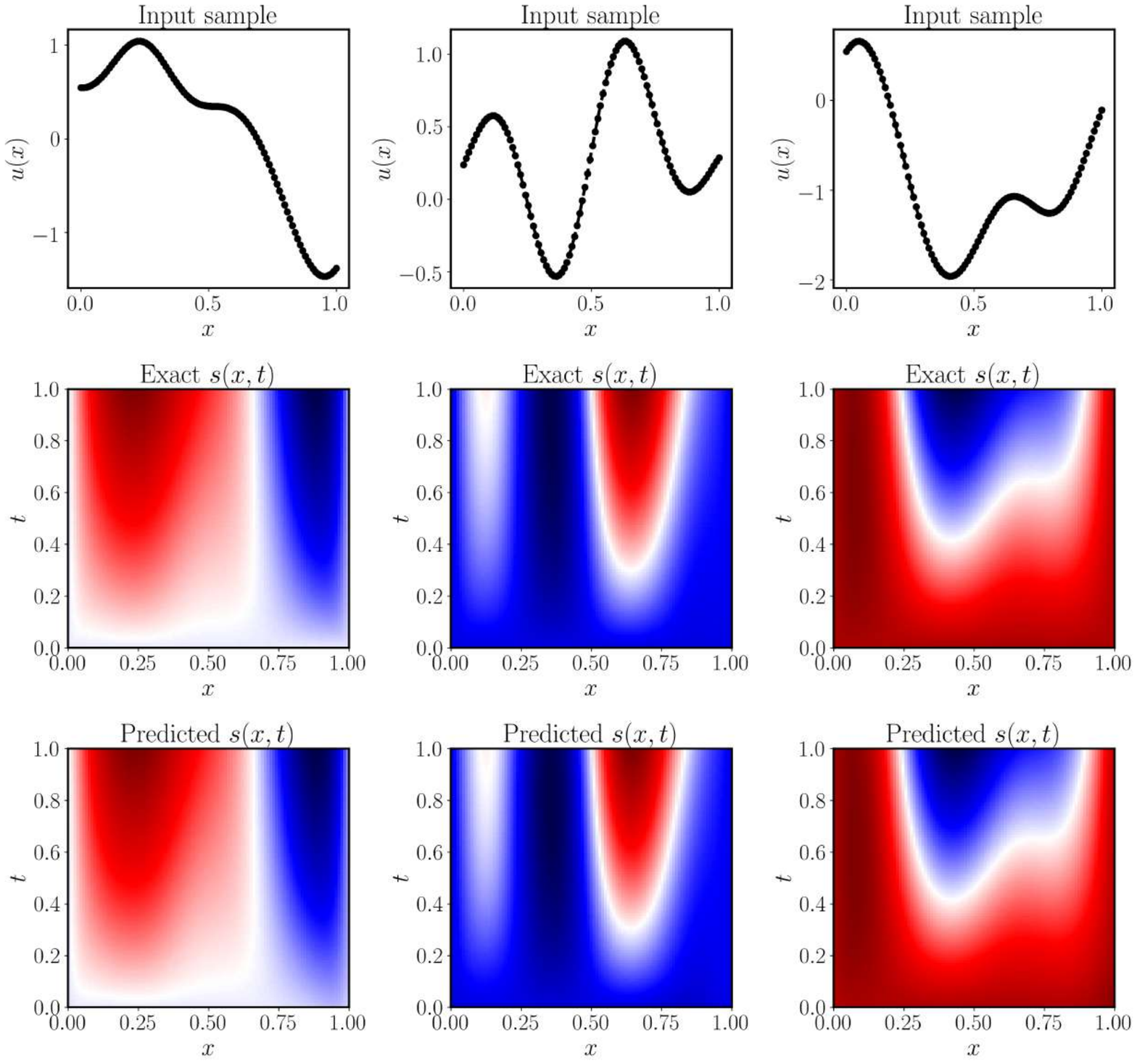}
    \caption{{\em Solving  a parametric diffusion-reaction system:} Predicted solution of a trained physics-informed DeepONet for three different examples in the test data-set.}
    \label{fig: physical_deeponet_DR_2}
\end{figure}

\clearpage
\section{Effect of the batch-size}
\label{appendix: DR_batch_size}

\begin{figure}[h]
    \centering
    \includegraphics[width=0.4\textwidth]{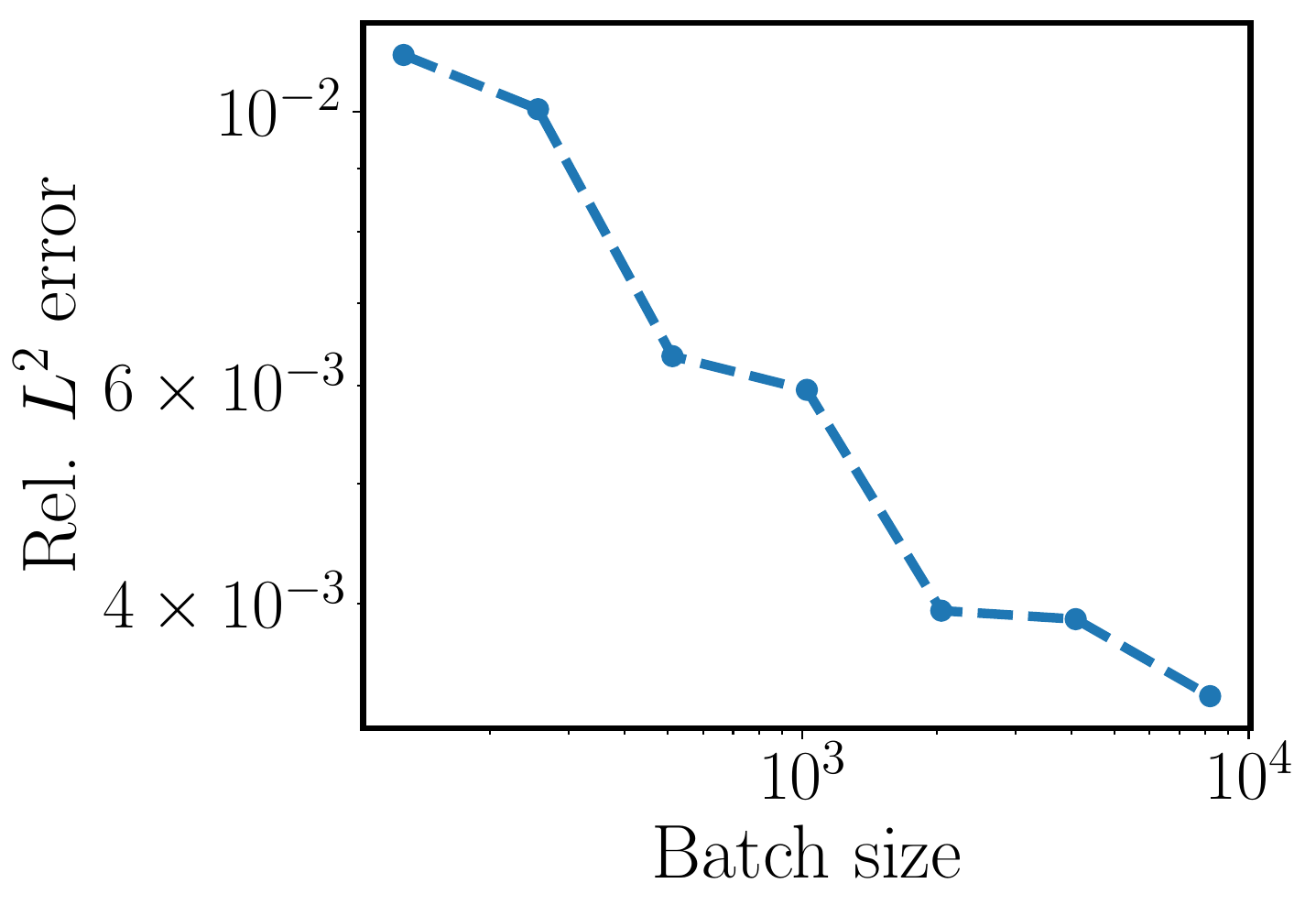}
    \caption{{\em Solving  a parametric diffusion-reaction system:} Relative $L^2$ prediction error of physics-informed DeepONets trained using a different batch-size, averaged over 1,000 examples in the test data-set.}
    \label{fig: DR_batch_size}
\end{figure}

\begin{figure}[h]
    \centering
        \begin{subfigure}[b]{0.4\textwidth}
         \centering
         \includegraphics[width=\textwidth]{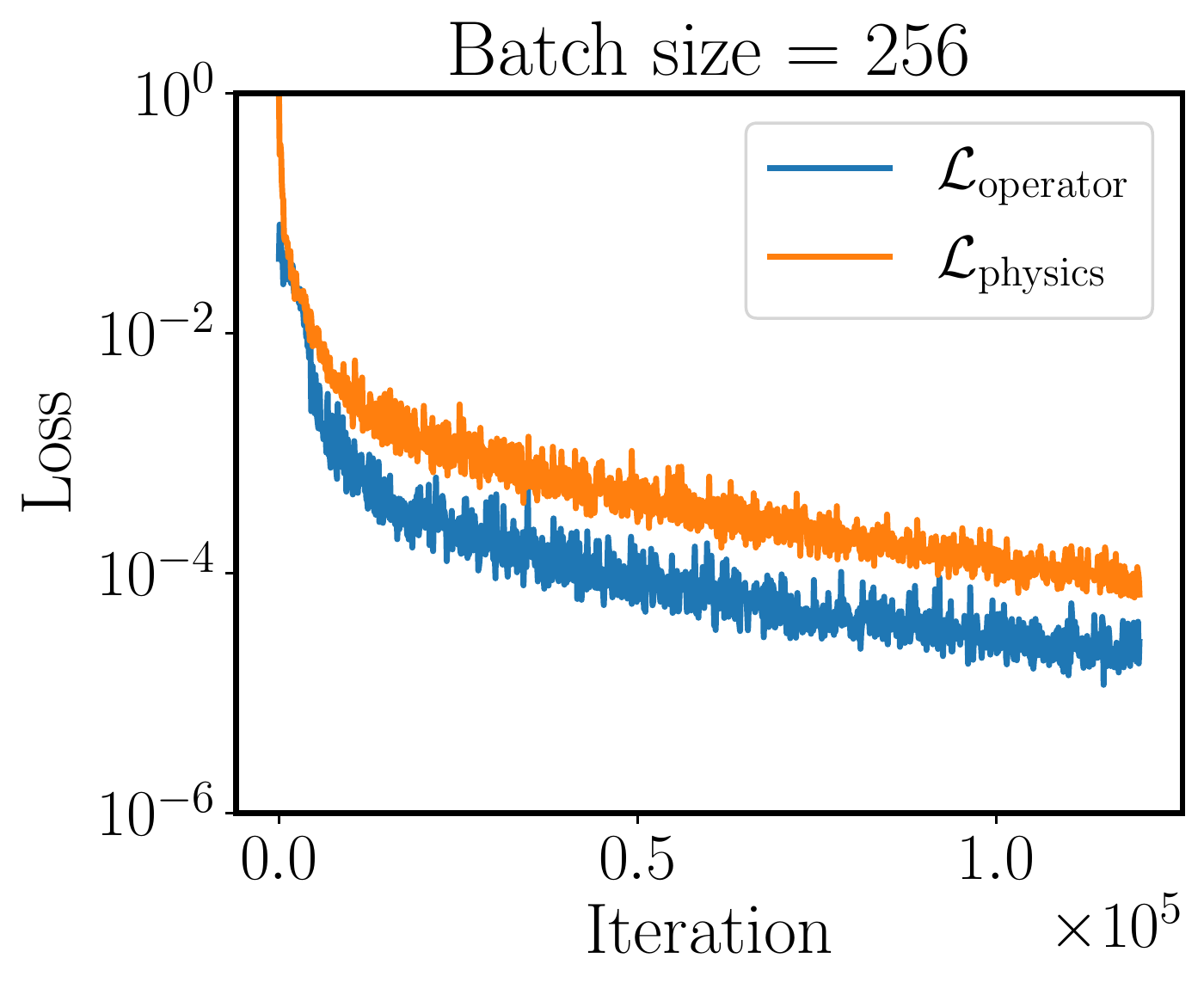}
     \end{subfigure}
        \begin{subfigure}[b]{0.4\textwidth}
         \centering
         \includegraphics[width=\textwidth]{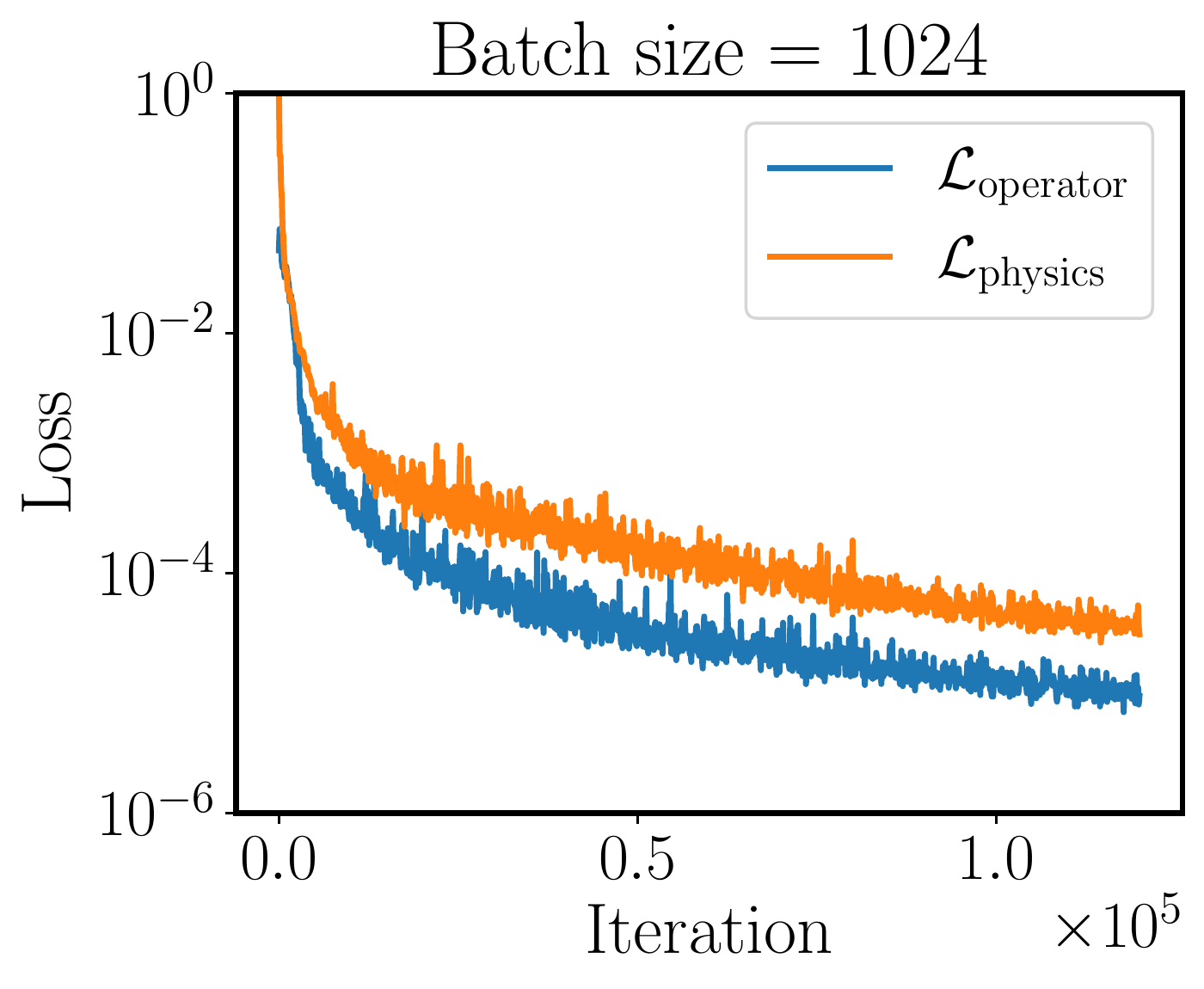}
     \end{subfigure}
        \begin{subfigure}[b]{0.4\textwidth}
         \centering
         \includegraphics[width=\textwidth]{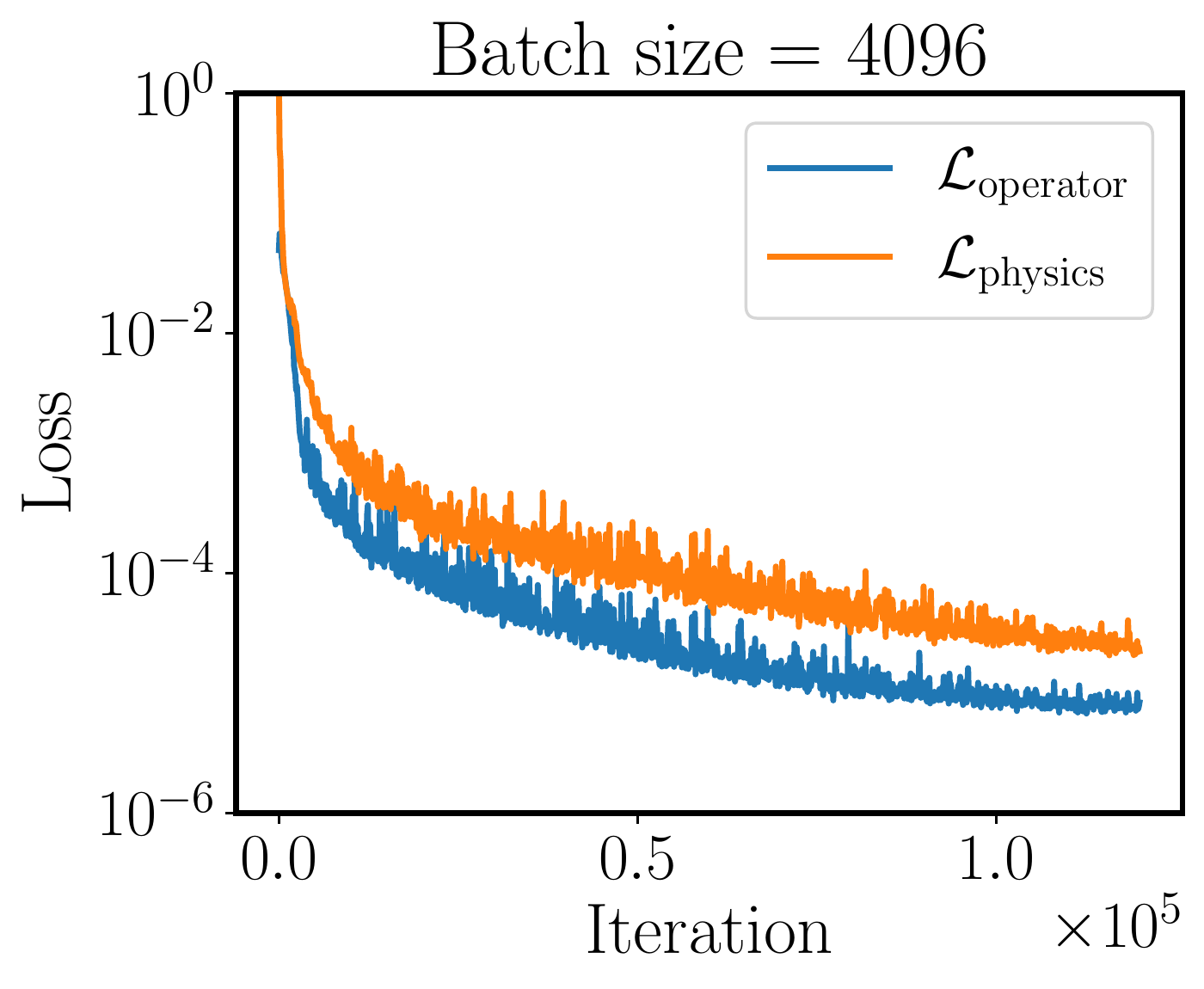}
     \end{subfigure}
        \begin{subfigure}[b]{0.4\textwidth}
         \centering
         \includegraphics[width=\textwidth]{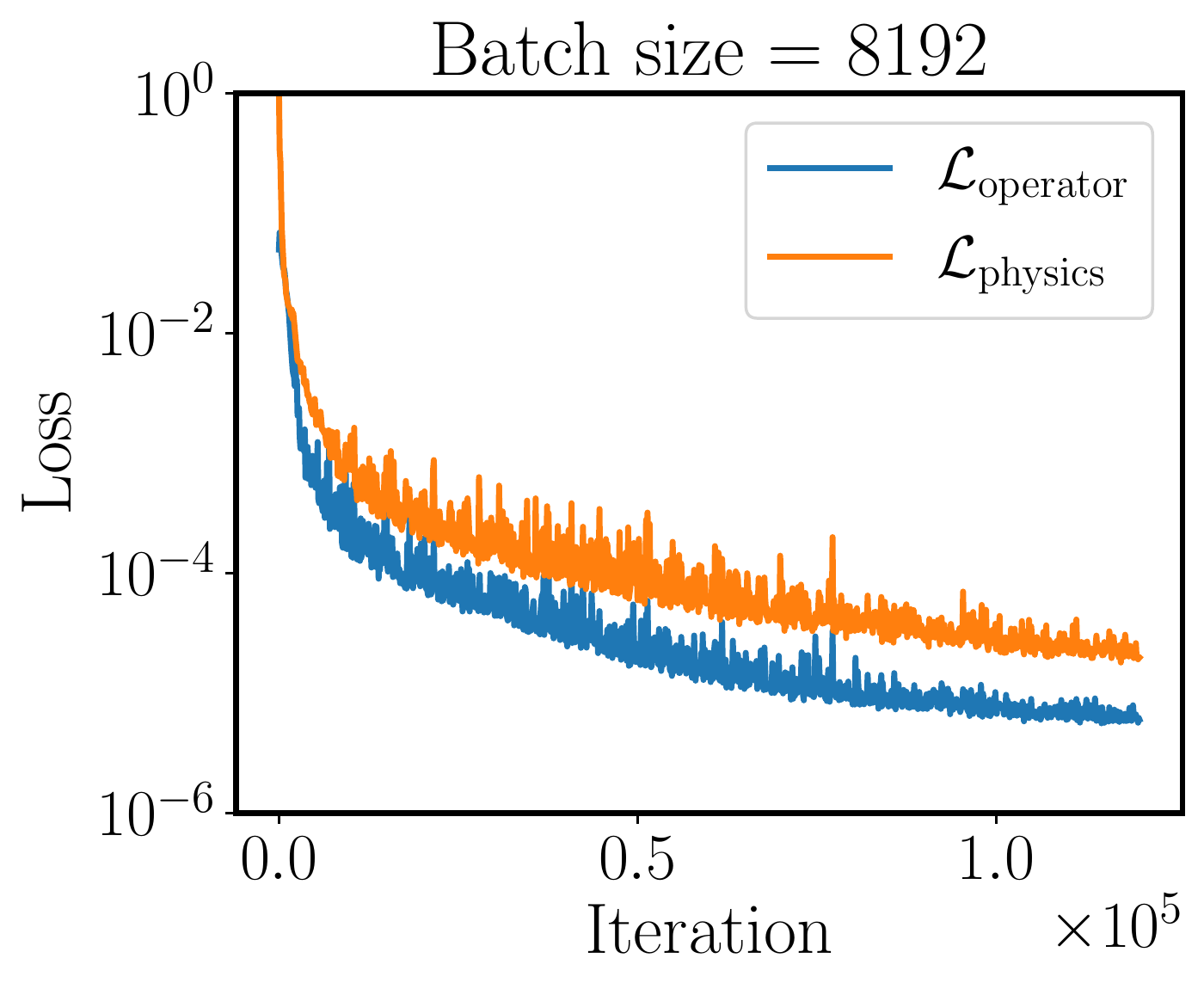}
     \end{subfigure}
   \caption{{\em Solving  a parametric diffusion-reaction system:}  Training loss convergence of physics-informed DeepONets trained using a different batch-size.}
    \label{fig: DR_batch_size_loss}
\end{figure}

\clearpage
\section{Burgers' equation}
\label{Appendix: Burger}

\subsection{Modified full-connected neural network}
The forward pass of the proposed modified MLP architecture is given by \cite{wang2020understanding}
\begin{align}
    &U = \phi(X W^1 + b^1), \ \  V = \phi(X W^2 + b^2) \\
    &H^{(1)} = \phi(X W^{z,1} + b^{z, 1}) \\
    &Z^{(k)} = \phi(H^{(k)}W^{z,k} + b^{z, k}), \ \ k=1, \dots, L \\
    &H^{(k+1)} = (1 - Z^{(k)}) \odot U  +  Z^{(k)}  \odot V, \ \  k=1, \dots, L \\
   & f_{\bm{\theta}}(x) = H^{(L+1)}W  + b,
\end{align}
where $X$ denotes the network inputs, and $\odot$ denotes element-wise multiplication. The parameters of this model are essentially the same as in a standard fully-connected architecture, with the addition of the weights and biases used by the two transformer networks, i.e.,
\begin{align}
    \theta = \{W^1, b^1, W^2, b^2, (W^{z,l}, b^{z,l})_{l=1}^L, W,b  \}
\end{align}

\begin{figure}[h]
    \centering
    \includegraphics[width=0.7\textwidth]{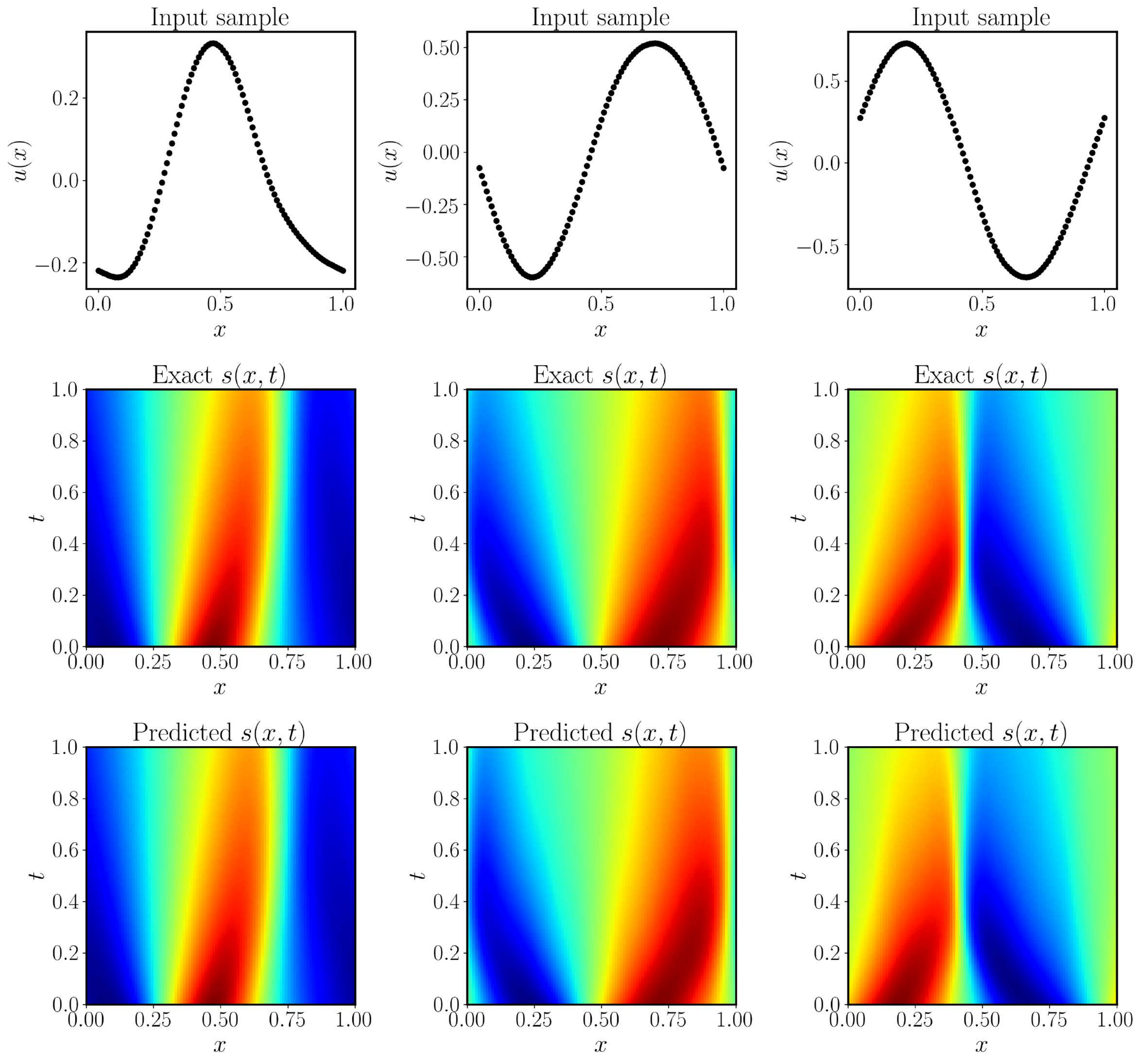}
    \caption{{\em Solving  a Burgers' equation:} Predicted solution of a trained physics-informed DeepONet with a modified MLP architecture for three different examples in the test data-set.}
    \label{fig: physical_deeponet_Burger_examples}
\end{figure}

\clearpage
\section{Eikonal equations}

\begin{figure}[h]
    \centering
   \begin{subfigure}[b]{0.4\textwidth}
         \centering
         \includegraphics[width=\textwidth]{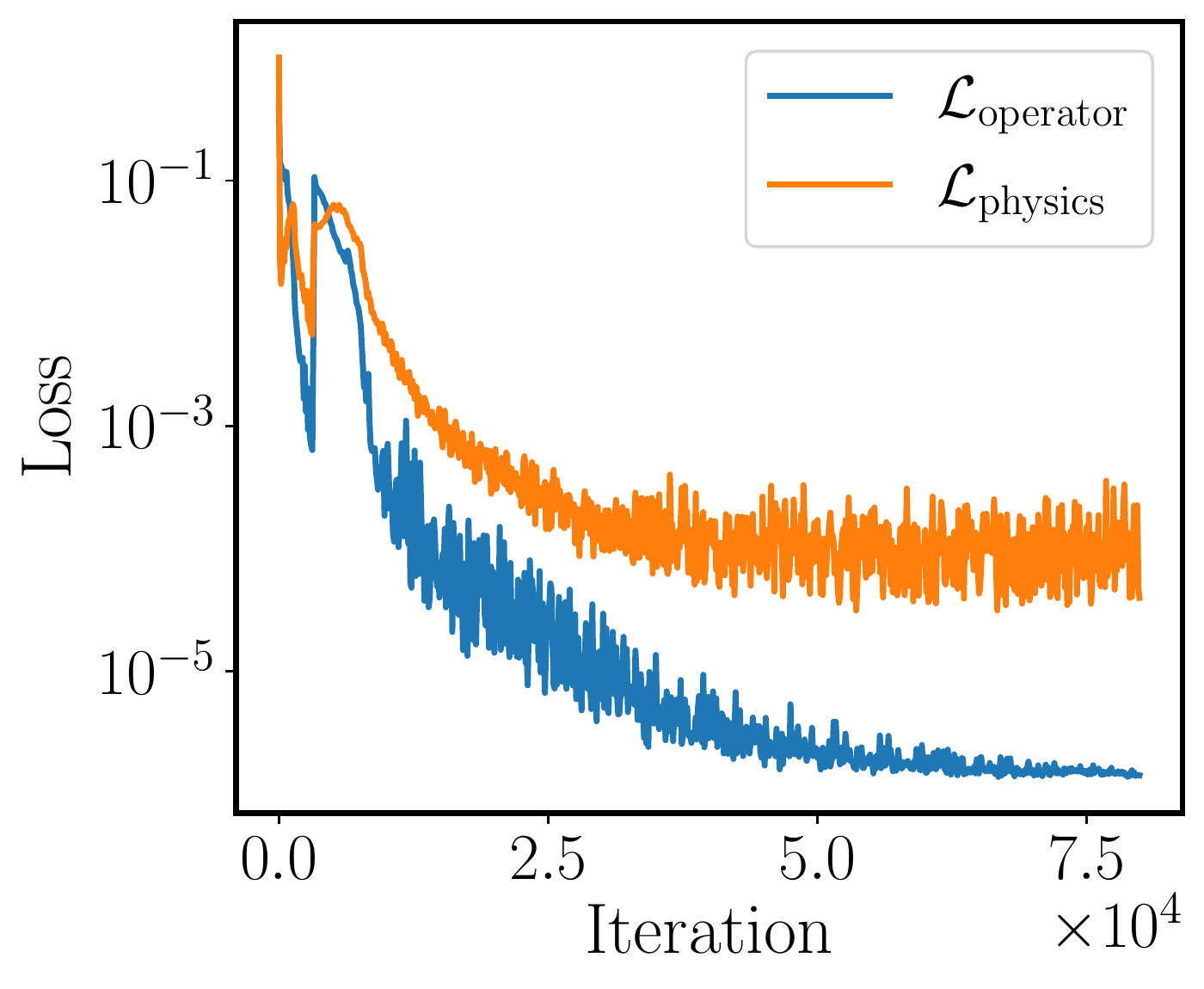}
         \caption{ }
         \label{fig: physical_deeponet_eikonal_circle_loss}
     \end{subfigure}
     \begin{subfigure}[b]{0.4\textwidth}
         \centering
         \includegraphics[width=\textwidth]{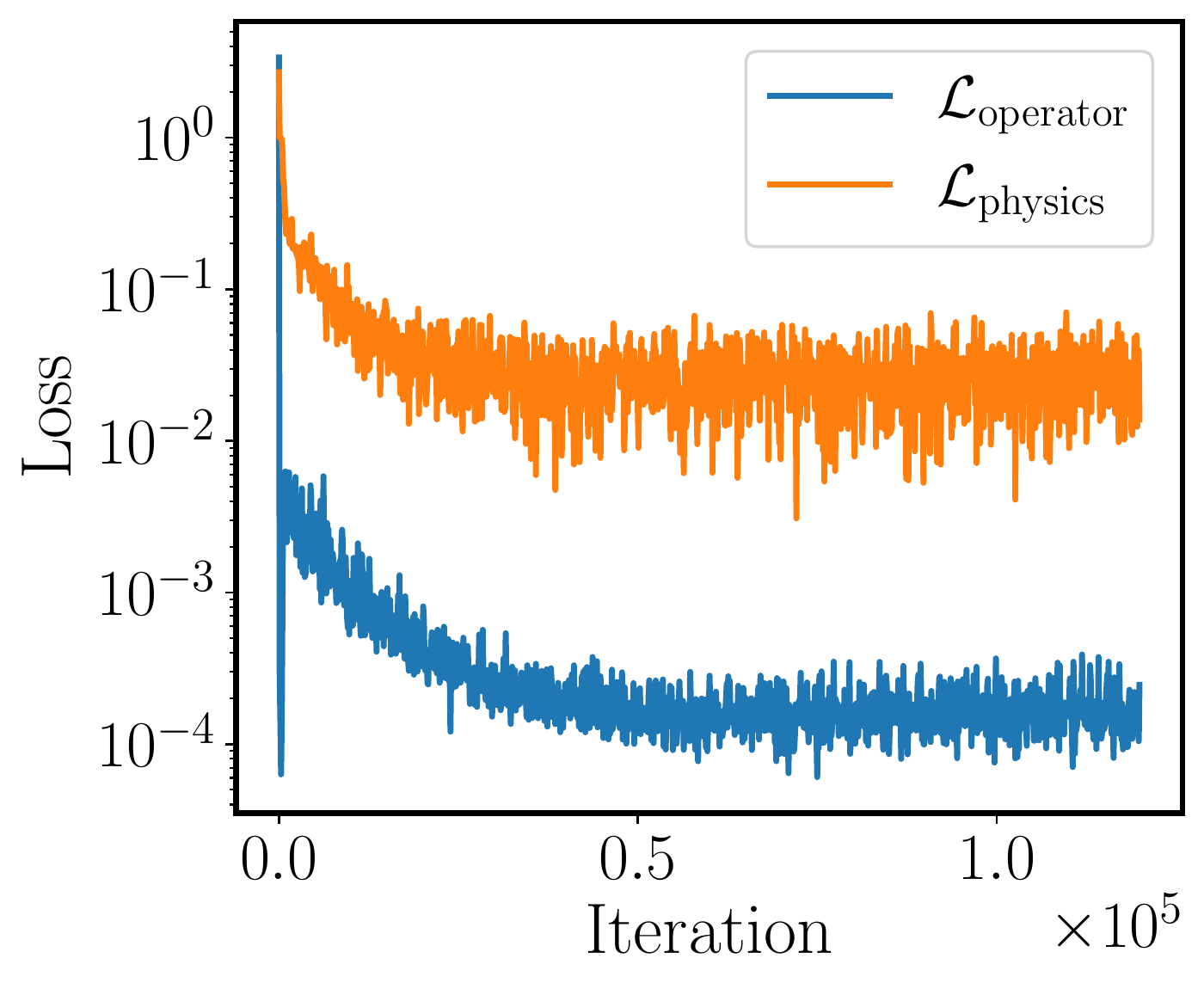}
        \caption{}
         \label{fig: physical_deeponet_eikonal_airfoil_loss}
     \end{subfigure}
   \caption{{\em Solving a parametric Eikonal equation:}  (a) Training loss convergence of a physics-informed DeepONet equipped with Tanh activations, for 80,000 iterations of gradient descent using the Adam optimizer. (b) Training loss convergence of a  physics-informed DeepONet equipped with ELU activations, for 120,000 iterations of gradient descent using the Adam optimizer. }
    \label{fig: physical_deeponet_eikonal_loss}
\end{figure}

\begin{figure}[h]
    \centering
    \includegraphics[width=0.8\textwidth]{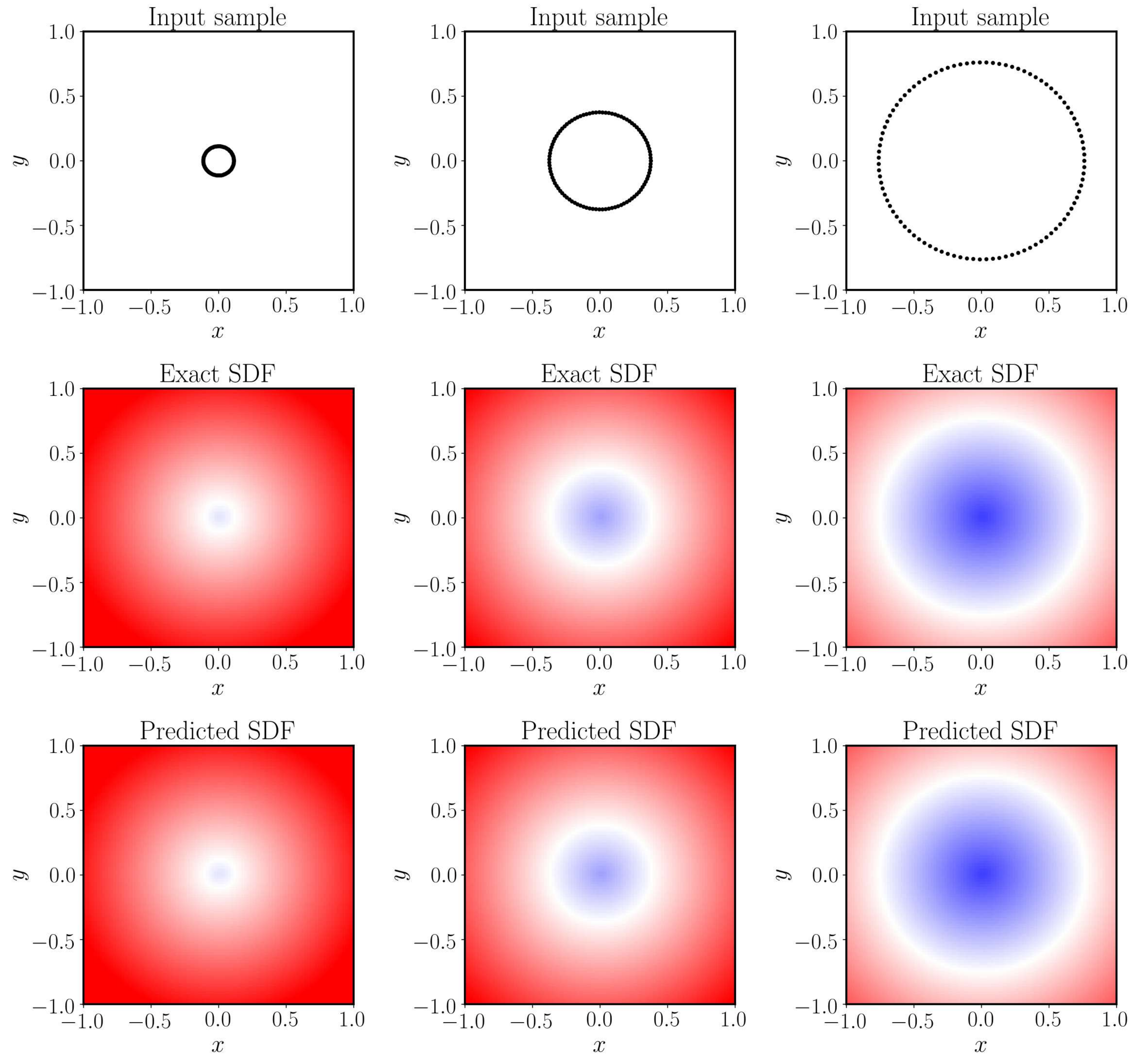}
   \caption{{\em Solving  a parametric Eikonal equation (circles):} Predicted signed distance functions by a trained physics-informed DeepONet for three different examples in the test data-set.}
    \label{fig: physical_deeponet_eikonal_circle_sdf_examples}
\end{figure}